%% file: main.tex
\documentclass[a4paper,fleqn]{cas-dc}

 \usepackage[numbers,compress]{natbib}

\usepackage{url}

\def\tsc#1{\csdef{#1}{\textsc{\lowercase{#1}}\xspace}}
\tsc{WGM}
\tsc{QE}
\tsc{EP}
\tsc{PMS}
\tsc{BEC}
\tsc{DE}
\usepackage{tikz}
\usepackage{float}
\usepackage{graphicx} 
\usepackage{svg}
\usepackage[percent]{overpic}
\usepackage{tabularx}
\usetikzlibrary{positioning}
\usepackage{amsmath,amsfonts,mathtools}
\newcommand*\diff{\mathop{}\!\mathrm{d}}
\DeclareMathOperator{\sign}{sign}
\usepackage{pgfplots}
\pgfplotsset{compat=newest}
\pgfplotsset{every x tick scale label/.style={ at={(1,0)},xshift=1pt,anchor=south west,inner sep=0pt}, /pgf/number format/fixed, /pgf/number format/precision=3, legend style={nodes={scale=0.7, transform shape}}, scaled ticks = false, every axis plot/.append style={thick}}
\newlength\fheight
\newlength\fwidth 
\graphicspath{{figs/}}

\usepackage{listings}
\usepackage{xcolor}
\begin{document}
\let\WriteBookmarks\relax
\def\floatpagepagefraction{1}
\def\textpagefraction{.001}
\shorttitle{Flatness Based Control of an Industrial Robot Joint Using Secondary Encoders}
\shortauthors{J. Weigand et~al.}

\title[mode = title]{Flatness Based Control of an Industrial Robot Joint Using Secondary Encoders}                      

 

 \author[1]{Jonas Weigand}[type=editor,
 orcid=0000-0001-5835-3106]
 \ead{jonas.weigand@mv.uni-kl.de} 
 \cormark[1]
 \author[1]{Nigora Gafur}[type=editor,
 orcid=0000-0002-3131-3574]  
 \author[1,2]{Martin Ruskowski}[type=editor,
 orcid=0000-0002-6534-9057]

\address[1]{Chair of Machine Tools and Control Systems (WSKL), TU Kaiserslautern, Gottlieb-Daimler-Str. 42, 67663 Kaiserslautern, Germany}
\address[2]{Innovative Factory Systems (IFS), German Research Center for Artificial Intelligence (DFKI), Trippstadter Str. 122, 67663 Kaiserslautern, Germany}

\begin{abstract}
Due to their compliant structure, industrial robots without precision-enhancing measures are only to a limited extent suitable for machining applications. Apart from structural, thermal and bearing deformations, the main cause for compliant structure is backlash of transmission drives. This paper proposes a method to improve trajectory tracking accuracy by using secondary encoders and applying a feedback and a flatness based feed forward control strategy. \ For this purpose, a novel nonlinear, continuously differentiable dynamical model of a flexible robot joint is presented. The robot joint is modeled as a two-mass oscillator with pose-dependent inertia, nonlinear friction and nonlinear stiffness, including backlash. A flatness based feed forward control is designed to improve the guiding behaviour and a feedback controller, based on secondary encoders, is implemented for disturbance compensation. Using Automatic Differentiation, the nonlinear feed forward controller can be computed in a few microseconds online. Finally, the proposed algorithms are evaluated in simulations and experimentally on a real KUKA Quantec KR300 Ultra SE.


\end{abstract}




\maketitle

\input{sec_intro}

\input{sec_modeling}

\input{sec_control_design_v2}

\input{sec_simulation}

\input{sec_results_v2}

\section*{Appendix}
\appendix
\input{sec_appendix_v3}




\bibliographystyle{model1-num-names}

\bibliography{references}


\end{document}

%% file: sec_intro.tex
\section{Introduction}
Industrial robots are highly flexible due to their open and complex kinematic chain and are mainly designed for a good repeatability but not for high precision tasks, such as milling \cite{Olabi2012}. The main drawback of using an industrial robot in machining processes is the lack of high position accuracy caused by low stiffness, meaning low eigenfrequencies and joint elasticity between the actuators and the driven links \cite{Freising2014,Schneider2014}. As a result, conventional computerized numerical control (CNC) machines, possessing high stiffness and simple kinematics, can still not be replaced by industrial robots for machining tasks \cite{Iglesias2015,Wu2018DynamicController}. The low and highly pose dependent overall stiffness of industrial robots, which is up to 100 times less than of conventional CNC machines, can lead to vibrations and chattering effects, so that the machining quality suffers \cite{Brunete2018,Yuan2018,Vieler2017}. Already the influence of the placement of a workpiece with respect to the robot 
can lead to different machining results due to pose dependent stiffness of industrial robots \cite{Lin2017PostureIndexes}. Two different methods to increase the machining accuracy of industrial robots are reported in the literature, either to use a model-based approach accounting for compliance of flexible joints or to use a sensor-based approach by tracking deformations \cite{Olabi2012,Schneider2014,Wang2009,Frommknecht2017Multi-sensorDrilling}. The sensor-based approach can lead to a higher position accuracy than a model-based approach \cite{Schneider2014}. 
Attaching secondary encoders (SE) to the link-side can reduce the oscillatory behaviour and chattering effects by providing sensor information to a feedback controller. An improvement of static precision by a factor of $10$ was achieved in \cite{Devlieg2011} using SE. However, the approach is associated with high costs, high implementation effort, communication delays and sensor noise \cite{Schneider2014}. Despite of that, reducing dynamic deflections, due to time-varying dynamical effects, still remains a challenge that can only be handled by a model-based approach. 

Modeling joint flexibility involves considering stiffness and friction between motor and the driven link, as well as effects occurring in transmission drives, such as backlash, lost-motion and hysteresis effects. High precision transmission drives, such as cycloid gears and harmonic drives, are typically used for industrial robots in order to meet high precision requirements. Simplified models taking into account linear stiffness and damping between the motor and the driven link are used in \cite{Spong1987,DeLuca1998,Wang1992,Albu-Schaffer2007,Mesmer2020}. An overview of further simplified models can be found in \cite{DeLuca2008}. 
Nonlinear modeling approaches account for torsional compliance by considering hysteresis effects \cite{Ruderman2009,Cordes2017}. Hysteresis behavior results from the structural damping of transmission elements and their piecewise elasto-plastic properties \cite{Ruderman2009}. It was shown that hysteresis effects can significantly contribute to a low path accuracy \cite{Cordes2017}. The authors point out, that it mostly impacts the positional error of the base joint due to the large lever arm. Moreover, hysteresis behavior, which has a bidirectional behaviour, leads to alternating stresses in the gears \cite{Bruning2016}. The non-linear characteristics of lost-motion and backlash in transmission drives have further considerable effects on lower path accuracy in machining tasks. Lost-motion is defined as the torsion angle at the midpoint of hysteresis curve, where not all tooth-flanks of the gearbox are in full contact, whereas backlash is defined as the angle difference in the output shaft at zero output torque, where gear teeth are not in contact \cite{Tran2016LostTolerances,Huynh2018}. The literature on modeling lost-motion effects is scarce. The impact of backlash caused by transmission drives was recognized by \cite{Kircanski1997} in his experimental study of harmonic drive, showing that torque transmission has a nonlinear characteristics and the input torque can not be entirely transmitted to the driven link. The impact of backlash is also investigated and modeled in \cite{Freising2014,Yang2015}. 
The authors in \cite{Huynh2018} show a considerable improvement in milling accuracy of aluminium by taking into account backlash effects and all joint flexibilities. Moreover, it was shown in \cite{Yang2016} that backlash leads to the accumulation of positioning errors while joints change their rotation direction, leading to torque oscillations and thus to earlier gear system failure.  

Enhancing trajectory tracking accuracy requires not only considering the most relevant dynamic effects but also designing a proper control law. Chattering effects, which significantly influence the surface quality in a milling process, can only be eliminated using a flexible model-based controller \cite{Kim2019}. For the flexible model-based controller, the trajectory should be continuously differentiable up to the $4^{th}$ order, i.e. up to jerk derivative, whereas a $2^{nd}$ order trajectory is sufficient for rigid model-based controllers. A continuously differentiable trajectory can be computed e.g.\ based on the dynamic model or estimated by numerical differentiation, which is error prone due to high sampling frequencies, measurement noise or model uncertainties. 
The authors in \cite{Yin2018} propose a robust adaptive control method for trajectory tracking and an online parameter estimation of a 6 degrees of freedom (DoF) industrial robot. The proposed controller differs from other controllers in the literature, as it is designed in the task space of the robot's end effector in order to achieve better trajectory tracking accuracy compared to controllers designed in joint space. The method shows a significant reduction of trajectory tracking error compared to a conventional PD controller. Similarly, the authors in \cite{Mesmer2020} show an improvement in dynamic path accuracy of a robot manipulator in a machining process, proposing a controller built on an independent joint control. A damping control algorithm is designed and validated experimentally, based on a velocity feedback using SE. The approach is extended by a state estimation. With the proposed controller, it is possible to alter the stiffness and damping of the controller system systematically with help of two proportional gains. The authors in \cite{Zhang2020AccurateRobot} show that modeling nonlinear effects, such as friction, and subsequent derivation of a model-based controller, can lead to a significant improvement in position and velocity accuracy of an industrial robot.

Providing a model which takes nonlinear effects into consideration is critical to trajectory tracking accuracy for manipulator systems. The aim of this paper is to improve trajectory tracking performance of an industrial robot by combining the model-based and sensor-based approaches. A novel nonlinear but continuous differentiable modeling approach for friction, backlash and lost-motion is introduced. This allows a derivation of a flatness based feed forward control law. The feed forward control law, which is mostly preferred as it shows more robustness against sampling rate, sensor noise and parameter uncertainty \cite{Kim2019} reduces positional errors up to 60 \% \cite{Wang2009}. In addition, model errors are compensated by applying a feedback law using SE. We propose an enhanced motor-side velocity controller, which takes flexible joint model into account.

The paper is structured as follows. At first, a nonlinear dynamic model is introduced, followed by Section ~\ref{sec:control}, explaining the advanced control design with feed forward and feedback controller. Further, simulation results are presented and discussed in Section ~\ref{sec:simulation}, as well as experimental results in Section ~\ref{sec:experiments}. The paper is finalized by concluding aspects for future research in Section~\ref{sec:conculsion}. 

%% file: sec_modeling.tex
\section{Modeling}
\label{sec:model}
\begin{figure}
\vspace{\baselineskip}
    \centering
	\begin{overpic}[width=0.5\textwidth]{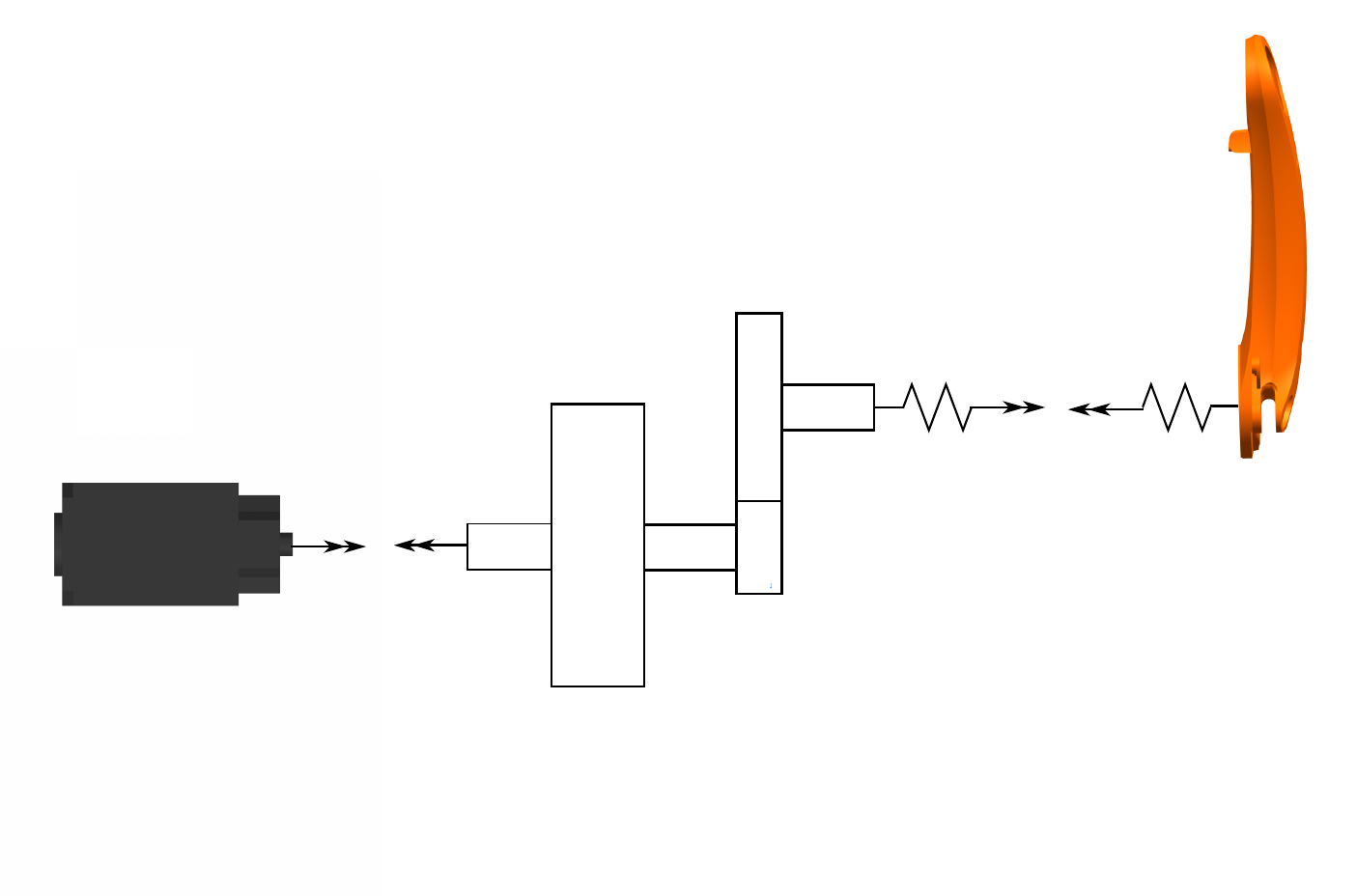}
	\put(42,24){$J_{i}$}
	\put(22,29){$\tau_{M,i}$}
	\put(29,22){$\tau_{M,i}$}
	\put(59,28){$u_i$}
	\put(73,32){$\tau_{E,i}$}
	\put(79,39){$\tau_{E,i}$}
	\end{overpic}
	\vspace{-40pt}
	\caption{Representation of a flexible robot joint.}
	\label{fig:two-mass-oscillator}
\end{figure}
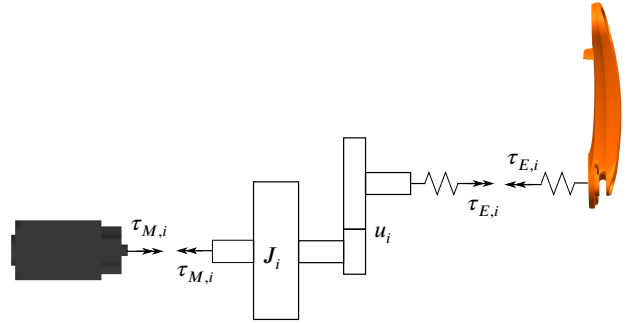
For modeling a joint flexibility, each motor is considered as a rigid body connected to a driven link via a transmission device and a rotational spring, illustrated in Fig.~\ref{fig:two-mass-oscillator}. Such a system can be modeled as a two mass oscillator leading to two generalized coordinates for each single joint. The generalized coordinates are composed of the motor positions $\boldsymbol{\theta} \in \mathbb{R}^N$ and the joint positions  $\mathbf{q} \in \mathbb{R}^N$, where $N$ stands for the number of DoF. That leads to an $2N$ DoF system of a robot, meaning a $12$ DoF system for robot kinematics with $6$ joints. The dynamical model of the considered robot manipulator is derived by means of Lagrange's equations of the second kind. The transmission ratios are assumed to be high enough, so that inertial couplings in the acceleration between the motors and the links can be neglected \cite{DeLuca2016}. The governing equations are 
\begin{align} 
\label{eq:robotdynamics}
\mathbf{J}\boldsymbol{\ddot{\theta}} + \mathbf{U}^{-1}\boldsymbol{\tau_E}(\boldsymbol{\theta},\mathbf{q}) &= \boldsymbol{\tau_M}, \\
\mathbf{M}(\mathbf{q})\ddot{\mathbf{q}}+\mathbf{C}(\mathbf{q},\dot{\mathbf{q}})\dot{\mathbf{q}} + \mathbf{g}(\mathbf{q})   + \boldsymbol{\tau_F}(\dot{\mathbf{q}}) &= \boldsymbol{\tau_E}(\boldsymbol{\theta},\mathbf{q}), 
\label{eq:robotdynamics2}
\end{align}
where \eqref{eq:robotdynamics} describes the dynamical model of the flexible joints and \eqref{eq:robotdynamics2} represents the dynamics of the links. \\
The two dynamical systems are coupled by the generalized elastic torque vector $\boldsymbol{\tau_E}(\boldsymbol{\theta},\mathbf{q}) \in \mathbb{R}^{N}$. $\mathbf{J} = \text{diag}(J_1,\dots,J_N) $ is the inertia matrix of the rotors, the vector of generalized friction torques induced by dissipative forces is represented by the vector $\boldsymbol{\tau_F}(\dot{\mathbf{q}}) \in \mathbb{R}^{N}$, the input torque transmitted from motors to the driven links is summarized in the generalized vector $\boldsymbol{\tau_M} \in \mathbb{R}^{N}$, the transmission ratio for each joint is summarized in the matrix $\mathbf{U} = \text{diag}(u_1,\dots,u_N)$. From \eqref{eq:robotdynamics2}, the pose dependent inertia matrix of the links is denoted by $\mathbf{M} \in \mathbb{R}^{N \times N}$, $\mathbf{C}(\mathbf{q},\dot{\mathbf{q}}) \in \mathbb{R}^{N \times N}$ represents the matrix, which coefficients encompass the centrifugal (proportional to $\dot{q}^2_i$) and Coriolis (proportional to $\dot{q}_i\dot{q}_j,i\neq j$) forces and $\mathbf{g}(\mathbf{q}) \in \mathbb{R}^{N}$ is the vector of gravitational torques. For the sake of simplicity, a single joint will be considered in the following, omitting the indices of the variables indicating a specific joint e.g. $v_i$ is the $i$-th element of the vector $\mathbf{v} \in \mathbb{R}^{N}$, simplified to $v_i = v$. 

\subsection{Friction Model}
The gearbox friction is modeled as Coulomb and viscous friction. The temperature dependency of the friction is neglected in this paper. With the viscous friction constant $f_{v}$ and the Coulomb friction torque $f_{c}$ we get

\begin{align}
\tau_{F} =
\begin{dcases}
0, &  \dot{{q}} = 0 \\
f_{v} \, \dot{{q}} + f_{c} \, \sign(\dot{{q}} ), & \text{otherwise}.
\end{dcases}
\label{eq:mod_friction}
\end{align}

For the implementation of a flatness based controller, a continuously differentiable function is needed. Therefore, an approximation of the friction is introduced, $\tilde{\tau}_{F}\approx  \tau_{F}$, using an exponential function and a friction smoothness factor $s_{F}$

\begin{equation}
\tilde{\tau}_{F}(\dot{q}) 
= f_{v} \, \dot{q} +\frac{2 \, f_{c}}{1+e^{-s_{F} \, \dot{q}}} - f_{c}.
\label{eq:mod_friction_app}
\end{equation}

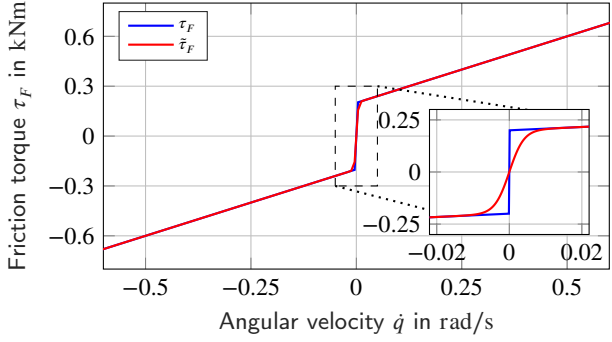
\begin{figure}
	\centering
	\setlength\fheight{3.5cm}
	\setlength\fwidth{7cm}
	\input{figs/friction_v3}
	\caption{Coulomb and viscous friction model with discontinuous and continuous approximation, exemplary for joint~$1$. Model parameters are given in Tab.~\ref{tab:alltableparameters}.}
    \label{fig:friction}
\end{figure}

The functions \eqref{eq:mod_friction} and \eqref{eq:mod_friction_app} are illustrated in Fig.~\ref{fig:friction}.

\subsection{Stiffness Model}
\begin{figure}
	\centering
	\setlength\fheight{3.5cm}
	\setlength\fwidth{7cm}
	\input{figs/stiffness_v2}
	\vspace{-10pt}
	\caption{Discontinuous and continuous approximation for stiffness model with the effects of backlash, lost-motion and linear elasticity, exemplary for joint~$1$. Model parameters are given in Tab.~\ref{tab:alltableparameters}.}
    \label{fig:stiffness_app}
\end{figure}
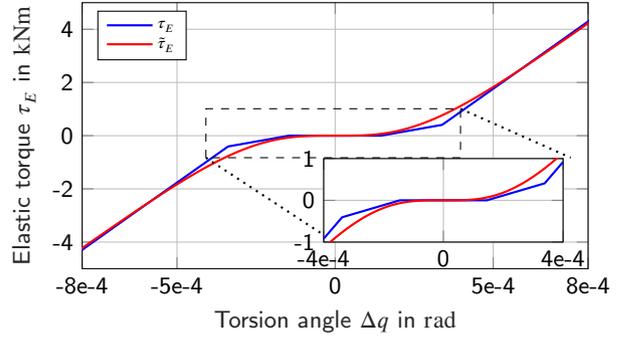

The stiffness model considers backlash, lost-motion, and linear elasticity. Lost-motion describes an effect in between backlash and linear elasticity, where not all tooth flanks are in full contact. During backlash the elastic torque vanishes, i.e. $\tau_E = 0$. In the lost-motion range, the stiffness is modeled as linear with a smaller slope and an offset. In the torsional rigidity range, the stiffness is modeled as linear with an offset, as shown in Fig.~\ref{fig:stiffness_app}. 
We define the stiffness coefficients $c_{LM}$ and $c_{TR}$, for lost-motion and torsional rigidity respectively. We further define the angular ranges $\phi_{B*}$ and $\phi_{LM}$, for backlash and lost-motion respectively. The torsional ridigity is defined as a measurement between $50 \%$ and $100 \%$ of the nominal torque. Lost-motion effects occur directly after the backlash and are measured between $\pm 3 \%$ of the nominal torque \cite{Tran2016LostTolerances}. Therefore, we introduce an effective backlash angle $\phi_{B} = \phi_{B*} + \phi_{LM}$ which ensures a correct modeling of the lost-motion and torsional rigidity range. With the link torsion angle and its sign function
\begin{align}
\Delta{q} &= \theta/u - {q},
\label{eq:torsion_diff_angle} \\
\sigma &= \sign(\Delta{q}) \nonumber
\end{align} 
and the elastic torque offset $\tau_{E,0} = c_{LM} \, \phi_{LM}$, the elastic torque can be obtained
\begin{align}
\tau_{E} &=\begin{dcases}
0, & |\Delta{q}| \leq \phi_{B*} \\
c_{LM} \left( \Delta{q} - \frac{\phi_{LM} \, \sigma}{2} \right), & \phi_{B*} < |\Delta{q}| \leq \phi_{B} \\
c_{TR} \, \Delta q + \tau_{E,0} \, \sigma, & \text{otherwise}.  
\end{dcases}
\label{eq:mod_stiff}
\end{align}
Although \eqref{eq:mod_stiff} describes the nonlinear stiffness precisely, it is difficult to use it in parameter estimation or advanced control strategies, such as Model Predictive Control (MPC), since the function is not continuously differentiable. Problems with \eqref{eq:mod_stiff} potentially arise with all derivative based algorithms. In particular, many Automatic Differentiation (AD) tools, such as implemented in the \textit{Symbolic Math Toolbox} from \textit{MATLAB} \cite{TheMathWorks2019SymbolicToolbox} require continuously differentiable equations. As a main contribution of this paper, we present a continuously differentiable elastic torque function, which origins from time-domain considerations as explained in Appendix~\ref{app:stiffness} and is applied in the following. To derive a differentiable approximation $ \tilde{\tau}_{E} \approx \tau_{E}$ of \eqref{eq:mod_stiff}, we define
\begin{align}
\tilde{\tau}_{E,+}( \Delta{q} )
= &\,c_{TR} \, \Delta{q} - c_{TR} \, \phi_B \nonumber \\
+ &\,c_{TR} \, \phi_B \, e^{-(3 \, \Delta{q}/\phi_B)} \label{eq:tau_e_plus_3rd} \\
+ &\,2 \, c_{TR} \, \Delta{q} \, e^{-(3 \, \Delta{q}/\phi_B)} \nonumber \\
+ &\,3 \, c_{TR} \, /(2 \, \phi_B) \, (\Delta{q})^2 \, e^{-(3 \, \Delta{q}/\phi_B)} \nonumber
\end{align}
for all $\Delta{q} \ge 0$ and
\begin{equation}
\tilde{\tau}_{E,-}( \Delta{q} ) = -\tilde{\tau}_{E,+}( -\Delta{q} ),
\label{eq:tau_e_minus_3rd}
\end{equation}
 for all $\Delta{q} < 0$. A complete continuously differentiable elastic torque function is given by

\begin{equation}
\tilde{\tau}_{E} = \tanh( s_{E1} \, \Delta{q} ) \, \tilde{\tau}_{E,+}\left( \Delta{q} \cdot \tanh( s_{E1} \, \Delta{q} ) \right)
\label{eq:tau_e_cont_3rd}
\end{equation}
for all $\Delta{q} \in \mathbb{R}$ with the hyperbolic tangent $\tanh( \cdot )$. It is essential for the tangent slope factor $s_{E1} $ that $s_{E1} \gg 3 / \phi_B $ holds. Note that the slope factor is not upper-bounded, except for limits due to numerical considerations. The continuously differentiable stiffness curve is depicted in Fig.~\ref{fig:stiffness_app}.

For many model applications, such as MPC or parameter identification, a continuously differentiable stiffness function is beneficial. However, a flatness based control is an exception, where both, \eqref{eq:tau_e_cont_3rd} and the inverse stiffness function \eqref{eq:vor_steifigkeit} can be employed. Although the inverse stiffness function is neither common nor applicable in general, it is advantageous in the case of flatness based control to use the following form
\begin{align}
\Delta q = 
\begin{dcases}
0, & \tau_{E} = 0 \\
\frac{\tau_{E}}{c_{LM}} + \frac{\phi_{B*}}{2} \, \sign(\tau_{E}), & 0 <  |\tau_{E}|  \leq \tau_{E,0}  \\
\frac{\tau_{E}}{c_{TR}} + \phi_{B}\, \sign(\tau_{E}), &  \tau_{E,0} < |\tau_{E}|
\end{dcases} \label{eq:vor_steifigkeit}
\end{align}
since the inverse stiffness function reduces the number of required exponential functions from eight in \eqref{eq:tau_e_cont_3rd} to one in \eqref{eq:mod_stiff_app}. For \eqref{eq:vor_steifigkeit} we can apply the same method as for the nonlinear friction, and we obtain the nonlinear, inverse, continuously differentiable stiffness function

\begin{align}
\Delta{q}
= \frac{\tau_{E}}{c_{TR}} + \frac{2 \, \phi_{B}}{1+e^{-s_{E2} \, \tau_{E}}} - \phi_B \label{eq:mod_stiff_app}
\end{align}
for all $\Delta{q} \in \mathbb{R}$ with the elastic smoothing factor $s_{E2}$.






%% file: figs/friction_v3.tex
%
%
%
\begin{tikzpicture}
\begin{axis}[%
width=0.951\fwidth,
height=\fheight,
at={(0\fwidth,0\fheight)},
scale only axis,
xmin=-0.6,
xmax=0.6,
xtick={0.5,0.25,...,-0.5},
xlabel style={font=\color{white!15!black}},
xlabel={Angular velocity $\dot{q}$ in $\mathrm{rad}/\mathrm{s}$},
ymin=-0.8,
ymax=0.8,
ytick={0.6, 0.3, ..., -0.6},
ylabel style={font=\color{white!15!black}},
ylabel={Friction torque $\tau_{F}$ in $\mathrm{kNm}$},
axis background/.style={fill=white},
xmajorgrids,
ymajorgrids,
legend style={at={(0.03,0.97)}, anchor=north west, legend cell align=left, align=left, draw=white!15!black}
]
\addplot [color=blue]
  table[row sep=crcr, x expr=\thisrow{X}, y expr=\thisrow{Y}*0.001]{
  X Y \\
    -0.8          -840 \\
    -0.7919598     -833.5678 \\
    -0.7839196     -827.1357 \\
    -0.7758794     -820.7035 \\
    -0.7678392     -814.2714 \\
    -0.759799     -807.8392 \\
    -0.7517588      -801.407 \\
    -0.7437186     -794.9749 \\
    -0.7356784     -788.5427 \\
    -0.7276382     -782.1106 \\
    -0.719598     -775.6784 \\
    -0.7115578     -769.2462 \\
    -0.7035176     -762.8141 \\
    -0.6954774     -756.3819 \\
    -0.6874372     -749.9497 \\
    -0.679397     -743.5176 \\
    -0.6713568     -737.0854 \\
    -0.6633166     -730.6533 \\
    -0.6552764     -724.2211 \\
    -0.6472362     -717.7889 \\
    -0.639196     -711.3568 \\
    -0.6311558     -704.9246 \\
    -0.6231156     -698.4925 \\
    -0.6150754     -692.0603 \\
    -0.6070352     -685.6281 \\
    -0.598995      -679.196 \\
    -0.5909548     -672.7638 \\
    -0.5829146     -666.3317 \\
    -0.5748744     -659.8995 \\
    -0.5668342     -653.4673 \\
    -0.558794     -647.0352 \\
    -0.5507538      -640.603 \\
    -0.5427136     -634.1709 \\
    -0.5346734     -627.7387 \\
    -0.5266332     -621.3065 \\
    -0.518593     -614.8744 \\
    -0.5105528     -608.4422 \\
    -0.5025126     -602.0101 \\
    -0.4944724     -595.5779 \\
    -0.4864322     -589.1457 \\
    -0.478392     -582.7136 \\
    -0.4703518     -576.2814 \\
    -0.4623116     -569.8492 \\
    -0.4542714     -563.4171 \\
    -0.4462312     -556.9849 \\
    -0.438191     -550.5528 \\
    -0.4301508     -544.1206 \\
    -0.4221106     -537.6884 \\
    -0.4140704     -531.2563 \\
    -0.4060302     -524.8241 \\
    -0.3979899      -518.392 \\
    -0.3899497     -511.9598 \\
    -0.3819095     -505.5276 \\
    -0.3738693     -499.0955 \\
    -0.3658291     -492.6633 \\
    -0.3577889     -486.2312 \\
    -0.3497487      -479.799 \\
    -0.3417085     -473.3668 \\
    -0.3336683     -466.9347 \\
    -0.3256281     -460.5025 \\
    -0.3175879     -454.0704 \\
    -0.3095477     -447.6382 \\
    -0.3015075      -441.206 \\
    -0.2934673     -434.7739 \\
    -0.2854271     -428.3417 \\
    -0.2773869     -421.9095 \\
    -0.2693467     -415.4774 \\
    -0.2613065     -409.0452 \\
    -0.2532663     -402.6131 \\
    -0.2452261     -396.1809 \\
    -0.2371859     -389.7487 \\
    -0.2291457     -383.3166 \\
    -0.2211055     -376.8844 \\
    -0.2130653     -370.4523 \\
    -0.2050251     -364.0201 \\
    -0.1969849     -357.5879 \\
    -0.1889447     -351.1558 \\
    -0.1809045     -344.7236 \\
    -0.1728643     -338.2915 \\
    -0.1648241     -331.8593 \\
    -0.1567839     -325.4271 \\
    -0.1487437      -318.995 \\
    -0.1407035     -312.5628 \\
    -0.1326633     -306.1307 \\
    -0.1246231     -299.6985 \\
    -0.1165829     -293.2663 \\
    -0.1085427     -286.8342 \\
    -0.1005025      -280.402 \\
    -0.09246231     -273.9698 \\
    -0.08442211     -267.5377 \\
    -0.07638191     -261.1055 \\
    -0.06834171     -254.6734 \\
    -0.06030151     -248.2412 \\
    -0.05226131      -241.809 \\
    -0.04422111     -235.3769 \\
    -0.0361809     -228.9447 \\
    -0.0281407     -222.5126 \\
    -0.0201005     -216.0804 \\
    -0.0120603     -209.6482 \\
    -0.004020101     -203.2161 \\
    0.004020101      203.2161 \\
    0.0120603      209.6482 \\
    0.0201005      216.0804 \\
    0.0281407      222.5126 \\
    0.0361809      228.9447 \\
    0.04422111      235.3769 \\
    0.05226131       241.809 \\
    0.06030151      248.2412 \\
    0.06834171      254.6734 \\
    0.07638191      261.1055 \\
    0.08442211      267.5377 \\
    0.09246231      273.9698 \\
    0.1005025       280.402 \\
    0.1085427      286.8342 \\
    0.1165829      293.2663 \\
    0.1246231      299.6985 \\
    0.1326633      306.1307 \\
    0.1407035      312.5628 \\
    0.1487437       318.995 \\
    0.1567839      325.4271 \\
    0.1648241      331.8593 \\
    0.1728643      338.2915 \\
    0.1809045      344.7236 \\
    0.1889447      351.1558 \\
    0.1969849      357.5879 \\
    0.2050251      364.0201 \\
    0.2130653      370.4523 \\
    0.2211055      376.8844 \\
    0.2291457      383.3166 \\
    0.2371859      389.7487 \\
    0.2452261      396.1809 \\
    0.2532663      402.6131 \\
    0.2613065      409.0452 \\
    0.2693467      415.4774 \\
    0.2773869      421.9095 \\
    0.2854271      428.3417 \\
    0.2934673      434.7739 \\
    0.3015075       441.206 \\
    0.3095477      447.6382 \\
    0.3175879      454.0704 \\
    0.3256281      460.5025 \\
    0.3336683      466.9347 \\
    0.3417085      473.3668 \\
    0.3497487       479.799 \\
    0.3577889      486.2312 \\
    0.3658291      492.6633 \\
    0.3738693      499.0955 \\
    0.3819095      505.5276 \\
    0.3899497      511.9598 \\
    0.3979899       518.392 \\
    0.4060302      524.8241 \\
    0.4140704      531.2563 \\
    0.4221106      537.6884 \\
    0.4301508      544.1206 \\
    0.438191      550.5528 \\
    0.4462312      556.9849 \\
    0.4542714      563.4171 \\
    0.4623116      569.8492 \\
    0.4703518      576.2814 \\
    0.478392      582.7136 \\
    0.4864322      589.1457 \\
    0.4944724      595.5779 \\
    0.5025126      602.0101 \\
    0.5105528      608.4422 \\
    0.518593      614.8744 \\
    0.5266332      621.3065 \\
    0.5346734      627.7387 \\
    0.5427136      634.1709 \\
    0.5507538       640.603 \\
    0.558794      647.0352 \\
    0.5668342      653.4673 \\
    0.5748744      659.8995 \\
    0.5829146      666.3317 \\
    0.5909548      672.7638 \\
    0.598995       679.196 \\
    0.6070352      685.6281 \\
    0.6150754      692.0603 \\
    0.6231156      698.4925 \\
    0.6311558      704.9246 \\
    0.639196      711.3568 \\
    0.6472362      717.7889 \\
    0.6552764      724.2211 \\
    0.6633166      730.6533 \\
    0.6713568      737.0854 \\
    0.679397      743.5176 \\
    0.6874372      749.9497 \\
    0.6954774      756.3819 \\
    0.7035176      762.8141 \\
    0.7115578      769.2462 \\
    0.719598      775.6784 \\
    0.7276382      782.1106 \\
    0.7356784      788.5427 \\
    0.7437186      794.9749 \\
    0.7517588       801.407 \\
    0.759799      807.8392 \\
    0.7678392      814.2714 \\
    0.7758794      820.7035 \\
    0.7839196      827.1357 \\
    0.7919598      833.5678 \\
    0.8           840 \\
};
\addlegendentry{$ \tau_{F} $}
\addplot [color=red]
  table[row sep=crcr, x expr=\thisrow{X}, y expr=\thisrow{Y}*0.001]{
  X Y \\
    -0.8          -840 \\
    -0.7919598     -833.5678 \\
    -0.7839196     -827.1357 \\
    -0.7758794     -820.7035 \\
    -0.7678392     -814.2714 \\
    -0.759799     -807.8392 \\
    -0.7517588      -801.407 \\
    -0.7437186     -794.9749 \\
    -0.7356784     -788.5427 \\
    -0.7276382     -782.1106 \\
    -0.719598     -775.6784 \\
    -0.7115578     -769.2462 \\
    -0.7035176     -762.8141 \\
    -0.6954774     -756.3819 \\
    -0.6874372     -749.9497 \\
    -0.679397     -743.5176 \\
    -0.6713568     -737.0854 \\
    -0.6633166     -730.6533 \\
    -0.6552764     -724.2211 \\
    -0.6472362     -717.7889 \\
    -0.639196     -711.3568 \\
    -0.6311558     -704.9246 \\
    -0.6231156     -698.4925 \\
    -0.6150754     -692.0603 \\
    -0.6070352     -685.6281 \\
    -0.598995      -679.196 \\
    -0.5909548     -672.7638 \\
    -0.5829146     -666.3317 \\
    -0.5748744     -659.8995 \\
    -0.5668342     -653.4673 \\
    -0.558794     -647.0352 \\
    -0.5507538      -640.603 \\
    -0.5427136     -634.1709 \\
    -0.5346734     -627.7387 \\
    -0.5266332     -621.3065 \\
    -0.518593     -614.8744 \\
    -0.5105528     -608.4422 \\
    -0.5025126     -602.0101 \\
    -0.4944724     -595.5779 \\
    -0.4864322     -589.1457 \\
    -0.478392     -582.7136 \\
    -0.4703518     -576.2814 \\
    -0.4623116     -569.8492 \\
    -0.4542714     -563.4171 \\
    -0.4462312     -556.9849 \\
    -0.438191     -550.5528 \\
    -0.4301508     -544.1206 \\
    -0.4221106     -537.6884 \\
    -0.4140704     -531.2563 \\
    -0.4060302     -524.8241 \\
    -0.3979899      -518.392 \\
    -0.3899497     -511.9598 \\
    -0.3819095     -505.5276 \\
    -0.3738693     -499.0955 \\
    -0.3658291     -492.6633 \\
    -0.3577889     -486.2312 \\
    -0.3497487      -479.799 \\
    -0.3417085     -473.3668 \\
    -0.3336683     -466.9347 \\
    -0.3256281     -460.5025 \\
    -0.3175879     -454.0704 \\
    -0.3095477     -447.6382 \\
    -0.3015075      -441.206 \\
    -0.2934673     -434.7739 \\
    -0.2854271     -428.3417 \\
    -0.2773869     -421.9095 \\
    -0.2693467     -415.4774 \\
    -0.2613065     -409.0452 \\
    -0.2532663     -402.6131 \\
    -0.2452261     -396.1809 \\
    -0.2371859     -389.7487 \\
    -0.2291457     -383.3166 \\
    -0.2211055     -376.8844 \\
    -0.2130653     -370.4523 \\
    -0.2050251     -364.0201 \\
    -0.1969849     -357.5879 \\
    -0.1889447     -351.1558 \\
    -0.1809045     -344.7236 \\
    -0.1728643     -338.2915 \\
    -0.1648241     -331.8593 \\
    -0.1567839     -325.4271 \\
    -0.1487437      -318.995 \\
    -0.1407035     -312.5628 \\
    -0.1326633     -306.1307 \\
    -0.1246231     -299.6985 \\
    -0.1165829     -293.2663 \\
    -0.1085427     -286.8342 \\
    -0.1005025      -280.402 \\
    -0.09246231     -273.9698 \\
    -0.08442211     -267.5377 \\
    -0.07638191     -261.1055 \\
    -0.06834171     -254.6734 \\
    -0.06030151     -248.2412 \\
    -0.05226131      -241.809 \\
    -0.04422111     -235.3769 \\
    -0.0361809     -228.9447 \\
    -0.0281407     -222.5123 \\
    -0.0201005     -216.0631 \\
    -0.0120603     -208.6885 \\
    -0.004020101     -155.9554 \\
    0.004020101      155.9554 \\
    0.0120603      208.6885 \\
    0.0201005      216.0631 \\
    0.0281407      222.5123 \\
    0.0361809      228.9447 \\
    0.04422111      235.3769 \\
    0.05226131       241.809 \\
    0.06030151      248.2412 \\
    0.06834171      254.6734 \\
    0.07638191      261.1055 \\
    0.08442211      267.5377 \\
    0.09246231      273.9698 \\
    0.1005025       280.402 \\
    0.1085427      286.8342 \\
    0.1165829      293.2663 \\
    0.1246231      299.6985 \\
    0.1326633      306.1307 \\
    0.1407035      312.5628 \\
    0.1487437       318.995 \\
    0.1567839      325.4271 \\
    0.1648241      331.8593 \\
    0.1728643      338.2915 \\
    0.1809045      344.7236 \\
    0.1889447      351.1558 \\
    0.1969849      357.5879 \\
    0.2050251      364.0201 \\
    0.2130653      370.4523 \\
    0.2211055      376.8844 \\
    0.2291457      383.3166 \\
    0.2371859      389.7487 \\
    0.2452261      396.1809 \\
    0.2532663      402.6131 \\
    0.2613065      409.0452 \\
    0.2693467      415.4774 \\
    0.2773869      421.9095 \\
    0.2854271      428.3417 \\
    0.2934673      434.7739 \\
    0.3015075       441.206 \\
    0.3095477      447.6382 \\
    0.3175879      454.0704 \\
    0.3256281      460.5025 \\
    0.3336683      466.9347 \\
    0.3417085      473.3668 \\
    0.3497487       479.799 \\
    0.3577889      486.2312 \\
    0.3658291      492.6633 \\
    0.3738693      499.0955 \\
    0.3819095      505.5276 \\
    0.3899497      511.9598 \\
    0.3979899       518.392 \\
    0.4060302      524.8241 \\
    0.4140704      531.2563 \\
    0.4221106      537.6884 \\
    0.4301508      544.1206 \\
    0.438191      550.5528 \\
    0.4462312      556.9849 \\
    0.4542714      563.4171 \\
    0.4623116      569.8492 \\
    0.4703518      576.2814 \\
    0.478392      582.7136 \\
    0.4864322      589.1457 \\
    0.4944724      595.5779 \\
    0.5025126      602.0101 \\
    0.5105528      608.4422 \\
    0.518593      614.8744 \\
    0.5266332      621.3065 \\
    0.5346734      627.7387 \\
    0.5427136      634.1709 \\
    0.5507538       640.603 \\
    0.558794      647.0352 \\
    0.5668342      653.4673 \\
    0.5748744      659.8995 \\
    0.5829146      666.3317 \\
    0.5909548      672.7638 \\
    0.598995       679.196 \\
    0.6070352      685.6281 \\
    0.6150754      692.0603 \\
    0.6231156      698.4925 \\
    0.6311558      704.9246 \\
    0.639196      711.3568 \\
    0.6472362      717.7889 \\
    0.6552764      724.2211 \\
    0.6633166      730.6533 \\
    0.6713568      737.0854 \\
    0.679397      743.5176 \\
    0.6874372      749.9497 \\
    0.6954774      756.3819 \\
    0.7035176      762.8141 \\
    0.7115578      769.2462 \\
    0.719598      775.6784 \\
    0.7276382      782.1106 \\
    0.7356784      788.5427 \\
    0.7437186      794.9749 \\
    0.7517588       801.407 \\
    0.759799      807.8392 \\
    0.7678392      814.2714 \\
    0.7758794      820.7035 \\
    0.7839196      827.1357 \\
    0.7919598      833.5678 \\
    0.8           840 \\
};
\addlegendentry{$\tilde{\tau}_{F} $}
\addplot [color=black, dotted, forget plot]
  table[row sep=crcr, x expr=\thisrow{X}, y expr=\thisrow{Y}]{
  X Y \\
-0.05 -0.3\\
0.18 -0.45 \\
}; 
\addplot [color=black, dotted, forget plot]
  table[row sep=crcr, x expr=\thisrow{X}, y expr=\thisrow{Y}]{
  X Y \\
0.05 0.3 \\
0.55	0.1 \\
}; 
\draw[dashed, draw=black] (axis cs: -0.05, -0.3) rectangle (axis cs: 0.05, 0.3);
\end{axis}
\begin{axis}[%
width=0.30\fwidth,
height=0.47\fheight,
at={(0.613\fwidth,0.13\fheight)},
scale only axis,
xmin=-0.022,
xmax=0.022,
ymin= -0.3,
ymax= 0.3,
xtick= {-0.02,0, 0.02},
ytick= {-0.25, 0, 0.25},
axis background/.style={fill=white},
xmajorgrids,
ymajorgrids,
legend style={legend cell align=left, align=left, draw=white!15!black}
]
\addplot [color=blue]
  table[row sep=crcr, x expr=\thisrow{X}, y expr=\thisrow{Y}*0.001]{
  X Y \\
    -0.022        -217.6 \\
    -0.02177889     -217.4231 \\
    -0.02155779     -217.2462 \\
    -0.02133668     -217.0693 \\
    -0.02111558     -216.8925 \\
    -0.02089447     -216.7156 \\
    -0.02067337     -216.5387 \\
    -0.02045226     -216.3618 \\
    -0.02023116     -216.1849 \\
    -0.02001005      -216.008 \\
    -0.01978894     -215.8312 \\
    -0.01956784     -215.6543 \\
    -0.01934673     -215.4774 \\
    -0.01912563     -215.3005 \\
    -0.01890452     -215.1236 \\
    -0.01868342     -214.9467 \\
    -0.01846231     -214.7698 \\
    -0.01824121      -214.593 \\
    -0.0180201     -214.4161 \\
    -0.01779899     -214.2392 \\
    -0.01757789     -214.0623 \\
    -0.01735678     -213.8854 \\
    -0.01713568     -213.7085 \\
    -0.01691457     -213.5317 \\
    -0.01669347     -213.3548 \\
    -0.01647236     -213.1779 \\
    -0.01625126      -213.001 \\
    -0.01603015     -212.8241 \\
    -0.01580905     -212.6472 \\
    -0.01558794     -212.4704 \\
    -0.01536683     -212.2935 \\
    -0.01514573     -212.1166 \\
    -0.01492462     -211.9397 \\
    -0.01470352     -211.7628 \\
    -0.01448241     -211.5859 \\
    -0.01426131      -211.409 \\
    -0.0140402     -211.2322 \\
    -0.0138191     -211.0553 \\
    -0.01359799     -210.8784 \\
    -0.01337688     -210.7015 \\
    -0.01315578     -210.5246 \\
    -0.01293467     -210.3477 \\
    -0.01271357     -210.1709 \\
    -0.01249246      -209.994 \\
    -0.01227136     -209.8171 \\
    -0.01205025     -209.6402 \\
    -0.01182915     -209.4633 \\
    -0.01160804     -209.2864 \\
    -0.01138693     -209.1095 \\
    -0.01116583     -208.9327 \\
    -0.01094472     -208.7558 \\
    -0.01072362     -208.5789 \\
    -0.01050251      -208.402 \\
    -0.01028141     -208.2251 \\
    -0.0100603     -208.0482 \\
    -0.009839196     -207.8714 \\
    -0.00961809     -207.6945 \\
    -0.009396985     -207.5176 \\
    -0.009175879     -207.3407 \\
    -0.008954774     -207.1638 \\
    -0.008733668     -206.9869 \\
    -0.008512563     -206.8101 \\
    -0.008291457     -206.6332 \\
    -0.008070352     -206.4563 \\
    -0.007849246     -206.2794 \\
    -0.007628141     -206.1025 \\
    -0.007407035     -205.9256 \\
    -0.00718593     -205.7487 \\
    -0.006964824     -205.5719 \\
    -0.006743719      -205.395 \\
    -0.006522613     -205.2181 \\
    -0.006301508     -205.0412 \\
    -0.006080402     -204.8643 \\
    -0.005859296     -204.6874 \\
    -0.005638191     -204.5106 \\
    -0.005417085     -204.3337 \\
    -0.00519598     -204.1568 \\
    -0.004974874     -203.9799 \\
    -0.004753769      -203.803 \\
    -0.004532663     -203.6261 \\
    -0.004311558     -203.4492 \\
    -0.004090452     -203.2724 \\
    -0.003869347     -203.0955 \\
    -0.003648241     -202.9186 \\
    -0.003427136     -202.7417 \\
    -0.00320603     -202.5648 \\
    -0.002984925     -202.3879 \\
    -0.002763819     -202.2111 \\
    -0.002542714     -202.0342 \\
    -0.002321608     -201.8573 \\
    -0.002100503     -201.6804 \\
    -0.001879397     -201.5035 \\
    -0.001658291     -201.3266 \\
    -0.001437186     -201.1497 \\
    -0.00121608     -200.9729 \\
    -0.0009949749      -200.796 \\
    -0.0007738693     -200.6191 \\
    -0.0005527638     -200.4422 \\
    -0.0003316583     -200.2653 \\
    -0.0001105528     -200.0884 \\
    0.0001105528      200.0884 \\
    0.0003316583      200.2653 \\
    0.0005527638      200.4422 \\
    0.0007738693      200.6191 \\
    0.0009949749       200.796 \\
    0.00121608      200.9729 \\
    0.001437186      201.1497 \\
    0.001658291      201.3266 \\
    0.001879397      201.5035 \\
    0.002100503      201.6804 \\
    0.002321608      201.8573 \\
    0.002542714      202.0342 \\
    0.002763819      202.2111 \\
    0.002984925      202.3879 \\
    0.00320603      202.5648 \\
    0.003427136      202.7417 \\
    0.003648241      202.9186 \\
    0.003869347      203.0955 \\
    0.004090452      203.2724 \\
    0.004311558      203.4492 \\
    0.004532663      203.6261 \\
    0.004753769       203.803 \\
    0.004974874      203.9799 \\
    0.00519598      204.1568 \\
    0.005417085      204.3337 \\
    0.005638191      204.5106 \\
    0.005859296      204.6874 \\
    0.006080402      204.8643 \\
    0.006301508      205.0412 \\
    0.006522613      205.2181 \\
    0.006743719       205.395 \\
    0.006964824      205.5719 \\
    0.00718593      205.7487 \\
    0.007407035      205.9256 \\
    0.007628141      206.1025 \\
    0.007849246      206.2794 \\
    0.008070352      206.4563 \\
    0.008291457      206.6332 \\
    0.008512563      206.8101 \\
    0.008733668      206.9869 \\
    0.008954774      207.1638 \\
    0.009175879      207.3407 \\
    0.009396985      207.5176 \\
    0.00961809      207.6945 \\
    0.009839196      207.8714 \\
    0.0100603      208.0482 \\
    0.01028141      208.2251 \\
    0.01050251       208.402 \\
    0.01072362      208.5789 \\
    0.01094472      208.7558 \\
    0.01116583      208.9327 \\
    0.01138693      209.1095 \\
    0.01160804      209.2864 \\
    0.01182915      209.4633 \\
    0.01205025      209.6402 \\
    0.01227136      209.8171 \\
    0.01249246       209.994 \\
    0.01271357      210.1709 \\
    0.01293467      210.3477 \\
    0.01315578      210.5246 \\
    0.01337688      210.7015 \\
    0.01359799      210.8784 \\
    0.0138191      211.0553 \\
    0.0140402      211.2322 \\
    0.01426131       211.409 \\
    0.01448241      211.5859 \\
    0.01470352      211.7628 \\
    0.01492462      211.9397 \\
    0.01514573      212.1166 \\
    0.01536683      212.2935 \\
    0.01558794      212.4704 \\
    0.01580905      212.6472 \\
    0.01603015      212.8241 \\
    0.01625126       213.001 \\
    0.01647236      213.1779 \\
    0.01669347      213.3548 \\
    0.01691457      213.5317 \\
    0.01713568      213.7085 \\
    0.01735678      213.8854 \\
    0.01757789      214.0623 \\
    0.01779899      214.2392 \\
    0.0180201      214.4161 \\
    0.01824121       214.593 \\
    0.01846231      214.7698 \\
    0.01868342      214.9467 \\
    0.01890452      215.1236 \\
    0.01912563      215.3005 \\
    0.01934673      215.4774 \\
    0.01956784      215.6543 \\
    0.01978894      215.8312 \\
    0.02001005       216.008 \\
    0.02023116      216.1849 \\
    0.02045226      216.3618 \\
    0.02067337      216.5387 \\
    0.02089447      216.7156 \\
    0.02111558      216.8925 \\
    0.02133668      217.0693 \\
    0.02155779      217.2462 \\
    0.02177889      217.4231 \\
    0.022         217.6 \\
};
\addplot [color=red]
  table[row sep=crcr, x expr=\thisrow{X}, y expr=\thisrow{Y}*0.001]{
  X Y \\
    -0.022     -217.5933 \\
    -0.02177889     -217.4157 \\
    -0.02155779     -217.2379 \\
    -0.02133668       -217.06 \\
    -0.02111558     -216.8821 \\
    -0.02089447      -216.704 \\
    -0.02067337     -216.5257 \\
    -0.02045226     -216.3473 \\
    -0.02023116     -216.1687 \\
    -0.02001005       -215.99 \\
    -0.01978894      -215.811 \\
    -0.01956784     -215.6317 \\
    -0.01934673     -215.4522 \\
    -0.01912563     -215.2724 \\
    -0.01890452     -215.0922 \\
    -0.01868342     -214.9117 \\
    -0.01846231     -214.7307 \\
    -0.01824121     -214.5492 \\
    -0.0180201     -214.3672 \\
    -0.01779899     -214.1846 \\
    -0.01757789     -214.0014 \\
    -0.01735678     -213.8173 \\
    -0.01713568     -213.6325 \\
    -0.01691457     -213.4467 \\
    -0.01669347     -213.2599 \\
    -0.01647236      -213.072 \\
    -0.01625126     -212.8827 \\
    -0.01603015      -212.692 \\
    -0.01580905     -212.4997 \\
    -0.01558794     -212.3055 \\
    -0.01536683     -212.1094 \\
    -0.01514573      -211.911 \\
    -0.01492462     -211.7101 \\
    -0.01470352     -211.5064 \\
    -0.01448241     -211.2996 \\
    -0.01426131     -211.0892 \\
    -0.0140402      -210.875 \\
    -0.0138191     -210.6564 \\
    -0.01359799     -210.4329 \\
    -0.01337688      -210.204 \\
    -0.01315578     -209.9691 \\
    -0.01293467     -209.7274 \\
    -0.01271357     -209.4781 \\
    -0.01249246     -209.2204 \\
    -0.01227136     -208.9533 \\
    -0.01205025     -208.6756 \\
    -0.01182915     -208.3863 \\
    -0.01160804     -208.0839 \\
    -0.01138693     -207.7669 \\
    -0.01116583     -207.4337 \\
    -0.01094472     -207.0823 \\
    -0.01072362     -206.7107 \\
    -0.01050251     -206.3166 \\
    -0.01028141     -205.8973 \\
    -0.0100603     -205.4501 \\
    -0.009839196     -204.9717 \\
    -0.00961809     -204.4586 \\
    -0.009396985     -203.9069 \\
    -0.009175879     -203.3122 \\
    -0.008954774     -202.6697 \\
    -0.008733668      -201.974 \\
    -0.008512563     -201.2193 \\
    -0.008291457     -200.3991 \\
    -0.008070352     -199.5061 \\
    -0.007849246     -198.5326 \\
    -0.007628141     -197.4697 \\
    -0.007407035     -196.3079 \\
    -0.00718593     -195.0368 \\
    -0.006964824     -193.6451 \\
    -0.006743719     -192.1203 \\
    -0.006522613      -190.449 \\
    -0.006301508     -188.6166 \\
    -0.006080402     -186.6074 \\
    -0.005859296     -184.4045 \\
    -0.005638191     -181.9902 \\
    -0.005417085     -179.3452 \\
    -0.00519598     -176.4496 \\
    -0.004974874     -173.2825 \\
    -0.004753769      -169.822 \\
    -0.004532663     -166.0461 \\
    -0.004311558     -161.9323 \\
    -0.004090452     -157.4581 \\
    -0.003869347     -152.6018 \\
    -0.003648241     -147.3425 \\
    -0.003427136     -141.6611 \\
    -0.00320603     -135.5406 \\
    -0.002984925      -128.967 \\
    -0.002763819     -121.9301 \\
    -0.002542714     -114.4241 \\
    -0.002321608     -106.4488 \\
    -0.0021005     -98.0097 \\
    -0.0018794      -89.119 \\
    -0.00165829     -79.7963 \\
    -0.00143719     -70.0686 \\
    -0.00121608     -59.9703 \\
    -0.000994975     -49.5435 \\
    -0.000773869     -38.8369 \\
    -0.000552764     -27.9058 \\
    -0.000331658     -16.8103 \\
    -0.00011055     -5.6147 \\
    0.00011055      5.6147 \\
    0.000331658      16.8103 \\
    0.000552764      27.9058 \\
    0.000773869      38.8369 \\
    0.000994975      49.5435 \\
    0.00121608      59.9703 \\
    0.00143719      70.0686 \\
    0.00165829      79.7963 \\
    0.0018794       89.119 \\
    0.0021005      98.0097 \\
    0.002321608      106.4488 \\
    0.002542714      114.4241 \\
    0.002763819      121.9301 \\
    0.002984925       128.967 \\
    0.00320603      135.5406 \\
    0.003427136      141.6611 \\
    0.003648241      147.3425 \\
    0.003869347      152.6018 \\
    0.004090452      157.4581 \\
    0.004311558      161.9323 \\
    0.004532663      166.0461 \\
    0.004753769       169.822 \\
    0.004974874      173.2825 \\
    0.00519598      176.4496 \\
    0.005417085      179.3452 \\
    0.005638191      181.9902 \\
    0.005859296      184.4045 \\
    0.006080402      186.6074 \\
    0.006301508      188.6166 \\
    0.006522613       190.449 \\
    0.006743719      192.1203 \\
    0.006964824      193.6451 \\
    0.00718593      195.0368 \\
    0.007407035      196.3079 \\
    0.007628141      197.4697 \\
    0.007849246      198.5326 \\
    0.008070352      199.5061 \\
    0.008291457      200.3991 \\
    0.008512563      201.2193 \\
    0.008733668       201.974 \\
    0.008954774      202.6697 \\
    0.009175879      203.3122 \\
    0.009396985      203.9069 \\
    0.00961809      204.4586 \\
    0.009839196      204.9717 \\
    0.0100603      205.4501 \\
    0.01028141      205.8973 \\
    0.01050251      206.3166 \\
    0.01072362      206.7107 \\
    0.01094472      207.0823 \\
    0.01116583      207.4337 \\
    0.01138693      207.7669 \\
    0.01160804      208.0839 \\
    0.01182915      208.3863 \\
    0.01205025      208.6756 \\
    0.01227136      208.9533 \\
    0.01249246      209.2204 \\
    0.01271357      209.4781 \\
    0.01293467      209.7274 \\
    0.01315578      209.9691 \\
    0.01337688       210.204 \\
    0.01359799      210.4329 \\
    0.0138191      210.6564 \\
    0.0140402       210.875 \\
    0.01426131      211.0892 \\
    0.01448241      211.2996 \\
    0.01470352      211.5064 \\
    0.01492462      211.7101 \\
    0.01514573       211.911 \\
    0.01536683      212.1094 \\
    0.01558794      212.3055 \\
    0.01580905      212.4997 \\
    0.01603015       212.692 \\
    0.01625126      212.8827 \\
    0.01647236       213.072 \\
    0.01669347      213.2599 \\
    0.01691457      213.4467 \\
    0.01713568      213.6325 \\
    0.01735678      213.8173 \\
    0.01757789      214.0014 \\
    0.01779899      214.1846 \\
    0.0180201      214.3672 \\
    0.01824121      214.5492 \\
    0.01846231      214.7307 \\
    0.01868342      214.9117 \\
    0.01890452      215.0922 \\
    0.01912563      215.2724 \\
    0.01934673      215.4522 \\
    0.01956784      215.6317 \\
    0.01978894       215.811 \\
    0.02001005        215.99 \\
    0.02023116      216.1687 \\
    0.02045226      216.3473 \\
    0.02067337      216.5257 \\
    0.02089447       216.704 \\
    0.02111558      216.8821 \\
    0.02133668        217.06 \\
    0.02155779      217.2379 \\
    0.02177889      217.4157 \\
    0.022      217.5933 \\
};
\end{axis}
\end{tikzpicture}%

%% file: figs/stiffness_v2.tex
%
%
\definecolor{mycolor1}{rgb}{0.00000,0.44700,0.74100}%
\definecolor{mycolor2}{rgb}{0.00000,0.44700,0.74100}
\begin{tikzpicture}

\begin{axis}[%
width=0.951\fwidth,
height=\fheight,
at={(0\fwidth,0\fheight)},
scale only axis,
xmin=-0.8e-3,
xmax=0.8e-3,
xtick={-8e-4,-5e-4,0,5e-4,8e-4},
xlabel style={font=\color{white!15!black}},
xlabel={Torsion angle $\Delta q$ in $\mathrm{rad}$},
xticklabels={-8e-4,-5e-4,0,5e-4,8e-4},
ymin=-5000,
ymax=5000,
ytick={6000,4000,...,-6000},
yticklabels={6,4,2,0,-2,-4,-6},
ylabel style={font=\color{white!15!black}},
ylabel={Elastic torque $\tau_E$ in $\mathrm{kNm}$},
axis background/.style={fill=white},
legend pos=north west,
xmajorgrids,
ymajorgrids
]
\addplot [color=blue]
  table[row sep=crcr]{%
-0.0008707  -4890\\
-0.0005944	-2557\\
-0.0003385	-406.6\\
-0.0001453	-0\\
0.0001453	0\\
0.0003385	406.6\\
0.0005944	2557\\
0.0008707   4890\\};
\addlegendentry{$ \tau_{E} $}
\addplot [color=red]
  table[row sep=crcr, x expr=\thisrow{X}, y expr=\thisrow{Y}*1000]{
  X Y \\
    -0.001     -5.8985 \\
    -0.0009798     -5.7289 \\
    -0.0009596     -5.5593 \\
    -0.00093939     -5.3899 \\
    -0.00091919     -5.2206 \\
    -0.00089899     -5.0514 \\
    -0.00087879     -4.8824 \\
    -0.00085859     -4.7136 \\
    -0.00083838     -4.5451 \\
    -0.00081818     -4.3768 \\
    -0.00079798     -4.2089 \\
    -0.00077778     -4.0413 \\
    -0.00075758     -3.8742 \\
    -0.00073737     -3.7075 \\
    -0.00071717     -3.5415 \\
    -0.00069697     -3.3761 \\
    -0.00067677     -3.2115 \\
    -0.00065657     -3.0479 \\
    -0.00063636     -2.8852 \\
    -0.00061616     -2.7238 \\
    -0.00059596     -2.5637 \\
    -0.00057576     -2.4053 \\
    -0.00055556     -2.2486 \\
    -0.00053535      -2.094 \\
    -0.00051515     -1.9417 \\
    -0.00049495     -1.7921 \\
    -0.00047475     -1.6454 \\
    -0.00045455     -1.5021 \\
    -0.00043434     -1.3626 \\
    -0.00041414     -1.2273 \\
    -0.00039394     -1.0967 \\
    -0.00037374    -0.97126 \\
    -0.00035354    -0.85156 \\
    -0.00033333    -0.73812 \\
    -0.00031313    -0.63149 \\
    -0.00029293    -0.53222 \\
    -0.00027273    -0.44083 \\
    -0.00025253    -0.35778 \\
    -0.00023232    -0.28351 \\
    -0.00021212    -0.21831 \\
    -0.00019192    -0.16236 \\
    -0.00017172    -0.11569 \\
    -0.00015152     -0.0781 \\
    -0.00013131   -0.049157 \\
    -0.00011111   -0.028161 \\
    -9.0909e-05   -0.014127 \\
    -7.0707e-05  -0.0057974 \\
    -5.0505e-05  -0.0016951 \\
    -3.0303e-05 -0.00024718 \\
    -1.0101e-05 -3.4392e-06 \\
    1.0101e-05  3.4392e-06 \\
    3.0303e-05  0.00024718 \\
    5.0505e-05   0.0016951 \\
    7.0707e-05   0.0057974 \\
    9.0909e-05    0.014127 \\
    0.00011111    0.028161 \\
    0.00013131    0.049157 \\
    0.00015152      0.0781 \\
    0.00017172     0.11569 \\
    0.00019192     0.16236 \\
    0.00021212     0.21831 \\
    0.00023232     0.28351 \\
    0.00025253     0.35778 \\
    0.00027273     0.44083 \\
    0.00029293     0.53222 \\
    0.00031313     0.63149 \\
    0.00033333     0.73812 \\
    0.00035354     0.85156 \\
    0.00037374     0.97126 \\
    0.00039394      1.0967 \\
    0.00041414      1.2273 \\
    0.00043434      1.3626 \\
    0.00045455      1.5021 \\
    0.00047475      1.6454 \\
    0.00049495      1.7921 \\
    0.00051515      1.9417 \\
    0.00053535       2.094 \\
    0.00055556      2.2486 \\
    0.00057576      2.4053 \\
    0.00059596      2.5637 \\
    0.00061616      2.7238 \\
    0.00063636      2.8852 \\
    0.00065657      3.0479 \\
    0.00067677      3.2115 \\
    0.00069697      3.3761 \\
    0.00071717      3.5415 \\
    0.00073737      3.7075 \\
    0.00075758      3.8742 \\
    0.00077778      4.0413 \\
    0.00079798      4.2089 \\
    0.00081818      4.3768 \\
    0.00083838      4.5451 \\
    0.00085859      4.7136 \\
    0.00087879      4.8824 \\
    0.00089899      5.0514 \\
    0.00091919      5.2206 \\
    0.00093939      5.3899 \\
    0.0009596      5.5593 \\
    0.0009798      5.7289 \\
    0.001      5.8985 \\
};
\addlegendentry{$ \Tilde{\tau}_E $}
\end{axis}

\begin{axis}[%
width=0.45\fwidth,
height=0.314\fheight,
at={(0.454\fwidth,0.1\fheight)}, 
scale only axis,
xmin=-0.0004,
xmax=0.0004,
xtick = {-4e-4,0,4e-4},
xticklabels = {-4e-4,0,4e-4},
ymin=-1000,
ymax=1000,
ytick={-1000,0,1000},
yticklabels={-1,0,1},
axis background/.style={fill=white},
xmajorgrids,
ymajorgrids,
legend style={legend cell align=left, align=left, draw=white!15!black}
]
\addplot [color=blue]
  table[row sep=crcr]{%
-0.0003996	-915.9\\
-0.0003385	-406.6\\
-0.0001453	-0\\
0.0001453	0\\
0.0003385	406.6\\
0.0003996	915.9\\
};
\addplot [color=red]
  table[row sep=crcr, x expr=\thisrow{X}, y expr=\thisrow{Y}*1000]{
  X Y \\
    -0.0005     -1.8292 \\
    -0.0004899     -1.7551 \\
    -0.0004798     -1.6818 \\
    -0.0004697     -1.6093 \\
    -0.0004596     -1.5376 \\
    -0.00044949     -1.4669 \\
    -0.00043939     -1.3971 \\
    -0.00042929     -1.3284 \\
    -0.00041919     -1.2607 \\
    -0.00040909     -1.1942 \\
    -0.00039899     -1.1289 \\
    -0.00038889     -1.0648 \\
    -0.00037879     -1.0021 \\
    -0.00036869    -0.94078 \\
    -0.00035859    -0.88092 \\
    -0.00034848    -0.82259 \\
    -0.00033838    -0.76586 \\
    -0.00032828     -0.7108 \\
    -0.00031818    -0.65748 \\
    -0.00030808    -0.60596 \\
    -0.00029798    -0.55632 \\
    -0.00028788    -0.50861 \\
    -0.00027778    -0.46291 \\
    -0.00026768    -0.41926 \\
    -0.00025758    -0.37774 \\
    -0.00024747    -0.33838 \\
    -0.00023737    -0.30124 \\
    -0.00022727    -0.26635 \\
    -0.00021717    -0.23374 \\
    -0.00020707    -0.20345 \\
    -0.00019697    -0.17548 \\
    -0.00018687    -0.14983 \\
    -0.00017677     -0.1265 \\
    -0.00016667    -0.10545 \\
    -0.00015657   -0.086665 \\
    -0.00014646   -0.070076 \\
    -0.00013636   -0.055615 \\
    -0.00012626   -0.043196 \\
    -0.00011616   -0.032714 \\
    -0.00010606   -0.024045 \\
    -9.596e-05   -0.017048 \\
    -8.5859e-05   -0.011565 \\
    -7.5758e-05   -0.007423 \\
    -6.5657e-05  -0.0044365 \\
    -5.5556e-05  -0.0024104 \\
    -4.5455e-05  -0.0011453 \\
    -3.5354e-05 -0.00044456 \\
    -2.5253e-05  -0.0001228 \\
    -1.5152e-05 -1.6896e-05 \\
    -5.0505e-06 -2.2144e-07 \\
    5.0505e-06  2.2144e-07 \\
    1.5152e-05  1.6896e-05 \\
    2.5253e-05   0.0001228 \\
    3.5354e-05  0.00044456 \\
    4.5455e-05   0.0011453 \\
    5.5556e-05   0.0024104 \\
    6.5657e-05   0.0044365 \\
    7.5758e-05    0.007423 \\
    8.5859e-05    0.011565 \\
    9.596e-05    0.017048 \\
    0.00010606    0.024045 \\
    0.00011616    0.032714 \\
    0.00012626    0.043196 \\
    0.00013636    0.055615 \\
    0.00014646    0.070076 \\
    0.00015657    0.086665 \\
    0.00016667     0.10545 \\
    0.00017677      0.1265 \\
    0.00018687     0.14983 \\
    0.00019697     0.17548 \\
    0.00020707     0.20345 \\
    0.00021717     0.23374 \\
    0.00022727     0.26635 \\
    0.00023737     0.30124 \\
    0.00024747     0.33838 \\
    0.00025758     0.37774 \\
    0.00026768     0.41926 \\
    0.00027778     0.46291 \\
    0.00028788     0.50861 \\
    0.00029798     0.55632 \\
    0.00030808     0.60596 \\
    0.00031818     0.65748 \\
    0.00032828      0.7108 \\
    0.00033838     0.76586 \\
    0.00034848     0.82259 \\
    0.00035859     0.88092 \\
    0.00036869     0.94078 \\
    0.00037879      1.0021 \\
    0.00038889      1.0648 \\
    0.00039899      1.1289 \\
    0.00040909      1.1942 \\
    0.00041919      1.2607 \\
    0.00042929      1.3284 \\
    0.00043939      1.3971 \\
    0.00044949      1.4669 \\
    0.0004596      1.5376 \\
    0.0004697      1.6093 \\
    0.0004798      1.6818 \\
    0.0004899      1.7551 \\
    0.0005      1.8292 \\
  };
\end{axis}

\begin{axis}[%
width=1.227\fwidth, 
height=1.227\fheight,
at={(-0.16\fwidth,-0.135\fheight)},
scale only axis,
xmin=0,
xmax=1,
ymin=0,
ymax=1,
axis line style={draw=none},
ticks=none,
axis x line*=bottom,
axis y line*=left,
legend style={legend cell align=left, align=left, draw=white!15!black}
]
\draw[dashed, draw=black] (axis cs:0.32,0.45) rectangle (axis cs:0.71,0.6);
\addplot [color=black, dotted, forget plot]
  table[row sep=crcr]{%
0.71 0.6 \\
0.878787878787879	0.45 \\ 
};
\addplot [color=black, dotted, forget plot]
  table[row sep=crcr]{%
0.32 0.45 \\  
0.5003885003885	0.213270142180095\\
};
\end{axis}
\end{tikzpicture}%

%% file: sec_control_design_v2.tex
\section{Controller Design}
\label{sec:control}
\begin{figure}
	\centering
	\resizebox{\columnwidth}{!}{%
	\input{figs/exact_linearization_tikz_v2}}
	\caption{Controller design with exact linearization}
	\label{fig:exact_linear_feedback}
\end{figure}
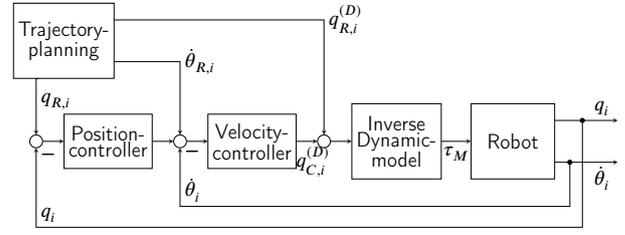

We propose a model-based feed forward controller to account for the dynamical system and a feedback control law for compensation of unknown disturbances and model errors. 
In Fig.~\ref{fig:closed-loop-model} and Fig.~\ref{fig:exact_linear_feedback} two different control architectures are presented. Both of them contain a feed forward controller, a feedback controller and trajectory planning. 

\begin{figure}
	\centering
	\resizebox{\columnwidth}{!}{%
	\input{figs/closed_loop_tikz_v2}}
	\caption{Controller design with feed forward and feedback control}
	\label{fig:closed-loop-model}
\end{figure}
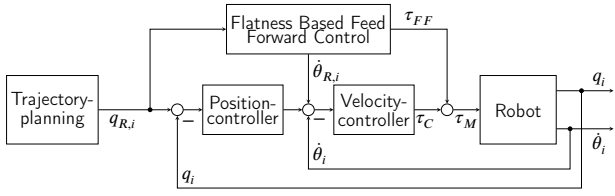

The trajectory planning generates the reference signals, i.e. the link angle $ q_{R} $ and its derivatives of a desired trajectory. A motor torque $\tau_M$ is applied to the robot and the link angle $q$ and the motor velocity $\dot{\theta}$ are the control variables of the feedback controller. The link angle $q$ can be measured directly using SE and the motor velocity $\dot{\theta}$ can be measured using the motor resolver. For both sensors, noise is addressed by implementing low pass filters. 

Regarding the feed forward controller, there are two different architectures, as presented in Fig.~\ref{fig:closed-loop-model} and Fig.~\ref{fig:exact_linear_feedback}. The design in Fig.~\ref{fig:closed-loop-model} is referred to flatness based feed forward controller, whereas the structure in Fig.~\ref{fig:exact_linear_feedback} is known as exact linearization. Depending on the literature, both are referred to as computed torque control \cite{Moberg2008}. To reduce the risk of confusion, we will avoid the term computed torque control in this contribution. The flatness based design is explained in detail in Section ~\ref{sec:flatness} and computes the feed forward torque independently of the current states. It relies exclusively on the reference trajectory and its derivatives and therefore, it can potentially be computed offline. As a consequence, it is invulnerable to any measurement noise. The flatness based architecture is independent of the feedback control and therefore, the feedback control is not time delayed or modified in any sense. It contains the disadvantage of a potentially less precise model since it relies on the reference trajectory rather than the measured trajectory. However, it is possible to partially include online measurements to improve the model accuracy. The inertia matrix, gravity, Coriolis and centripetal terms can be updated at run time based on online measurements. 

The exact linearization architecture as presented in Fig.~\ref{fig:exact_linear_feedback}, attempts to eliminate the nonlinearity in robot dynamics in an inner loop and therefore control a simple, linear system with the feedback controller. A detailed explanation is given in \cite{Spong1987}. Basically, the feedback controller computes an angular acceleration for a dynamic inversion-based controller in order to compensate  nonlinear dynamics. 
Its advantage is that the feedback controller is automatically adjusted to the current robot state. Disadvantages are a necessity of a fast computation of the inverse dynamic model in the inner loop and a potential time delay of the feedback control. In \cite{Nguyen-Tuong2008} both designs are compared and the flatness based architecture outperforms the exact linearization structure with several feed forward models regarding the achieved trajectory accuracy. The authors in \cite{Nguyen-Tuong2008} argue that the main advantage of a feed forward structure compared to exact linearization is the direct and non-delayed impact of the feedback control law. Based on these arguments, we decide to apply a flatness based architecture rather than exact linearization.
The exact linearization module does not account for joint elasticity and therefore, the inner loop reduces to a double integrator. The main idea, to apply the inverse of the model, remains the same in the flatness based feed forward and the exact linearization case. Assume a model inverse of the robot is given by the following form
\begin{equation}
    \tau_M = f( q_{U}, \dot{q}_{U}, \cdots, {q}^{(D)}_{U} ),
    \label{eq:model_inverse}
\end{equation}
with an input angle $q_{U}$ and its derivatives. We define the derivative as ${q}^{(D)} = \frac{d^D{q}}{d t^D}$. In the flatness based architecture, we set $q_{U} = q_{R}$. In the exact linearization formulation, we apply
\begin{equation}
    {q}^{(D)}_{U} = {q}^{(D)}_{R} + {q}^{(D)}_{C},
\end{equation}
with the feedback control variable ${q}^{(D)}_{C}$. With an elastic joint model the derivative order is $D = 4$, whereas with the rigid joint model the derivative order is $D = 2$ \cite{Moberg2008}. So, for an elastic joint model we obtain
\begin{equation}
    \ddot{q}_{U} = \ddot{q}_{R} + \ddot{q}_{C},
\end{equation}
and for a model with elastic joints we get
\begin{equation}
    {q}^{(4)}_{U} = {q}^{(4)}_{R} + {q}^{(4)}_{C}.
\end{equation}

 In the exact linearization case it is beneficial to apply the highest input derivative. Note that the output of the velocity controller $u_{C}$ depends on the architecture. In the flatness based case, it is equal to the motor control torque, i.e. $u_{C} = \tau_C$. In the exact linearization case, it computes the control input of the inverse dynamics model, i.e. $ u_{C} = {q}^{(D)}_{C}$. In Section ~\ref{sec:flatness}, we derive the flatness based module and thus estimate \eqref{eq:model_inverse}.

\subsection{Feedback Controller}
\label{sec:control_fb}

The enhancement of the feedback controller is a core contribution of this paper. We implement a PD-controller with a link-side position controller using SE and a motor-side velocity controller based on the resolver signal. Considering elastic joints it is difficult to precisely calculate the motor reference signals since the equation $\theta_{R} = u \, q_{R}$ is only valid for rigid joints. A precise calculation of the motor reference signals requires the full, nonlinear stiffness model \eqref{eq:robotdynamics} and \eqref{eq:robotdynamics2}. Therefore, the exact calculation of the motor reference signals is sensitive to modelling errors. In most previous works, this consideration is simply neglected and the rigid link equation $\theta_{R} = u \, q_{R}$ is used for the motor-side position and velocity controller. Regarding position control, this problem can be solved by measuring $q$ and implementing a link-side position control. Unfortunately, a link-side velocity control only partially solves this issue. A link velocity speed control utilizes the correct reference speed $\dot{q}_{R}$ but it causes two additional issues. First, a comparative low sensor resolution due to the missing transmission factor requires a significant low pass filtering. Second, to avoid stability issues caused by the elastic joint a soft velocity control parameter is necessary. Both drawbacks can be addressed by implementing a motor-side velocity controller utilizing the full nonlinear elastic joint model for calculating the correct motor reference speed. We present in experiments on the real robot the significance of this effect.

For a compact notation, we define the conventional feedback controller, which neglects joint elasticity as C-FB and the model-based feedback controller, which accounts for joint elasticity as MB-FB. The estimation of the reference velocity $\dot{\theta}_R$ model-based feedback controller is presented in Section~\ref{sec:flatness}. The source code of the implementation is given in Appendix~\ref{app:code} and is calculated jointly with the flatness based controller.

Both, the position and velocity controllers are realized as proportional controllers. The position controller is given by
 \begin{equation}
\dot{\theta}_{C} =  K_{P} \, (q_{R}  - q),
 \label{eq:pos_controller}
 \end{equation}
 with the proportional position gain $K_{P}$ and the motor velocity $\dot{\theta}_{C}$. The velocity controller using the proportional speed gain $K_{V}$ has the form
 \begin{equation}
 \tau_{C} =  K_{V} \, (\dot{\theta}_{C} + \dot{\theta}_{R} - \dot{\theta}),
 \label{eq:vel_controller}
 \end{equation}
 All in all, the control signal sums up to
\begin{equation}
  u_{vel} = \tau_{FF} + \tau_{C}.  
  \label{eq:complete_controller}
 \end{equation}


\subsection{Flatness Based Feed Forward Controller}
\label{sec:flatness}

The main task of the feed forward control law is to achieve a good guiding behavior, whereby the feedback control law can be applied exclusively for the compensation of disturbances and model uncertainties. For the design of the feed forward control law, the inverse nonlinear model of the robot link is used. Therefore, the nonlinear model \eqref{eq:robotdynamics} and \eqref{eq:robotdynamics2} is solved for the motor torque $ \tau_{M} $, valid for each link, in the form
\begin{align}
\tau_{M,i} = J_{i,i} \, \ddot{\theta}_i +  \frac{M_{i,i} \, \ddot{q}_i + \tau_{A,i} + \tau_{CC,i} + g_i + \tilde{\tau}_{F,i}(\dot{q}_i)}{u_i}.
\label{eq:motor_torque}
\end{align}

Equation \eqref{eq:motor_torque} describes the motor torque applied in a specific operating point. We distinguish between variables such as $q_i$ and $\theta_i$ and parameters such as $M_{i,i}$, $J_{i,i}$, $\tau_{A,i}$, $\tau_{CC,i}$ and $g_i$. We neglect all parameter changes during one time interval, in the presented case for $0.8\, \mathrm{ms}$. The parameters are updated each $0.8\, \mathrm{ms}$ according to the state trajectory but are not considered as differentiable variables in the flatness based feed forward control law. 

For a good guiding behavior, i.e. $q_i = q_{R,i}$, the motor position is set to the motor reference position, i.e. $\theta_i = \theta_{R,i}$ and $\tau_{M,i} = \tau_{FF,i}$ is applied. It follows 
\newpage
\begin{align}
\label{eq:motor_torque_ff}
\tau_{FF,i} &= J_{i,i} \, \ddot{\theta}_{R,i} \\
&+  \frac{M_{i,i} \, \ddot{q}_{R,i} + \tau_{A,i} + \tau_{CC,i} + g_i + \tilde{\tau}_{F,i}(\dot{q}_{R,i})}{u} \nonumber.
\end{align}

In order to solve \eqref{eq:motor_torque_ff} we need to substitute $\ddot{\theta}_{R,i}$ with a function of $q_{R,i}$ and its derivatives. The following Section shows how to determine each term in \eqref{eq:motor_torque_ff}.

\begin{enumerate}
\item The continuous differentiable friction torque $ \tilde{\tau}_{F,i}(\dot{q_{R,i}}) $ is defined in \eqref{eq:mod_friction_app}.
\item The Coriolis and centripetal torques are summarized for each link in $\tau_{CC,i}$ with
\begin{equation}
    \tau_{CC,i} = \sum_{j = 1}^{N} c_{i,j} \, q_j, \qquad i \neq j.
\end{equation}
Note that $\tau_{CC,i}$ is not a function of $q_i$. $\tau_{CC,i}$ can be calculated using the kinematics of the robot.
\item The link inertia $M_{i,i}$ and the link gravity torque $g_i$ can be obtained from the kinematics of the robot. We summarize the inertia torques caused by other links in the acceleration torque $\tau_{A,i}$
\begin{equation}
    \tau_{A,i} = \sum_{j = 1}^{N} M_{i,j} \, \ddot{q}_{R, j}, \qquad i \neq j.
\end{equation}
\item The motor angle $\theta_i$ is substituted in \eqref{eq:motor_torque_ff} using the inverse nonlinear stiffness model. We use the torsion angle $ \Delta q_i $ as defined in \eqref{eq:torsion_diff_angle} and solve for $\theta_i$ 

\begin{equation}
 \theta_i = u_i \, ( \Delta q_i + q_i ).
\end{equation}

The torsion angle $\Delta q_i$ is substituted using the inverse, continuously differentiable stiffness model \eqref{eq:mod_stiff_app} leading to

\begin{equation}
 \theta_i = u_i \, \left( \frac{\tilde{\tau}_{E,i}}{c_{TR,i}} + \frac{2 \, \phi_{B,i}}{1+e^{-s_{E,i} \, \tilde{\tau}_{E,i}}} - \phi_{B,i} + q_i \right).
\end{equation}

The elastic torque is substituted using the sum of torques of the link. Furthermore, we set $\theta_i = \theta_{R,i} $ and $q_i = q_{R,i}$ and obtain

\begin{equation} \label{eq:feedforward}
\begin{split}
 \theta_{R,i} &= u_i \, ( \frac{M_{i,i} \, \ddot{q}_{R,i} + \tau_{A,i} + \tau_{CC,i} + g_i + \tilde{\tau}_{F,i}(\dot{q}_{R,i})}{c_{TR,i}} \\
 &\qquad + \frac{2 \, \phi_{B,i}}{1+e^{-s_{E,i} \, \left( M_{i,i} \, \ddot{q}_{R,i} + \tau_{A,i} + \tau_{CC,i} + g_i + \tilde{\tau}_{F,i}(\dot{q}_{R,i}) \right)}} \\ &\qquad - \phi_{B,i} + q_{R,i} ).
\end{split}
\end{equation}

Obtaining the second derivative of \eqref{eq:feedforward}, i.e. $ \ddot{\theta}_{R,i} = \frac{d^2 \theta_{R,i}}{dt^2} $, completes the solution for $\tau_{FF,i}$  in \eqref{eq:motor_torque_ff}.  \textit{MATLAB Symbolic Math Toolbox} was used for solving the equations. 
It should be pointed out, that due to the pre-computation of the solution using AD, the run time of the feed forward calculation requires only a few microseconds. This is only possible because the continuously differentiable formulation of the model allows to use AD tools. The complete source code in C language is given in Appendix \ref{app:code}. \\
We applied the \textit{Robotics Toolbox} \cite{Corke2017RoboticsControl} for the calculation of the inertia matrix $\mathbf{M}(\mathbf{q})$, the gravity load $\mathbf{g}$, the Coriolis and centripetal matrix $\mathbf{C}(\mathbf{q}, \mathbf{\dot{q}})$, as well as for the forward and inverse kinematics. Denavit-Hartenberg parameters are derived using a CAD model of the robot. We included material densities in the CAD model and  utilized the CAD model to calculate mass, inertia, and center of gravity for each link with finite element method (FEM).
\end{enumerate}



%% file: figs/closed_loop_tikz_v2.tex
\tikzstyle{block} = [draw, rectangle, 
minimum height=1.3cm, minimum width=2cm]
\tikzstyle{sum} = [draw, circle]
\tikzstyle{input} = [coordinate]
\tikzstyle{output} = [coordinate]
\tikzstyle{pinstyle} = [pin edge={<- ,black}]
\tikzstyle{branch} = [circle,inner sep=0pt,minimum size=1mm,fill=black,draw=black]

\begin{tikzpicture}[auto, node distance=0.5cm,>=stealth,every text node part/.style={align=center}]

\node [block,minimum height=1.8cm, name=input] {\Large Trajectory-\\\Large planning};
\node [branch,right= 1.3 of input] (bul) {};
\node [sum,  right= of bul] (sum) {};
\node [block, right= of sum] (controller) {\Large Position-\\ \Large controller};
\node [sum, right= of controller] (sum2) {};
\node [block, right= of sum2] (controller2) {\Large Velocity-\\\Large controller};
\node [block, above= 1.3cm of sum2] (flat) {\Large Flatness Based Feed\\\Large Forward Control};
\node [sum, right= 0.7cm of controller2] (sum3) {};
\node [block, right= 0.7cm of sum3, minimum height=1.8cm] (system) {\Large Robot};
\node [branch, right= 0.6 of system, yshift=0.5cm] (y1) {};
\node [branch, right= 0.3 of system, yshift=-0.5cm] (y2) {};
\node [output, right= 1.5cm of system, yshift=0.5cm] (aus1) {};
\node [output, right= 1.5cm of system, yshift=-0.5cm] (aus2) {};

\draw [->] (flat.south) -| node[right, pos=0.68] {\Large $\dot{\theta}_{R,i}$} (sum2);
\draw [->] (input.east) -- node[name=u2, below, pos=0.33] {\Large $q_{R,i}$} (sum);
\draw [->] (bul) |-  (flat);
\draw [->] (flat) -| node[near start] {\Large $\tau_{FF}$} (sum3);
\draw [->] (sum) -- (controller);
\draw [->] (controller) -- (sum2);
\draw [->] (sum2) -- (controller2);
\draw [->] (controller2) -- node[below,pos=0.4] {\Large $\tau_C$} (sum3);
\draw [->] (sum3) -- node[below] {\Large $\tau_M$} (system);
\draw [->] ([yshift=0.5cm] system.east) -- node[near end] {\Large $q_{i}$} (aus1);
\draw [->] ([yshift=-0.5cm] system.east) -- node[below, near end] {\Large $\dot{\theta}_{i}$}(aus2);
\draw [->] (y1.south) -- ++(0,-2.5cm) -| node[pos=0.95, right] {\Large $-$}  node[above right] {\Large $q_{i}$} (sum.south);
\draw [->] (y2.south) -- ++(0,-1cm) -| node[pos=0.95, right] {\Large $-$}  node[above right] {\Large $\dot{\theta}_{i}$} (sum2.south);


\end{tikzpicture}

%% file: figs/exact_linearization_tikz_v2.tex
\tikzstyle{block} = [draw, rectangle, 
minimum height=1.3cm, minimum width=2cm]
\tikzstyle{sum} = [draw, circle]
\tikzstyle{input} = [coordinate]
\tikzstyle{output} = [coordinate]
\tikzstyle{pinstyle} = [pin edge={<- ,black}]
\tikzstyle{branch} = [circle,inner sep=0pt,minimum size=1mm,fill=black,draw=black]

\begin{tikzpicture}[auto, node distance=0.5cm,>=stealth,every text node part/.style={align=center}]

\node [block,minimum height=1.8cm] (input) {\Large Trajectory-\\\Large planning};
\node [coordinate,below = 1.5cm of input, yshift=0cm] (bul) {};
\node [sum, left = 0.5cm of bul] (sum) {};
\node [block, right= of sum] (controller) {\Large Position-\\ \Large controller};
\node [sum, right= of controller] (sum2) {};
\node [block, right= of sum2] (controller2) {\Large Velocity-\\\Large controller};
\node [sum, right= 0.5cm of controller2] (sum4) {};
\node [block, right= 0.5cm of sum4, minimum height=1.8cm] (dynamics) {\Large Inverse \\ \Large Dynamic-\\ \Large model};
\node [block, right= 0.7cm of dynamics, minimum height=1.8cm] (system) {\Large Robot};
\node [branch, right= 0.6 of system, yshift=0.5cm] (y1) {};
\node [branch, right= 0.3 of system, yshift=-0.5cm] (y2) {};
\node [output, right= 1.5cm of system, yshift=0.5cm] (aus1) {};
\node [output, right= 1.5cm of system, yshift=-0.5cm] (aus2) {};
\draw [->] ([yshift=-0.5cm] input.east) -| node[name=u1] {\Large $\dot{\theta}_{R,i}$} (sum2);
\draw [->] ([yshift=0.5cm] input.east) -| node[name=u3] {\Large $q^{(D)}_{R,i}$} (sum4);
\draw [->] ([xshift=-0.6cm] input.south) -| node[name=u2, right, pos=0.7] {\Large $q_{R,i}$} (sum);
\draw [->] (sum) -- (controller);
\draw [->] (controller) -- (sum2);
\draw [->] (sum2) -- (controller2);
\draw [->] (controller2) -- node[below,pos=0.78] {\Large $q^{(D)}_{C,i}$} (sum4);
\draw [->] (sum4) -- (dynamics);
\draw [->] (dynamics) -- node[below] {\Large $\tau_M$} (system);
\draw [->] ([yshift=0.5cm] system.east) -- node[near end] {\Large $q_{i}$} (aus1);
\draw [->] ([yshift=-0.5cm] system.east) -- node[below, near end] {\Large $\dot{\theta}_{i}$}(aus2);
\draw [->] (y1.south) -- ++(0,-2.5cm) -| node[pos=0.95, right] {\Large $-$}  node[above right] {\Large $q_{i}$} (sum.south);
\draw [->] (y2.south) -- ++(0,-1cm) -| node[pos=0.95, right] {\Large $-$}  node[above right] {\Large $\dot{\theta}_{i}$} (sum2.south);

\end{tikzpicture}

%% file: sec_simulation.tex
\section{Simulation Results}
\label{sec:simulation}

Before applying the presented algorithm on the \textit{KUKA Quantec KR300 Ultra SE}, we analyze the feed forward and feedback algorithms in simulation. We compare the flatness based feed forward controller with a nonlinear rigid model feed forward controller. We employ perfect model knowledge, neglect sensor noise and neglect any disturbances in order to focus on pure modelling differences. Unlike on the real robot, we can disable the feedback controller if needed in simulation. We compare the presented algorithm with a nonlinear feed forward control law without elastic joints, i.e. $\boldsymbol{\theta} = \mathbf{U} \, \mathbf{q}$. As a result, the model in \eqref{eq:robotdynamics2} reduces to

\begin{align} 
\left(\mathbf{M}(\mathbf{q}) + \mathbf{J} \mathbf{U}^2\right) \ddot{\mathbf{q}}+\mathbf{C}(\mathbf{q},\dot{\mathbf{q}})\dot{\mathbf{q}} + \mathbf{g}(\mathbf{q})   + \mathbf{\tau_F}(\dot{\mathbf{q}}) = \mathbf{U} \mathbf{\tau_M}.
\label{eq:robotdynamics3}
\end{align}

Using \eqref{eq:robotdynamics3}, the nonlinear feed forward control can be obtained straightforward

\begin{align} 
 \mathbf{\tau_{FF, R}} =& \, \mathbf{U}^{-1}\left(\left(\mathbf{M}(\mathbf{q}_R) + \mathbf{J} \mathbf{U}^2\right) \ddot{\mathbf{q}}_R \right.\nonumber\\*
 &+\left.\mathbf{C}(\mathbf{q}_R,\dot{\mathbf{q}}_R)\dot{\mathbf{q}}_R + \mathbf{g}(\mathbf{q}_R)   + \mathbf{\tau_F}(\dot{\mathbf{q}}_R) \right).
\label{eq:nonlinFeedForward}
\end{align}

For an unbiased comparison, we apply the same nonlinear friction \eqref{eq:mod_friction}, Coriolis, centripetal and gravity torque as in the flatness based controller. All model parameters are identical, including the pose-dependent inertia matrix. As a consequence, the only difference between the flatness based and nonlinear rigid model feed forward control law is neglecting the joint elasticity. We independently simulate both feed forward controllers on the same model, \eqref{eq:robotdynamics} and \eqref{eq:robotdynamics2}. For a compact notation, we define flatness based feed forward control as FB-FF and nonlinear rigid model feed forward control as R-FF, presented in \eqref{eq:motor_torque_ff} and \eqref{eq:nonlinFeedForward} respectively.

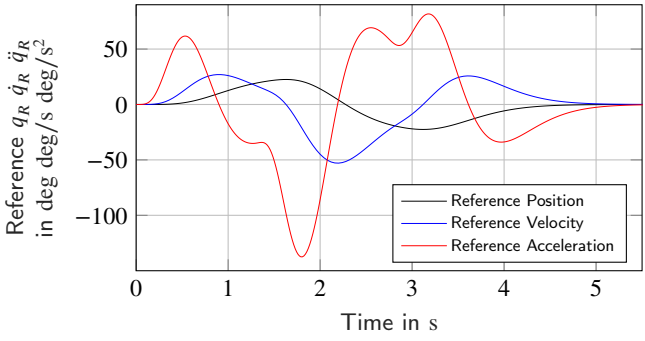
\begin{figure}
	\centering
	\setlength\fheight{3.5cm}
	\setlength\fwidth{7cm}
	\input{figs/sim_reference_plot}
	\caption{Reference trajectory with angular position, velocity and acceleration used for simulation. 
	}
  \label{fig:sim_ref}
\end{figure}

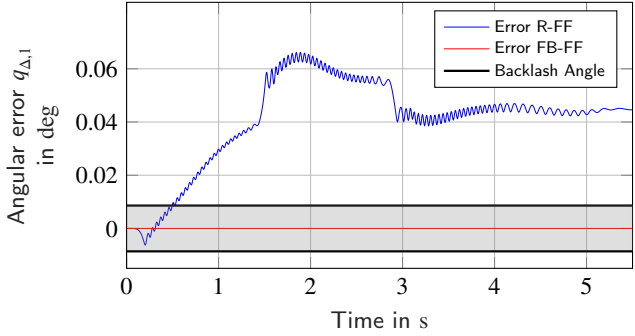
\begin{figure}
	\centering
	\setlength\fheight{3.5cm}
	\setlength\fwidth{7cm}
	\input{figs/sim_error_plot_v4}
	\caption{Angular error in simulation with perfect model knowledge, without noise, without feedback controller and without any disturbances. Gray area represents the backlash angle for comparison. FB-FF refers to the flatness based controller and R-FF to the nonlinear rigid model controller.}
   \label{fig:sim_error}
\end{figure}

Fig.~\ref{fig:sim_ref} presents the reference trajectory, velocity and acceleration used for simulation. For a clear overview, we did not illustrate the jerk and the jerk derivative, although both are continuously differentiable and are used in the FB-FF. For a demanding simulation, we did not choose a standard acceleration and deceleration phase, but rather the trajectory acceleration shown in Fig.~\ref{fig:sim_ref}. 
In theory, since the simulation does not contain any noise, model errors or external disturbances, the flatness based feed forward controller should perfectly follow an arbitrary trajectory. As presented in Fig.~\ref{fig:sim_error}, this can be achieved in simulation. Fig.~\ref{fig:sim_error} shows the angular error, i.e. $q_{\Delta,i} = q_{R,i} - q_i$, for the FB-FF and R-FF in simulation. For comparison, the range of the backlash angle $\phi_{B*}$ is displayed as a gray area in Fig.~\ref{fig:sim_error}. As expected, the R-FF is slightly worse than the FB-FF. However, a core insight of Fig.~\ref{fig:sim_error} is the performance of the R-FF controller on the elastic joint model. Considering the dynamic trajectory, we expected angular errors significantly greater than $0.06 \, \mathrm{deg}$. Fig.~\ref{fig:sim_error} is representative for many simulations confirming this result. If a model error is simulated, i.e. the feed forward model parameters do not match the simulation model parameters, the angular error easily exceeds $>1 \, \mathrm{deg}$. Note that the simulation presented in Fig.~\ref{fig:sim_error} does not apply any feedback control. If an additional feedback controller is applied, which does not account for elastic joints, the angular error increases by an order of magnitude. Therefore, it is very important to implement model based feedback (MB-FB) instead of conventional feedback (C-FB) as explained in Section~\ref{sec:control_fb}.

\begin{figure}
	\centering
	\setlength\fheight{3.5cm}
	\setlength\fwidth{7cm}
	\input{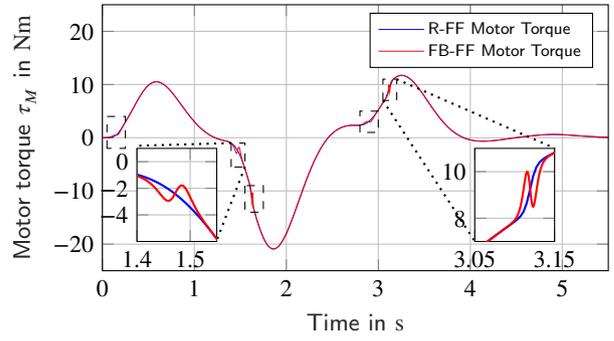}
	\caption{Feed forward motor torque in simulation. FB-FF refers to the flatness based controller and R-FF to the nonlinear rigid model controller.}
    \label{fig:sim_ff_torque}
\end{figure}

\begin{figure}
	\centering
	\setlength\fheight{3.5cm}
	\setlength\fwidth{7cm}
	\input{figs/sim_elastic_torque_plot_v3}
	\caption{Elastic joint torques in simulation. FB-FF refers to the flatness based controller and R-FF to the nonlinear rigid model controller.}
    \label{fig:sim_elastic_torque}
\end{figure}
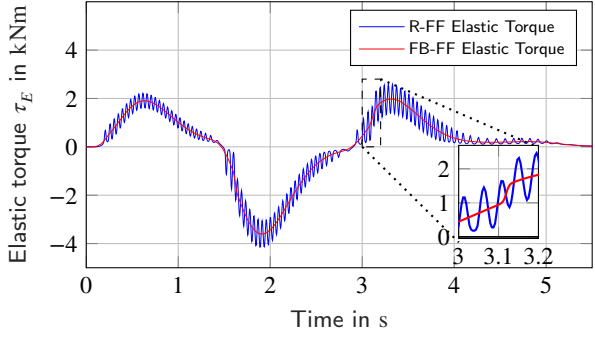

Fig.~\ref{fig:sim_ff_torque} compares the feed forward torque of R-FF and FB-FF. Due to the identical friction, inertia, gravity, Coriolis and centripetal terms of FB-FF and R-FF, the computed motor torques in Fig.~\ref{fig:sim_ff_torque} is broadly similar. Differences only occur in the Sections where the model traverses backlash or Coulomb friction. Besides compensating backlash, see Fig.~\ref{fig:sim_error}, these minor changes in motor torque have huge effects on the elastic torque of the model. As presented in Fig.~\ref{fig:sim_elastic_torque}, oscillations induced by backlash within the joint are compensated. Note that we did not employ a SE feedback controller in the simulations, which would be able to damp these oscillations. However, we argue that it is beneficial, if these oscillations are not induced in the first place by a proper flatness based control. 

Nevertheless, we identified some detrimental aspects during our analysis in simulation. First of all, modeling backlash as a flat function yields increased motor torque change rates. It is generally known that the less flat a system is, the more dynamic the input variable should be. On a real robot a dynamic input is not desirable. As presented in Fig.~\ref{fig:sim_ff_torque}, the torque change rates can be reduced to a reasonable level for the \textit{KUKA Quantec KR300 Ultra SE}. However, the demanded torque change rate is the bottleneck of the presented algorithm. This affects the change rate only and the absolute limit of the motor torque was not problematic in any analysis.

Second, backlash is only one source of positional errors. Model errors, sensor noise, external disturbances and conventional feedback controllers have a huge impact on positioning accuracy. Neither, the nonlinear nor the flatness based feed forward torque can account for these effects. For achieving a high positioning accuracy, a model-based feedback controller, ideally with SE, is absolutely necessary.

The parameters used for simulation correspond to joint~$1$ and are given in Tab.~\ref{tab:alltableparameters}.

%% file: figs/sim_reference_plot.tex
\begin{tikzpicture}
\begin{axis}[%
width=0.951\fwidth,
height=\fheight,
at={(0\fwidth,0\fheight)},
scale only axis,
xmin=0,
xmax=5.5,
xtick={0,1,...,5},
xlabel style={font=\color{white!15!black}},
xlabel={Time in $\mathrm{s}$},
ymin=-150,
ymax= 90,
ytick={100, 50,...,-100},
ylabel= Reference $q_R$ $\dot{q}_R$ $\ddot{q}_R$\\ in $\mathrm{deg}$ $\mathrm{deg}/\mathrm{s}$  $\mathrm{deg}/\mathrm{s^2}$,
ylabel style={font=\color{white!15!black}, align=center},
axis background/.style={fill=white},
xmajorgrids,
ymajorgrids,
legend cell align={left},
legend pos=south east
]
\addplot[line width=0.1mm, color=black]
  table[row sep=crcr, x expr=\thisrow{X}, y expr=\thisrow{Y}*57.2958]{
    X Y \\
      0  0 \\
    0.04  6.6325e-09 \\
    0.08  6.7317e-07 \\
    0.12  9.1287e-06 \\
    0.16  5.4332e-05 \\
    0.2  0.00020603 \\
    0.24  0.00058774 \\
    0.28    0.001378 \\
    0.32      0.0028 \\
    0.36   0.0051014 \\
    0.4   0.0085309 \\
    0.44    0.013314 \\
    0.48    0.019632 \\
    0.52    0.027609 \\
    0.56    0.037302 \\
    0.6    0.048697 \\
    0.64    0.061717 \\
    0.68    0.076227 \\
    0.72    0.092046 \\
    0.76     0.10896 \\
    0.8     0.12672 \\
    0.84     0.14509 \\
    0.88     0.16381 \\
    0.92     0.18265 \\
    0.96     0.20138 \\
    1      0.2198 \\
    1.04     0.23774 \\
    1.08     0.25504 \\
    1.12      0.2716 \\
    1.16     0.28731 \\
    1.2     0.30212 \\
    1.24     0.31596 \\
    1.28     0.32884 \\
    1.32     0.34073 \\
    1.36     0.35166 \\
    1.4     0.36163 \\
    1.44     0.37061 \\
    1.48     0.37852 \\
    1.52     0.38515 \\
    1.56     0.39019 \\
    1.6     0.39325 \\
    1.64     0.39389 \\
    1.68     0.39167 \\
    1.72     0.38619 \\
    1.76     0.37715 \\
    1.8     0.36434 \\
    1.84     0.34769 \\
    1.88     0.32729 \\
    1.92     0.30333 \\
    1.96     0.27611 \\
    2     0.24603 \\
    2.04     0.21355 \\
    2.08     0.17919 \\
    2.12     0.14346 \\
    2.16     0.10689 \\
    2.2    0.069964 \\
    2.24    0.033151 \\
    2.28  -0.0031338 \\
    2.32   -0.038528 \\
    2.36   -0.072727 \\
    2.4    -0.10548 \\
    2.44    -0.13659 \\
    2.48    -0.16593 \\
    2.52    -0.19339 \\
    2.56    -0.21894 \\
    2.6    -0.24255 \\
    2.64    -0.26426 \\
    2.68     -0.2841 \\
    2.72    -0.30214 \\
    2.76    -0.31847 \\
    2.8    -0.33319 \\
    2.84    -0.34636 \\
    2.88    -0.35803 \\
    2.92     -0.3682 \\
    2.96    -0.37682 \\
    3    -0.38376 \\
    3.04     -0.3889 \\
    3.08    -0.39206 \\
    3.12    -0.39312 \\
    3.16    -0.39196 \\
    3.2    -0.38853 \\
    3.24    -0.38284 \\
    3.28    -0.37494 \\
    3.32    -0.36496 \\
    3.36    -0.35309 \\
    3.4    -0.33953 \\
    3.44    -0.32455 \\
    3.48    -0.30843 \\
    3.52    -0.29143 \\
    3.56    -0.27384 \\
    3.6    -0.25594 \\
    3.64    -0.23796 \\
    3.68    -0.22013 \\
    3.72    -0.20265 \\
    3.76    -0.18569 \\
    3.8    -0.16939 \\
    3.84    -0.15385 \\
    3.88    -0.13916 \\
    3.92    -0.12537 \\
    3.96    -0.11251 \\
    4     -0.1006 \\
    4.04   -0.089634 \\
    4.08   -0.079591 \\
    4.12   -0.070442 \\
    4.16    -0.06215 \\
    4.2   -0.054668 \\
    4.24   -0.047946 \\
    4.28   -0.041934 \\
    4.32   -0.036577 \\
    4.36   -0.031821 \\
    4.4   -0.027614 \\
    4.44   -0.023906 \\
    4.48   -0.020647 \\
    4.52   -0.017793 \\
    4.56     -0.0153 \\
    4.6   -0.013129 \\
    4.64   -0.011243 \\
    4.68  -0.0096088 \\
    4.72  -0.0081969 \\
    4.76  -0.0069796 \\
    4.8  -0.0059325 \\
    4.84  -0.0050339 \\
    4.88  -0.0042643 \\
    4.92  -0.0036065 \\
    4.96  -0.0030454 \\
    5  -0.0025677 \\
    5.04  -0.0021618 \\
    5.08  -0.0018174 \\
    5.12  -0.0015258 \\
    5.16  -0.0012792 \\
    5.2  -0.0010711 \\
    5.24 -0.00089569 \\
    5.28 -0.00074808 \\
    5.32 -0.00062405 \\
    5.36 -0.00051997 \\
    5.4 -0.00043275 \\
    5.44 -0.00035976 \\
    5.48 -0.00029876 \\
    5.52 -0.00024784 \\
    5.56 -0.00020539 \\
};
\addlegendentry{Reference Position}
\addplot[line width=0.1mm, color=blue]
  table[row sep=crcr, x expr=\thisrow{X}, y expr=\thisrow{Y}*57.2958]{
    X Y \\
    0  0 \\
    0.04  1.1221e-06 \\
    0.08  5.5005e-05 \\
    0.12  0.00047988 \\
    0.16   0.0020652 \\
    0.2    0.006034 \\
    0.24      0.0138 \\
    0.28    0.026653 \\
    0.32    0.045487 \\
    0.36     0.07063 \\
    0.4     0.10179 \\
    0.44     0.13812 \\
    0.48     0.17831 \\
    0.52     0.22077 \\
    0.56     0.26378 \\
    0.6     0.30564 \\
    0.64      0.3448 \\
    0.68     0.37996 \\
    0.72     0.41007 \\
    0.76     0.43445 \\
    0.8     0.45267 \\
    0.84     0.46463 \\
    0.88     0.47045 \\
    0.92     0.47047 \\
    0.96     0.46518 \\
    1     0.45518 \\
    1.04     0.44114 \\
    1.08     0.42374 \\
    1.12      0.4037 \\
    1.16     0.38167 \\
    1.2     0.35827 \\
    1.24     0.33407 \\
    1.28     0.30957 \\
    1.32     0.28519 \\
    1.36     0.26117 \\
    1.4     0.23715 \\
    1.44     0.21178 \\
    1.48     0.18274 \\
    1.52     0.14734 \\
    1.56     0.10307 \\
    1.6    0.048191 \\
    1.64   -0.017927 \\
    1.68   -0.094715 \\
    1.72    -0.18049 \\
    1.76    -0.27268 \\
    1.8    -0.36818 \\
    1.84    -0.46361 \\
    1.88    -0.55566 \\
    1.92    -0.64133 \\
    1.96    -0.71808 \\
    2    -0.78395 \\
    2.04     -0.8376 \\
    2.08    -0.87831 \\
    2.12    -0.90591 \\
    2.16    -0.92075 \\
    2.2    -0.92355 \\
    2.24    -0.91534 \\
    2.28    -0.89739 \\
    2.32    -0.87107 \\
    2.36    -0.83781 \\
    2.4    -0.79903 \\
    2.44     -0.7561 \\
    2.48    -0.71029 \\
    2.52    -0.66274 \\
    2.56    -0.61446 \\
    2.6    -0.56633 \\
    2.64    -0.51906 \\
    2.68    -0.47325 \\
    2.72    -0.42935 \\
    2.76    -0.38768 \\
    2.8    -0.34823 \\
    2.84    -0.31044 \\
    2.88    -0.27324 \\
    2.92    -0.23524 \\
    2.96    -0.19504 \\
    3     -0.1516 \\
    3.04    -0.10434 \\
    3.08   -0.053307 \\
    3.12  0.00089047 \\
    3.16    0.057182 \\
    3.2     0.11418 \\
    3.24     0.17033 \\
    3.28     0.22408 \\
    3.32     0.27401 \\
    3.36     0.31888 \\
    3.4     0.35776 \\
    3.44     0.38999 \\
    3.48     0.41523 \\
    3.52     0.43342 \\
    3.56     0.44475 \\
    3.6     0.44958 \\
    3.64     0.44847 \\
    3.68     0.44207 \\
    3.72      0.4311 \\
    3.76      0.4163 \\
    3.8     0.39842 \\
    3.84     0.37817 \\
    3.88     0.35623 \\
    3.92     0.33319 \\
    3.96      0.3096 \\
    4     0.28591 \\
    4.04     0.26252 \\
    4.08     0.23975 \\
    4.12     0.21784 \\
    4.16     0.19699 \\
    4.2     0.17733 \\
    4.24     0.15895 \\
    4.28      0.1419 \\
    4.32     0.12619 \\
    4.36     0.11181 \\
    4.4     0.09873 \\
    4.44    0.086889 \\
    4.48    0.076227 \\
    4.52    0.066671 \\
    4.56    0.058146 \\
    4.6     0.05057 \\
    4.64    0.043866 \\
    4.68    0.037953 \\
    4.72    0.032758 \\
    4.76    0.028208 \\
    4.8    0.024235 \\
    4.84    0.020777 \\
    4.88    0.017775 \\
    4.92    0.015177 \\
    4.96    0.012933 \\
    5       0.011 \\
    5.04   0.0093395 \\
    5.08   0.0079157 \\
    5.12   0.0066977 \\
    5.16   0.0056579 \\
    5.2    0.004772 \\
    5.24   0.0040187 \\
    5.28   0.0033793 \\
    5.32   0.0028375 \\
    5.36   0.0023793 \\
    5.4   0.0019924 \\
    5.44   0.0016663 \\
    5.48   0.0013917 \\
    5.52    0.001161 \\
    5.56  0.00096734 \\
  };
  \addlegendentry{Reference Velocity}
 \addplot[line width=0.1mm,color=red]
   table[row sep=crcr, x expr=\thisrow{X}, y expr=\thisrow{Y}*57.2958]{
     X Y \\
    0  0 \\
    0.016  1.9867e-06 \\
    0.032  5.6108e-05 \\
    0.048   0.0003759 \\
    0.064    0.001397 \\
    0.08   0.0037587 \\
    0.096   0.0082425 \\
    0.112    0.015694 \\
    0.128    0.026944 \\
    0.144    0.042737 \\
    0.16    0.063676 \\
    0.176    0.090182 \\
    0.192     0.12247 \\
    0.208     0.16054 \\
    0.224     0.20419 \\
    0.24       0.253 \\
    0.256     0.30639 \\
    0.272     0.36364 \\
    0.288      0.4239 \\
    0.304     0.48622 \\
    0.32     0.54963 \\
    0.336     0.61312 \\
    0.352     0.67568 \\
    0.368     0.73633 \\
    0.384     0.79415 \\
    0.4     0.84828 \\
    0.416     0.89796 \\
    0.432     0.94251 \\
    0.448     0.98135 \\
    0.464       1.014 \\
    0.48      1.0402 \\
    0.496      1.0595 \\
    0.512       1.072 \\
    0.528      1.0774 \\
    0.544       1.076 \\
    0.56      1.0677 \\
    0.576      1.0528 \\
    0.592      1.0317 \\
    0.608      1.0045 \\
    0.624     0.97176 \\
    0.64     0.93384 \\
    0.656     0.89122 \\
    0.672     0.84437 \\
    0.688     0.79379 \\
    0.704     0.73998 \\
    0.72     0.68346 \\
    0.736     0.62472 \\
    0.752     0.56426 \\
    0.768     0.50255 \\
    0.784     0.44006 \\
    0.8     0.37723 \\
    0.816     0.31445 \\
    0.832     0.25213 \\
    0.848     0.19062 \\
    0.864     0.13023 \\
    0.88    0.071281 \\
    0.896    0.014022 \\
    0.912   -0.041308 \\
    0.928   -0.094504 \\
    0.944    -0.14539 \\
    0.96    -0.19383 \\
    0.976    -0.23969 \\
    0.992    -0.28288 \\
    1.008    -0.32334 \\
    1.024    -0.36102 \\
    1.04    -0.39589 \\
    1.056    -0.42796 \\
    1.072    -0.45723 \\
    1.088    -0.48374 \\
    1.104    -0.50753 \\
    1.12    -0.52865 \\
    1.136    -0.54719 \\
    1.152    -0.56321 \\
    1.168    -0.57681 \\
    1.184    -0.58808 \\
    1.2    -0.59712 \\
    1.216    -0.60404 \\
    1.232    -0.60894 \\
    1.248    -0.61194 \\
    1.264    -0.61316 \\
    1.28    -0.61269 \\
    1.296    -0.61067 \\
    1.312    -0.60732 \\
    1.328    -0.60318 \\
    1.344    -0.59926 \\
    1.36    -0.59695 \\
    1.376    -0.59794 \\
    1.392      -0.604 \\
    1.408    -0.61691 \\
    1.424    -0.63822 \\
    1.44    -0.66923 \\
    1.456    -0.71083 \\
    1.472    -0.76353 \\
    1.488    -0.82739 \\
    1.504    -0.90203 \\
    1.52     -0.9867 \\
    1.536     -1.0803 \\
    1.552     -1.1814 \\
    1.568     -1.2883 \\
    1.584     -1.3993 \\
    1.6     -1.5123 \\
    1.616     -1.6255 \\
    1.632     -1.7368 \\
    1.648     -1.8444 \\
    1.664     -1.9464 \\
    1.68     -2.0411 \\
    1.696      -2.127 \\
    1.712     -2.2029 \\
    1.728     -2.2676 \\
    1.744     -2.3202 \\
    1.76     -2.3599 \\
    1.776     -2.3864 \\
    1.792     -2.3994 \\
    1.808     -2.3986 \\
    1.824     -2.3844 \\
    1.84     -2.3569 \\
    1.856     -2.3166 \\
    1.872     -2.2639 \\
    1.888     -2.1997 \\
    1.904     -2.1247 \\
    1.92     -2.0396 \\
    1.936     -1.9456 \\
    1.952     -1.8434 \\
    1.968     -1.7341 \\
    1.984     -1.6187 \\
    2     -1.4982 \\
    2.016     -1.3737 \\
    2.032      -1.246 \\
    2.048     -1.1161 \\
    2.064    -0.98501 \\
    2.08    -0.85351 \\
    2.096    -0.72242 \\
    2.112    -0.59253 \\
    2.128    -0.46451 \\
    2.144    -0.33903 \\
    2.16    -0.21665 \\
    2.176   -0.097906 \\
    2.192    0.016751 \\
    2.208     0.12692 \\
    2.224     0.23227 \\
    2.24      0.3325 \\
    2.256      0.4274 \\
    2.272     0.51678 \\
    2.288     0.60052 \\
    2.304     0.67854 \\
    2.32      0.7508 \\
    2.336      0.8173 \\
    2.352     0.87807 \\
    2.368     0.93318 \\
    2.384     0.98273 \\
    2.4      1.0268 \\
    2.416      1.0657 \\
    2.432      1.0993 \\
    2.448      1.1281 \\
    2.464      1.1521 \\
    2.48      1.1715 \\
    2.496      1.1867 \\
    2.512      1.1977 \\
    2.528      1.2048 \\
    2.544      1.2083 \\
    2.56      1.2084 \\
    2.576      1.2053 \\
    2.592      1.1993 \\
    2.608      1.1905 \\
    2.624      1.1792 \\
    2.64      1.1656 \\
    2.656        1.15 \\
    2.672      1.1324 \\
    2.688      1.1132 \\
    2.704      1.0925 \\
    2.72      1.0705 \\
    2.736      1.0476 \\
    2.752      1.0245 \\
    2.768      1.0018 \\
    2.784     0.98069 \\
    2.8     0.96198 \\
    2.816     0.94666 \\
    2.832     0.93559 \\
    2.848     0.92949 \\
    2.864     0.92886 \\
    2.88     0.93401 \\
    2.896     0.94503 \\
    2.912     0.96176 \\
    2.928     0.98388 \\
    2.944      1.0109 \\
    2.96       1.042 \\
    2.976      1.0765 \\
    2.992      1.1136 \\
    3.008      1.1522 \\
    3.024      1.1913 \\
    3.04        1.23 \\
    3.056      1.2674 \\
    3.072      1.3025 \\
    3.088      1.3345 \\
    3.104      1.3627 \\
    3.12      1.3864 \\
    3.136       1.405 \\
    3.152       1.418 \\
    3.168      1.4252 \\
    3.184      1.4262 \\
    3.2      1.4209 \\
    3.216      1.4093 \\
    3.232      1.3914 \\
    3.248      1.3674 \\
    3.264      1.3373 \\
    3.28      1.3016 \\
    3.296      1.2604 \\
    3.312      1.2143 \\
    3.328      1.1636 \\
    3.344      1.1087 \\
    3.36      1.0502 \\
    3.376     0.98844 \\
    3.392     0.92399 \\
    3.408     0.85733 \\
    3.424     0.78893 \\
    3.44     0.71928 \\
    3.456     0.64884 \\
    3.472     0.57806 \\
    3.488     0.50734 \\
    3.504     0.43709 \\
    3.52     0.36768 \\
    3.536     0.29945 \\
    3.552     0.23271 \\
    3.568     0.16774 \\
    3.584     0.10478 \\
    3.6    0.044069 \\
    3.616   -0.014215 \\
    3.632   -0.069909 \\
    3.648    -0.12288 \\
    3.664    -0.17302 \\
    3.68    -0.22025 \\
    3.696    -0.26452 \\
    3.712    -0.30578 \\
    3.728    -0.34404 \\
    3.744    -0.37929 \\
    3.76    -0.41156 \\
    3.776    -0.44089 \\
    3.792    -0.46733 \\
    3.808    -0.49095 \\
    3.824    -0.51183 \\
    3.84    -0.53006 \\
    3.856    -0.54573 \\
    3.872    -0.55895 \\
    3.888    -0.56982 \\
    3.904    -0.57846 \\
    3.92    -0.58497 \\
    3.936    -0.58949 \\
    3.952    -0.59212 \\
    3.968    -0.59299 \\
    3.984    -0.59222 \\
    4    -0.58992 \\
    4.016     -0.5862 \\
    4.032    -0.58119 \\
    4.048    -0.57499 \\
    4.064    -0.56771 \\
    4.08    -0.55945 \\
    4.096    -0.55031 \\
    4.112    -0.54039 \\
    4.128    -0.52979 \\
    4.144    -0.51858 \\
    4.16    -0.50685 \\
    4.176    -0.49468 \\
    4.192    -0.48214 \\
    4.208     -0.4693 \\
    4.224    -0.45623 \\
    4.24    -0.44298 \\
    4.256    -0.42961 \\
    4.272    -0.41619 \\
    4.288    -0.40274 \\
    4.304    -0.38931 \\
    4.32    -0.37596 \\
    4.336     -0.3627 \\
    4.352    -0.34958 \\
    4.368    -0.33663 \\
    4.384    -0.32387 \\
    4.4    -0.31132 \\
    4.416      -0.299 \\
    4.432    -0.28694 \\
    4.448    -0.27514 \\
    4.464    -0.26363 \\
    4.48    -0.25241 \\
    4.496    -0.24149 \\
    4.512    -0.23088 \\
    4.528    -0.22058 \\
    4.544    -0.21059 \\
    4.56    -0.20093 \\
    4.576    -0.19159 \\
    4.592    -0.18256 \\
    4.608    -0.17386 \\
    4.624    -0.16547 \\
    4.64    -0.15739 \\
    4.656    -0.14962 \\
    4.672    -0.14216 \\
    4.688    -0.13499 \\
    4.704    -0.12811 \\
    4.72    -0.12153 \\
    4.736    -0.11522 \\
    4.752    -0.10918 \\
    4.768    -0.10341 \\
    4.784   -0.097897 \\
    4.8   -0.092634 \\
    4.816   -0.087613 \\
    4.832   -0.082827 \\
    4.848   -0.078266 \\
    4.864   -0.073925 \\
    4.88   -0.069794 \\
    4.896   -0.065866 \\
    4.912   -0.062134 \\
    4.928   -0.058589 \\
    4.944   -0.055224 \\
    4.96   -0.052032 \\
    4.976   -0.049006 \\
    4.992   -0.046138 \\
    5.008   -0.043422 \\
    5.024   -0.040851 \\
    5.04   -0.038418 \\
    5.056   -0.036117 \\
    5.072   -0.033942 \\
    5.088   -0.031887 \\
    5.104   -0.029947 \\
    5.12   -0.028115 \\
    5.136   -0.026387 \\
    5.152   -0.024756 \\
    5.168    -0.02322 \\
    5.184   -0.021772 \\
    5.2   -0.020407 \\
    5.216   -0.019123 \\
    5.232   -0.017914 \\
    5.248   -0.016777 \\
    5.264   -0.015707 \\
    5.28   -0.014701 \\
    5.296   -0.013756 \\
    5.312   -0.012868 \\
    5.328   -0.012034 \\
    5.344   -0.011251 \\
    5.36   -0.010516 \\
    5.376   -0.009827 \\
    5.392  -0.0091805 \\
    5.408  -0.0085743 \\
    5.424  -0.0080061 \\
    5.44  -0.0074737 \\
    5.456  -0.0069751 \\
    5.472  -0.0065081 \\
    5.488  -0.0060709 \\
    5.504  -0.0056618 \\
    5.52   -0.005279 \\
    5.536   -0.004921 \\
    5.552  -0.0045862 \\
    5.568  -0.0042733 \\
    5.584  -0.0039808 \\
  };
 \addlegendentry{Reference Acceleration}
\end{axis}
\end{tikzpicture}

%% file: figs/sim_error_plot_v4.tex
\begin{tikzpicture}
\begin{axis}[%
width=0.951\fwidth,
height=\fheight,
at={(0\fwidth,0\fheight)},
scale only axis,
xmin=0,
xmax=5.5,
xtick={0,1,...,5},
xlabel style={font=\color{white!15!black}},
xlabel={Time in $\mathrm{s}$},
ymin= -0.015,
ymax= 0.085,
ytick={0, 0.02, ..., 0.08},
scaled ticks=false,
ylabel style={font=\color{white!15!black}},
ylabel=Angular error $q_{\Delta, 1}$  \\ in $\mathrm{deg}$,
ylabel style={align=center},
axis background/.style={fill=white},
xmajorgrids,
ymajorgrids,
legend cell align={left},
legend pos=north east
]
\addplot[line width=0.1mm, color=blue]
  table[row sep=crcr, x expr=\thisrow{X}, y expr=\thisrow{Y}*57.2958]{
  X Y \\
     0  0 \\
    0.004 -8.1793e-16 \\
    0.008  -1.022e-13 \\
    0.012 -1.7039e-12 \\
    0.016 -1.2452e-11 \\
    0.02 -5.7891e-11 \\
    0.024 -2.0217e-10 \\
    0.028 -5.7943e-10 \\
    0.032 -1.4369e-09 \\
    0.036 -3.1898e-09 \\
    0.04  -6.489e-09 \\
    0.044 -1.2298e-08 \\
    0.048 -2.1984e-08 \\
    0.052 -3.7411e-08 \\
    0.056 -6.1048e-08 \\
    0.06 -9.6076e-08 \\
    0.064  -1.465e-07 \\
    0.068 -2.1726e-07 \\
    0.072 -3.1434e-07 \\
    0.076 -4.4489e-07 \\
    0.08 -6.1729e-07 \\
    0.084 -8.4128e-07 \\
    0.088  -1.128e-06 \\
    0.092 -1.4901e-06 \\
    0.096 -1.9418e-06 \\
    0.1 -2.4989e-06 \\
    0.104 -3.1785e-06 \\
    0.108 -3.9998e-06 \\
    0.112 -4.9832e-06 \\
    0.116 -6.1507e-06 \\
    0.12 -7.5258e-06 \\
    0.124 -9.1332e-06 \\
    0.128 -1.0999e-05 \\
    0.132 -1.3151e-05 \\
    0.136 -1.5616e-05 \\
    0.14 -1.8423e-05 \\
    0.144 -2.1601e-05 \\
    0.148 -2.5179e-05 \\
    0.152 -2.9185e-05 \\
    0.156 -3.3645e-05 \\
    0.16 -3.8584e-05 \\
    0.164 -4.4021e-05 \\
    0.168  -4.997e-05 \\
    0.172 -5.6436e-05 \\
    0.176 -6.3407e-05 \\
    0.18  -7.085e-05 \\
    0.184 -7.8693e-05 \\
    0.188 -8.6795e-05 \\
    0.192 -9.4876e-05 \\
    0.196 -0.00010232 \\
    0.2 -0.00010764 \\
    0.204 -0.00010855 \\
    0.208  -0.0001036 \\
    0.212 -9.3086e-05 \\
    0.216 -7.8886e-05 \\
    0.22 -6.3874e-05 \\
    0.224 -5.0624e-05 \\
    0.228 -4.0608e-05 \\
    0.232 -3.4545e-05 \\
    0.236 -3.2735e-05 \\
    0.24 -3.5132e-05 \\
    0.244 -4.1156e-05 \\
    0.248 -4.9054e-05 \\
    0.252 -5.5703e-05 \\
    0.256 -5.8053e-05 \\
    0.26  -5.435e-05 \\
    0.264 -4.4675e-05 \\
    0.268 -3.0923e-05 \\
    0.272 -1.6222e-05 \\
    0.276 -3.9285e-06 \\
    0.28  3.5276e-06 \\
    0.284  5.1667e-06 \\
    0.288  1.3875e-06 \\
    0.292  -5.831e-06 \\
    0.296 -1.3278e-05 \\
    0.3 -1.7636e-05 \\
    0.304 -1.6556e-05 \\
    0.308 -9.3742e-06 \\
    0.312  2.6871e-06 \\
    0.316  1.6907e-05 \\
    0.32  2.9905e-05 \\
    0.324  3.8693e-05 \\
    0.328  4.1586e-05 \\
    0.332  3.8691e-05 \\
    0.336  3.1873e-05 \\
    0.34  2.4201e-05 \\
    0.344  1.8984e-05 \\
    0.348  1.8761e-05 \\
    0.352  2.4516e-05 \\
    0.356  3.5388e-05 \\
    0.36  4.8945e-05 \\
    0.364  6.1933e-05 \\
    0.368  7.1285e-05 \\
    0.372   7.506e-05 \\
    0.376  7.3033e-05 \\
    0.38  6.6758e-05 \\
    0.384  5.9082e-05 \\
    0.388  5.3266e-05 \\
    0.392  5.1981e-05 \\
    0.396  5.6492e-05 \\
    0.4  6.6276e-05 \\
    0.404  7.9188e-05 \\
    0.408  9.2132e-05 \\
    0.412  0.00010201 \\
    0.416  0.00010669 \\
    0.42  0.00010563 \\
    0.424  0.00010005 \\
    0.428  9.2575e-05 \\
    0.432  8.6369e-05 \\
    0.436  8.4207e-05 \\
    0.44  8.7608e-05 \\
    0.444  9.6371e-05 \\
    0.448  0.00010865 \\
    0.452  0.00012153 \\
    0.456  0.00013193 \\
    0.46  0.00013752 \\
    0.464  0.00013749 \\
    0.468  0.00013274 \\
    0.472  0.00012563 \\
    0.476  0.00011921 \\
    0.48  0.00011634 \\
    0.484  0.00011875 \\
    0.488  0.00012654 \\
    0.492  0.00013819 \\
    0.496  0.00015097 \\
    0.5  0.00016182 \\
    0.504  0.00016832 \\
    0.508  0.00016935 \\
    0.512  0.00016552 \\
    0.516  0.00015891 \\
    0.52  0.00015244 \\
    0.524  0.00014899 \\
    0.528  0.00015049 \\
    0.532  0.00015736 \\
    0.536  0.00016835 \\
    0.54  0.00018096 \\
    0.544  0.00019222 \\
    0.548  0.00019956 \\
    0.552  0.00020166 \\
    0.556  0.00019879 \\
    0.56  0.00019277 \\
    0.564  0.00018636 \\
    0.568  0.00018245 \\
    0.572  0.00018314 \\
    0.576  0.00018909 \\
    0.58  0.00019939 \\
    0.584  0.00021176 \\
    0.588  0.00022331 \\
    0.592  0.00023143 \\
    0.596  0.00023454 \\
    0.6  0.00023266 \\
    0.604   0.0002273 \\
    0.608  0.00022105 \\
    0.612  0.00021677 \\
    0.616  0.00021669 \\
    0.62  0.00022176 \\
    0.624  0.00023131 \\
    0.628  0.00024335 \\
    0.632  0.00025508 \\
    0.636  0.00026387 \\
    0.64  0.00026794 \\
    0.644  0.00026703 \\
    0.648  0.00026237 \\
    0.652  0.00025635 \\
    0.656  0.00025177 \\
    0.66  0.00025098 \\
    0.664  0.00025515 \\
    0.668  0.00026391 \\
    0.672  0.00027551 \\
    0.676  0.00028732 \\
    0.68  0.00029665 \\
    0.684   0.0003016 \\
    0.688  0.00030162 \\
    0.692  0.00029768 \\
    0.696  0.00029194 \\
    0.7  0.00028713 \\
    0.704  0.00028568 \\
    0.708  0.00028895 \\
    0.712  0.00029687 \\
    0.716  0.00030794 \\
    0.72  0.00031968 \\
    0.724  0.00032944 \\
    0.728  0.00033517 \\
    0.732  0.00033608 \\
    0.736  0.00033285 \\
    0.74  0.00032743 \\
    0.744  0.00032245 \\
    0.748  0.00032037 \\
    0.752  0.00032275 \\
    0.756  0.00032979 \\
    0.76  0.00034023 \\
    0.764  0.00035179 \\
    0.768  0.00036184 \\
    0.772  0.00036825 \\
    0.776  0.00036998 \\
    0.78  0.00036746 \\
    0.784   0.0003624 \\
    0.788   0.0003573 \\
    0.792  0.00035464 \\
    0.796  0.00035614 \\
    0.8  0.00036225 \\
    0.804  0.00037198 \\
    0.808  0.00038324 \\
    0.812  0.00039347 \\
    0.816  0.00040045 \\
    0.82  0.00040294 \\
    0.824  0.00040111 \\
    0.828  0.00039644 \\
    0.832  0.00039126 \\
    0.836  0.00038807 \\
    0.84  0.00038872 \\
    0.844  0.00039388 \\
    0.848  0.00040283 \\
    0.852  0.00041368 \\
    0.856  0.00042397 \\
    0.86  0.00043141 \\
    0.864  0.00043459 \\
    0.868  0.00043344 \\
    0.872  0.00042917 \\
    0.876  0.00042397 \\
    0.88  0.00042031 \\
    0.884  0.00042013 \\
    0.888  0.00042434 \\
    0.892  0.00043245 \\
    0.896   0.0004428 \\
    0.9  0.00045303 \\
    0.904  0.00046083 \\
    0.908  0.00046464 \\
    0.912  0.00046413 \\
    0.916   0.0004603 \\
    0.92  0.00045514 \\
    0.924  0.00045105 \\
    0.928   0.0004501 \\
    0.932  0.00045335 \\
    0.936  0.00046059 \\
    0.94  0.00047036 \\
    0.944  0.00048043 \\
    0.948   0.0004885 \\
    0.952  0.00049287 \\
    0.956  0.00049298 \\
    0.96  0.00048959 \\
    0.964  0.00048452 \\
    0.968  0.00048009 \\
    0.972  0.00047841 \\
    0.976  0.00048073 \\
    0.98  0.00048706 \\
    0.984  0.00049618 \\
    0.988  0.00050601 \\
    0.992  0.00051425 \\
    0.996  0.00051911 \\
    1  0.00051983 \\
    1.004  0.00051691 \\
    1.008  0.00051199 \\
    1.012  0.00050728 \\
    1.016  0.00050493 \\
    1.02  0.00050634 \\
    1.024  0.00051176 \\
    1.028  0.00052018 \\
    1.032  0.00052968 \\
    1.036  0.00053802 \\
    1.04  0.00054331 \\
    1.044   0.0005446 \\
    1.048  0.00054218 \\
    1.052  0.00053746 \\
    1.056  0.00053254 \\
    1.06  0.00052959 \\
    1.064  0.00053013 \\
    1.068  0.00053462 \\
    1.072   0.0005423 \\
    1.076  0.00055141 \\
    1.08  0.00055976 \\
    1.084  0.00056543 \\
    1.088   0.0005673 \\
    1.092  0.00056542 \\
    1.096  0.00056097 \\
    1.1  0.00055593 \\
    1.104  0.00055242 \\
    1.108   0.0005521 \\
    1.112  0.00055565 \\
    1.116  0.00056251 \\
    1.12  0.00057115 \\
    1.124  0.00057948 \\
    1.128  0.00058551 \\
    1.132  0.00058801 \\
    1.136  0.00058682 \\
    1.14  0.00058281 \\
    1.144  0.00057773 \\
    1.148  0.00057366 \\
    1.152  0.00057239 \\
    1.156  0.00057482 \\
    1.16   0.0005807 \\
    1.164  0.00058874 \\
    1.168  0.00059698 \\
    1.172  0.00060346 \\
    1.176  0.00060686 \\
    1.18  0.00060676 \\
    1.184  0.00060363 \\
    1.188  0.00059879 \\
    1.192  0.00059412 \\
    1.196  0.00059153 \\
    1.2  0.00059227 \\
    1.204  0.00059657 \\
    1.208  0.00060358 \\
    1.212   0.0006116 \\
    1.216  0.00061874 \\
    1.22  0.00062353 \\
    1.224  0.00062533 \\
    1.228  0.00062412 \\
    1.232   0.0006205 \\
    1.236  0.00061567 \\
    1.24  0.00061138 \\
    1.244  0.00060932 \\
    1.248  0.00061054 \\
    1.252  0.00061507 \\
    1.256  0.00062193 \\
    1.26   0.0006295 \\
    1.264  0.00063613 \\
    1.268  0.00064075 \\
    1.272  0.00064288 \\
    1.276  0.00064248 \\
    1.28  0.00063986 \\
    1.284  0.00063569 \\
    1.288  0.00063113 \\
    1.292   0.0006277 \\
    1.296  0.00062675 \\
    1.3  0.00062892 \\
    1.304  0.00063392 \\
    1.308  0.00064063 \\
    1.312   0.0006476 \\
    1.316  0.00065363 \\
    1.32  0.00065802 \\
    1.324  0.00066048 \\
    1.328  0.00066093 \\
    1.332  0.00065952 \\
    1.336  0.00065656 \\
    1.34  0.00065262 \\
    1.344  0.00064856 \\
    1.348  0.00064551 \\
    1.352  0.00064456 \\
    1.356  0.00064629 \\
    1.36  0.00065052 \\
    1.364  0.00065644 \\
    1.368  0.00066295 \\
    1.372  0.00066918 \\
    1.376  0.00067457 \\
    1.38   0.0006788 \\
    1.384  0.00068175 \\
    1.388  0.00068338 \\
    1.392  0.00068377 \\
    1.396   0.0006831 \\
    1.4  0.00068158 \\
    1.404  0.00067952 \\
    1.408   0.0006773 \\
    1.412  0.00067532 \\
    1.416  0.00067402 \\
    1.42  0.00067382 \\
    1.424    0.000675 \\
    1.428  0.00067769 \\
    1.432  0.00068187 \\
    1.436  0.00068738 \\
    1.44  0.00069406 \\
    1.444  0.00070175 \\
    1.448  0.00071032 \\
    1.452  0.00071971 \\
    1.456  0.00072988 \\
    1.46  0.00074084 \\
    1.464  0.00075262 \\
    1.468  0.00076528 \\
    1.472  0.00077889 \\
    1.476  0.00079354 \\
    1.48  0.00080934 \\
    1.484  0.00082638 \\
    1.488  0.00084477 \\
    1.492  0.00086461 \\
    1.496  0.00088597 \\
    1.5  0.00090887 \\
    1.504  0.00093326 \\
    1.508  0.00095893 \\
    1.512  0.00098507 \\
    1.516   0.0010083 \\
    1.52   0.0010224 \\
    1.524   0.0010234 \\
    1.528   0.0010112 \\
    1.532  0.00098987 \\
    1.536  0.00096614 \\
    1.54  0.00094563 \\
    1.544  0.00093066 \\
    1.548  0.00092237 \\
    1.552  0.00092154 \\
    1.556  0.00092871 \\
    1.56  0.00094414 \\
    1.564  0.00096761 \\
    1.568  0.00099677 \\
    1.572   0.0010243 \\
    1.576   0.0010422 \\
    1.58   0.0010452 \\
    1.584   0.0010329 \\
    1.588   0.0010095 \\
    1.592  0.00098261 \\
    1.596  0.00096076 \\
    1.6  0.00095031 \\
    1.604  0.00095354 \\
    1.608  0.00097039 \\
    1.612  0.00099723 \\
    1.616   0.0010263 \\
    1.62   0.0010491 \\
    1.624   0.0010592 \\
    1.628   0.0010547 \\
    1.632   0.0010404 \\
    1.636    0.001024 \\
    1.64   0.0010122 \\
    1.644   0.0010097 \\
    1.648   0.0010182 \\
    1.652   0.0010356 \\
    1.656   0.0010573 \\
    1.66   0.0010775 \\
    1.664   0.0010906 \\
    1.668   0.0010933 \\
    1.672   0.0010856 \\
    1.676   0.0010705 \\
    1.68   0.0010535 \\
    1.684   0.0010404 \\
    1.688   0.0010358 \\
    1.692    0.001042 \\
    1.696   0.0010574 \\
    1.7   0.0010781 \\
    1.704   0.0010982 \\
    1.708   0.0011122 \\
    1.712   0.0011164 \\
    1.716   0.0011103 \\
    1.72   0.0010962 \\
    1.724   0.0010792 \\
    1.728    0.001065 \\
    1.732   0.0010585 \\
    1.736   0.0010623 \\
    1.74   0.0010756 \\
    1.744   0.0010948 \\
    1.748   0.0011146 \\
    1.752   0.0011292 \\
    1.756   0.0011348 \\
    1.76     0.00113 \\
    1.764    0.001117 \\
    1.768      0.0011 \\
    1.772   0.0010849 \\
    1.776   0.0010766 \\
    1.78    0.001078 \\
    1.784    0.001089 \\
    1.788   0.0011066 \\
    1.792   0.0011257 \\
    1.796   0.0011407 \\
    1.8   0.0011474 \\
    1.804    0.001144 \\
    1.808   0.0011319 \\
    1.812   0.0011152 \\
    1.816   0.0010992 \\
    1.82   0.0010891 \\
    1.824   0.0010882 \\
    1.828    0.001097 \\
    1.832   0.0011128 \\
    1.836   0.0011311 \\
    1.84   0.0011462 \\
    1.844   0.0011539 \\
    1.848   0.0011518 \\
    1.852   0.0011408 \\
    1.856   0.0011244 \\
    1.86   0.0011078 \\
    1.864   0.0010961 \\
    1.868   0.0010931 \\
    1.872   0.0010995 \\
    1.876   0.0011135 \\
    1.88   0.0011307 \\
    1.884   0.0011459 \\
    1.888   0.0011544 \\
    1.892   0.0011536 \\
    1.896   0.0011437 \\
    1.9   0.0011279 \\
    1.904   0.0011108 \\
    1.908   0.0010978 \\
    1.912   0.0010927 \\
    1.916   0.0010969 \\
    1.92   0.0011091 \\
    1.924   0.0011252 \\
    1.928   0.0011403 \\
    1.932   0.0011495 \\
    1.936   0.0011499 \\
    1.94   0.0011413 \\
    1.944   0.0011262 \\
    1.948    0.001109 \\
    1.952   0.0010949 \\
    1.956   0.0010879 \\
    1.96   0.0010901 \\
    1.964   0.0011004 \\
    1.968   0.0011154 \\
    1.972   0.0011302 \\
    1.976   0.0011401 \\
    1.98   0.0011418 \\
    1.984   0.0011345 \\
    1.988   0.0011202 \\
    1.992   0.0011031 \\
    1.996   0.0010881 \\
    2   0.0010796 \\
    2.004   0.0010798 \\
    2.008   0.0010883 \\
    2.012   0.0011021 \\
    2.016   0.0011167 \\
    2.02   0.0011272 \\
    2.024   0.0011301 \\
    2.028   0.0011242 \\
    2.032   0.0011109 \\
    2.036   0.0010941 \\
    2.04   0.0010786 \\
    2.044   0.0010687 \\
    2.048   0.0010671 \\
    2.052   0.0010739 \\
    2.056   0.0010865 \\
    2.06   0.0011007 \\
    2.064   0.0011118 \\
    2.068   0.0011159 \\
    2.072   0.0011114 \\
    2.076   0.0010994 \\
    2.08   0.0010831 \\
    2.084   0.0010673 \\
    2.088   0.0010562 \\
    2.092    0.001053 \\
    2.096   0.0010582 \\
    2.1   0.0010696 \\
    2.104   0.0010834 \\
    2.108   0.0010949 \\
    2.112   0.0011002 \\
    2.116   0.0010972 \\
    2.12   0.0010865 \\
    2.124    0.001071 \\
    2.128    0.001055 \\
    2.132   0.0010431 \\
    2.136   0.0010385 \\
    2.14    0.001042 \\
    2.144   0.0010523 \\
    2.148   0.0010656 \\
    2.152   0.0010775 \\
    2.156   0.0010839 \\
    2.16   0.0010824 \\
    2.164   0.0010732 \\
    2.168   0.0010586 \\
    2.172   0.0010427 \\
    2.176   0.0010301 \\
    2.18   0.0010242 \\
    2.184   0.0010263 \\
    2.188   0.0010354 \\
    2.192   0.0010482 \\
    2.196   0.0010604 \\
    2.2   0.0010678 \\
    2.204   0.0010678 \\
    2.208     0.00106 \\
    2.212   0.0010465 \\
    2.216    0.001031 \\
    2.22   0.0010179 \\
    2.224   0.0010109 \\
    2.228   0.0010116 \\
    2.232   0.0010195 \\
    2.236   0.0010317 \\
    2.24    0.001044 \\
    2.244   0.0010524 \\
    2.248   0.0010539 \\
    2.252   0.0010476 \\
    2.256   0.0010353 \\
    2.26   0.0010203 \\
    2.264    0.001007 \\
    2.268  0.00099897 \\
    2.272  0.00099842 \\
    2.276   0.0010051 \\
    2.28   0.0010166 \\
    2.284   0.0010291 \\
    2.288   0.0010383 \\
    2.292   0.0010411 \\
    2.296   0.0010364 \\
    2.3   0.0010254 \\
    2.304    0.001011 \\
    2.308  0.00099757 \\
    2.312  0.00098877 \\
    2.316  0.00098702 \\
    2.32  0.00099254 \\
    2.324   0.0010033 \\
    2.328   0.0010157 \\
    2.332   0.0010256 \\
    2.336   0.0010298 \\
    2.34   0.0010266 \\
    2.344   0.0010169 \\
    2.348   0.0010034 \\
    2.352  0.00098993 \\
    2.356  0.00098046 \\
    2.36  0.00097761 \\
    2.364  0.00098196 \\
    2.368  0.00099192 \\
    2.372   0.0010042 \\
    2.376   0.0010147 \\
    2.38     0.00102 \\
    2.384   0.0010184 \\
    2.388     0.00101 \\
    2.392  0.00099738 \\
    2.396  0.00098412 \\
    2.4  0.00097412 \\
    2.404  0.00097024 \\
    2.408  0.00097344 \\
    2.412  0.00098252 \\
    2.416  0.00099449 \\
    2.42   0.0010055 \\
    2.424   0.0010119 \\
    2.428   0.0010117 \\
    2.432   0.0010047 \\
    2.436  0.00099308 \\
    2.44  0.00098012 \\
    2.444  0.00096972 \\
    2.448  0.00096488 \\
    2.452  0.00096692 \\
    2.456  0.00097503 \\
    2.46  0.00098658 \\
    2.464  0.00099789 \\
    2.468   0.0010053 \\
    2.472   0.0010066 \\
    2.476   0.0010011 \\
    2.48  0.00099047 \\
    2.484  0.00097794 \\
    2.488  0.00096721 \\
    2.492   0.0009614 \\
    2.496    0.000962 \\
    2.5  0.00096869 \\
    2.504  0.00097949 \\
    2.508  0.00099104 \\
    2.512  0.00099971 \\
    2.516   0.0010028 \\
    2.52  0.00099927 \\
    2.524  0.00099019 \\
    2.528  0.00097831 \\
    2.532  0.00096719 \\
    2.536  0.00095994 \\
    2.54  0.00095808 \\
    2.544  0.00096179 \\
    2.548  0.00097025 \\
    2.552  0.00098141 \\
    2.556  0.00099199 \\
    2.56  0.00099872 \\
    2.564  0.00099953 \\
    2.568  0.00099415 \\
    2.572  0.00098425 \\
    2.576  0.00097283 \\
    2.58  0.00096312 \\
    2.584  0.00095711 \\
    2.588  0.00095545 \\
    2.592  0.00095816 \\
    2.596  0.00096483 \\
    2.6   0.0009745 \\
    2.604  0.00098517 \\
    2.608  0.00099389 \\
    2.612  0.00099804 \\
    2.616  0.00099635 \\
    2.62   0.0009894 \\
    2.624  0.00097932 \\
    2.628    0.000969 \\
    2.632  0.00096056 \\
    2.636  0.00095496 \\
    2.64  0.00095251 \\
    2.644  0.00095326 \\
    2.648  0.00095704 \\
    2.652   0.0009635 \\
    2.656  0.00097203 \\
    2.66  0.00098158 \\
    2.664  0.00099032 \\
    2.668  0.00099597 \\
    2.672  0.00099688 \\
    2.676  0.00099285 \\
    2.68  0.00098514 \\
    2.684   0.0009759 \\
    2.688  0.00096699 \\
    2.692  0.00095937 \\
    2.696  0.00095355 \\
    2.7  0.00094976 \\
    2.704  0.00094809 \\
    2.708  0.00094853 \\
    2.712  0.00095097 \\
    2.716  0.00095525 \\
    2.72  0.00096109 \\
    2.724  0.00096814 \\
    2.728  0.00097591 \\
    2.732   0.0009837 \\
    2.736  0.00099053 \\
    2.74  0.00099516 \\
    2.744  0.00099652 \\
    2.748  0.00099431 \\
    2.752  0.00098917 \\
    2.756  0.00098227 \\
    2.76  0.00097465 \\
    2.764  0.00096704 \\
    2.768  0.00095992 \\
    2.772  0.00095358 \\
    2.776  0.00094825 \\
    2.78  0.00094403 \\
    2.784  0.00094101 \\
    2.788  0.00093919 \\
    2.792  0.00093858 \\
    2.796   0.0009391 \\
    2.8  0.00094068 \\
    2.804   0.0009432 \\
    2.808  0.00094651 \\
    2.812  0.00095044 \\
    2.816  0.00095479 \\
    2.82  0.00095935 \\
    2.824  0.00096389 \\
    2.828  0.00096815 \\
    2.832  0.00097187 \\
    2.836  0.00097483 \\
    2.84  0.00097678 \\
    2.844  0.00097755 \\
    2.848  0.00097701 \\
    2.852  0.00097509 \\
    2.856  0.00097179 \\
    2.86  0.00096715 \\
    2.864  0.00096124 \\
    2.868  0.00095415 \\
    2.872  0.00094596 \\
    2.876  0.00093675 \\
    2.88  0.00092661 \\
    2.884   0.0009156 \\
    2.888  0.00090377 \\
    2.892  0.00089117 \\
    2.896  0.00087785 \\
    2.9  0.00086385 \\
    2.904  0.00084919 \\
    2.908  0.00083393 \\
    2.912  0.00081812 \\
    2.916  0.00080181 \\
    2.92  0.00078511 \\
    2.924  0.00076815 \\
    2.928  0.00075116 \\
    2.932  0.00073449 \\
    2.936  0.00071892 \\
    2.94  0.00070632 \\
    2.944  0.00069995 \\
    2.948  0.00070196 \\
    2.952  0.00071199 \\
    2.956  0.00072727 \\
    2.96  0.00074405 \\
    2.964  0.00075993 \\
    2.968  0.00077366 \\
    2.972  0.00078445 \\
    2.976   0.0007917 \\
    2.98  0.00079498 \\
    2.984  0.00079396 \\
    2.988  0.00078843 \\
    2.992  0.00077833 \\
    2.996  0.00076379 \\
    3  0.00074541 \\
    3.004  0.00072552 \\
    3.008  0.00070947 \\
    3.012  0.00070235 \\
    3.016  0.00070654 \\
    3.02  0.00072089 \\
    3.024  0.00074117 \\
    3.028  0.00076139 \\
    3.032  0.00077668 \\
    3.036  0.00078499 \\
    3.04  0.00078556 \\
    3.044  0.00077817 \\
    3.048  0.00076329 \\
    3.052  0.00074337 \\
    3.056  0.00072413 \\
    3.06  0.00071166 \\
    3.064  0.00070995 \\
    3.068  0.00071966 \\
    3.072  0.00073796 \\
    3.076  0.00075939 \\
    3.08  0.00077756 \\
    3.084  0.00078754 \\
    3.088  0.00078733 \\
    3.092  0.00077699 \\
    3.096  0.00075911 \\
    3.1   0.0007392 \\
    3.104  0.00072356 \\
    3.108  0.00071714 \\
    3.112  0.00072175 \\
    3.116  0.00073496 \\
    3.12  0.00075054 \\
    3.124   0.0007616 \\
    3.128  0.00076337 \\
    3.132  0.00075441 \\
    3.136  0.00073683 \\
    3.14  0.00071552 \\
    3.144  0.00069663 \\
    3.148  0.00068564 \\
    3.152  0.00068563 \\
    3.156  0.00069632 \\
    3.16  0.00071415 \\
    3.164  0.00073338 \\
    3.168  0.00074791 \\
    3.172  0.00075307 \\
    3.176  0.00074713 \\
    3.18  0.00073177 \\
    3.184  0.00071156 \\
    3.188  0.00069256 \\
    3.192  0.00068042 \\
    3.196  0.00067868 \\
    3.2  0.00068766 \\
    3.204  0.00070441 \\
    3.208  0.00072361 \\
    3.212  0.00073921 \\
    3.216  0.00074632 \\
    3.22  0.00074266 \\
    3.224  0.00072931 \\
    3.228  0.00071029 \\
    3.232  0.00069138 \\
    3.236  0.00067829 \\
    3.24  0.00067493 \\
    3.244  0.00068221 \\
    3.248  0.00069778 \\
    3.252  0.00071676 \\
    3.256  0.00073324 \\
    3.26  0.00074213 \\
    3.264   0.0007407 \\
    3.268  0.00072939 \\
    3.272   0.0007117 \\
    3.276  0.00069307 \\
    3.28  0.00067922 \\
    3.284  0.00067436 \\
    3.288  0.00067995 \\
    3.292  0.00069424 \\
    3.296   0.0007128 \\
    3.3  0.00072995 \\
    3.304  0.00074044 \\
    3.308  0.00074112 \\
    3.312  0.00073187 \\
    3.316  0.00071561 \\
    3.32  0.00069743 \\
    3.324  0.00068297 \\
    3.328  0.00067672 \\
    3.332  0.00068063 \\
    3.336  0.00069353 \\
    3.34   0.0007115 \\
    3.344  0.00072908 \\
    3.348  0.00074097 \\
    3.352  0.00074363 \\
    3.356   0.0007364 \\
    3.36  0.00072165 \\
    3.364  0.00070405 \\
    3.368  0.00068914 \\
    3.372  0.00068161 \\
    3.376  0.00068386 \\
    3.38  0.00069528 \\
    3.384  0.00071247 \\
    3.388  0.00073025 \\
    3.392  0.00074331 \\
    3.396  0.00074781 \\
    3.4  0.00074255 \\
    3.404  0.00072935 \\
    3.408  0.00071246 \\
    3.412  0.00069724 \\
    3.416  0.00068853 \\
    3.42  0.00068914 \\
    3.424  0.00069901 \\
    3.428  0.00071525 \\
    3.432  0.00073301 \\
    3.436  0.00074701 \\
    3.44  0.00075317 \\
    3.444   0.0007498 \\
    3.448  0.00073818 \\
    3.452   0.0007221 \\
    3.456  0.00070671 \\
    3.46  0.00069694 \\
    3.464  0.00069595 \\
    3.468  0.00070421 \\
    3.472  0.00071934 \\
    3.476  0.00073686 \\
    3.48  0.00075159 \\
    3.484  0.00075923 \\
    3.488  0.00075766 \\
    3.492  0.00074762 \\
    3.496  0.00073244 \\
    3.5  0.00071702 \\
    3.504  0.00070631 \\
    3.508  0.00070378 \\
    3.512  0.00071039 \\
    3.516  0.00072428 \\
    3.52  0.00074137 \\
    3.524   0.0007566 \\
    3.528  0.00076553 \\
    3.532  0.00076566 \\
    3.536   0.0007572 \\
    3.54  0.00074301 \\
    3.544  0.00072768 \\
    3.548  0.00071615 \\
    3.552  0.00071215 \\
    3.556   0.0007171 \\
    3.56  0.00072964 \\
    3.564  0.00074611 \\
    3.568  0.00076164 \\
    3.572  0.00077169 \\
    3.576   0.0007734 \\
    3.58   0.0007665 \\
    3.584  0.00075337 \\
    3.588  0.00073826 \\
    3.592  0.00072605 \\
    3.596  0.00072067 \\
    3.6  0.00072396 \\
    3.604  0.00073506 \\
    3.608  0.00075075 \\
    3.612  0.00076639 \\
    3.616  0.00077737 \\
    3.62  0.00078055 \\
    3.624  0.00077519 \\
    3.628  0.00076317 \\
    3.632  0.00074841 \\
    3.636  0.00073565 \\
    3.64    0.000729 \\
    3.644  0.00073066 \\
    3.648  0.00074027 \\
    3.652  0.00075504 \\
    3.656  0.00077059 \\
    3.66  0.00078234 \\
    3.664  0.00078687 \\
    3.668  0.00078301 \\
    3.672  0.00077217 \\
    3.676  0.00075788 \\
    3.68  0.00074472 \\
    3.684   0.0007369 \\
    3.688  0.00073696 \\
    3.692  0.00074503 \\
    3.696  0.00075876 \\
    3.7  0.00077407 \\
    3.704  0.00078641 \\
    3.708  0.00079219 \\
    3.712  0.00078984 \\
    3.716  0.00078028 \\
    3.72  0.00076659 \\
    3.724  0.00075315 \\
    3.728  0.00074424 \\
    3.732  0.00074272 \\
    3.736  0.00074916 \\
    3.74   0.0007617 \\
    3.744   0.0007766 \\
    3.748  0.00078941 \\
    3.752   0.0007964 \\
    3.756  0.00079572 \\
    3.76  0.00078774 \\
    3.764  0.00077492 \\
    3.768  0.00076132 \\
    3.772  0.00075123 \\
    3.776  0.00074786 \\
    3.78  0.00075234 \\
    3.784  0.00076336 \\
    3.788  0.00077764 \\
    3.792  0.00079088 \\
    3.796  0.00079927 \\
    3.8  0.00080079 \\
    3.804  0.00079528 \\
    3.808  0.00078416 \\
    3.812  0.00077068 \\
    3.816  0.00075905 \\
    3.82  0.00075292 \\
    3.824  0.00075425 \\
    3.828  0.00076268 \\
    3.832  0.00077569 \\
    3.836  0.00078935 \\
    3.84  0.00079972 \\
    3.844  0.00080445 \\
    3.848  0.00080294 \\
    3.852  0.00079554 \\
    3.856  0.00078377 \\
    3.86   0.0007708 \\
    3.864  0.00076062 \\
    3.868  0.00075641 \\
    3.872   0.0007595 \\
    3.876  0.00076897 \\
    3.88  0.00078196 \\
    3.884  0.00079458 \\
    3.888  0.00080364 \\
    3.892  0.00080776 \\
    3.896  0.00080662 \\
    3.9  0.00080048 \\
    3.904  0.00079019 \\
    3.908  0.00077777 \\
    3.912  0.00076667 \\
    3.916  0.00076029 \\
    3.92  0.00076062 \\
    3.924  0.00076756 \\
    3.928    0.000779 \\
    3.932   0.0007915 \\
    3.936  0.00080188 \\
    3.94   0.0008086 \\
    3.944  0.00081112 \\
    3.948   0.0008094 \\
    3.952  0.00080365 \\
    3.956  0.00079447 \\
    3.96  0.00078306 \\
    3.964  0.00077188 \\
    3.968  0.00076414 \\
    3.972  0.00076221 \\
    3.976  0.00076667 \\
    3.98  0.00077614 \\
    3.984  0.00078777 \\
    3.988   0.0007986 \\
    3.992  0.00080699 \\
    3.996  0.00081228 \\
    4  0.00081423 \\
    4.004  0.00081282 \\
    4.008  0.00080822 \\
    4.012  0.00080078 \\
    4.016  0.00079113 \\
    4.02  0.00078041 \\
    4.024  0.00077076 \\
    4.028  0.00076484 \\
    4.032  0.00076442 \\
    4.036   0.0007696 \\
    4.04  0.00077878 \\
    4.044  0.00078937 \\
    4.048  0.00079925 \\
    4.052  0.00080735 \\
    4.056  0.00081315 \\
    4.06  0.00081642 \\
    4.064  0.00081708 \\
    4.068  0.00081517 \\
    4.072  0.00081083 \\
    4.076  0.00080433 \\
    4.08  0.00079607 \\
    4.084  0.00078667 \\
    4.088  0.00077718 \\
    4.092  0.00076934 \\
    4.096  0.00076517 \\
    4.1  0.00076586 \\
    4.104  0.00077116 \\
    4.108   0.0007795 \\
    4.112  0.00078888 \\
    4.116  0.00079789 \\
    4.12  0.00080575 \\
    4.124  0.00081201 \\
    4.128  0.00081644 \\
    4.132   0.0008189 \\
    4.136  0.00081933 \\
    4.14  0.00081778 \\
    4.144  0.00081433 \\
    4.148  0.00080915 \\
    4.152  0.00080249 \\
    4.156   0.0007947 \\
    4.16  0.00078624 \\
    4.164  0.00077785 \\
    4.168  0.00077056 \\
    4.172  0.00076581 \\
    4.176  0.00076476 \\
    4.18  0.00076765 \\
    4.184  0.00077366 \\
    4.188  0.00078139 \\
    4.192  0.00078961 \\
    4.196  0.00079752 \\
    4.2   0.0008046 \\
    4.204  0.00081054 \\
    4.208  0.00081509 \\
    4.212  0.00081813 \\
    4.216  0.00081957 \\
    4.22   0.0008194 \\
    4.224  0.00081764 \\
    4.228  0.00081436 \\
    4.232  0.00080971 \\
    4.236  0.00080385 \\
    4.24  0.00079702 \\
    4.244  0.00078955 \\
    4.248  0.00078186 \\
    4.252   0.0007745 \\
    4.256  0.00076826 \\
    4.26  0.00076407 \\
    4.264  0.00076276 \\
    4.268  0.00076461 \\
    4.272  0.00076912 \\
    4.276  0.00077538 \\
    4.28  0.00078247 \\
    4.284  0.00078971 \\
    4.288  0.00079658 \\
    4.292  0.00080274 \\
    4.296  0.00080793 \\
    4.3  0.00081196 \\
    4.304  0.00081471 \\
    4.308   0.0008161 \\
    4.312   0.0008161 \\
    4.316  0.00081472 \\
    4.32  0.00081201 \\
    4.324  0.00080806 \\
    4.328  0.00080299 \\
    4.332    0.000797 \\
    4.336   0.0007903 \\
    4.34  0.00078319 \\
    4.344  0.00077606 \\
    4.348   0.0007694 \\
    4.352  0.00076383 \\
    4.356  0.00076008 \\
    4.36  0.00075879 \\
    4.364  0.00076016 \\
    4.368  0.00076388 \\
    4.372  0.00076926 \\
    4.376  0.00077556 \\
    4.38  0.00078215 \\
    4.384  0.00078859 \\
    4.388   0.0007945 \\
    4.392  0.00079962 \\
    4.396  0.00080376 \\
    4.4  0.00080676 \\
    4.404  0.00080854 \\
    4.408  0.00080904 \\
    4.412  0.00080825 \\
    4.416  0.00080619 \\
    4.42  0.00080292 \\
    4.424  0.00079855 \\
    4.428  0.00079322 \\
    4.432  0.00078711 \\
    4.436  0.00078048 \\
    4.44  0.00077364 \\
    4.444  0.00076699 \\
    4.448  0.00076106 \\
    4.452  0.00075646 \\
    4.456  0.00075386 \\
    4.46  0.00075368 \\
    4.464  0.00075593 \\
    4.468  0.00076013 \\
    4.472  0.00076562 \\
    4.476  0.00077173 \\
    4.48  0.00077794 \\
    4.484  0.00078383 \\
    4.488   0.0007891 \\
    4.492  0.00079351 \\
    4.496  0.00079688 \\
    4.5  0.00079909 \\
    4.504  0.00080008 \\
    4.508   0.0007998 \\
    4.512  0.00079826 \\
    4.516  0.00079551 \\
    4.52  0.00079163 \\
    4.524  0.00078675 \\
    4.528  0.00078105 \\
    4.532  0.00077476 \\
    4.536  0.00076819 \\
    4.54  0.00076173 \\
    4.544  0.00075588 \\
    4.548  0.00075127 \\
    4.552  0.00074856 \\
    4.556  0.00074821 \\
    4.56  0.00075026 \\
    4.564  0.00075427 \\
    4.568  0.00075956 \\
    4.572  0.00076547 \\
    4.576  0.00077145 \\
    4.58  0.00077708 \\
    4.584  0.00078205 \\
    4.588  0.00078612 \\
    4.592  0.00078912 \\
    4.596  0.00079094 \\
    4.6  0.00079151 \\
    4.604   0.0007908 \\
    4.608  0.00078884 \\
    4.612   0.0007857 \\
    4.616  0.00078147 \\
    4.62  0.00077632 \\
    4.624  0.00077046 \\
    4.628  0.00076419 \\
    4.632  0.00075787 \\
    4.636  0.00075201 \\
    4.64  0.00074722 \\
    4.644  0.00074419 \\
    4.648  0.00074346 \\
    4.652  0.00074515 \\
    4.656  0.00074888 \\
    4.66  0.00075397 \\
    4.664  0.00075971 \\
    4.668  0.00076553 \\
    4.672  0.00077098 \\
    4.676  0.00077575 \\
    4.68  0.00077957 \\
    4.684   0.0007823 \\
    4.688  0.00078381 \\
    4.692  0.00078405 \\
    4.696  0.00078301 \\
    4.7  0.00078072 \\
    4.704  0.00077727 \\
    4.708   0.0007728 \\
    4.712  0.00076749 \\
    4.716  0.00076161 \\
    4.72  0.00075551 \\
    4.724  0.00074965 \\
    4.728  0.00074464 \\
    4.732  0.00074118 \\
    4.736  0.00073987 \\
    4.74  0.00074098 \\
    4.744  0.00074422 \\
    4.748  0.00074897 \\
    4.752   0.0007545 \\
    4.756  0.00076018 \\
    4.76  0.00076553 \\
    4.764  0.00077022 \\
    4.768  0.00077398 \\
    4.772  0.00077664 \\
    4.776  0.00077809 \\
    4.78  0.00077828 \\
    4.784   0.0007772 \\
    4.788  0.00077489 \\
    4.792  0.00077147 \\
    4.796  0.00076707 \\
    4.8  0.00076192 \\
    4.804  0.00075631 \\
    4.808  0.00075061 \\
    4.812  0.00074533 \\
    4.816  0.00074109 \\
    4.82  0.00073853 \\
    4.824  0.00073814 \\
    4.828  0.00073998 \\
    4.832  0.00074367 \\
    4.836  0.00074855 \\
    4.84  0.00075394 \\
    4.844  0.00075931 \\
    4.848  0.00076423 \\
    4.852  0.00076842 \\
    4.856  0.00077166 \\
    4.86  0.00077381 \\
    4.864  0.00077479 \\
    4.868  0.00077459 \\
    4.872  0.00077322 \\
    4.876  0.00077077 \\
    4.88  0.00076737 \\
    4.884  0.00076318 \\
    4.888  0.00075845 \\
    4.892  0.00075347 \\
    4.896  0.00074861 \\
    4.9  0.00074432 \\
    4.904   0.0007411 \\
    4.908  0.00073943 \\
    4.912  0.00073959 \\
    4.916  0.00074155 \\
    4.92  0.00074499 \\
    4.924  0.00074939 \\
    4.928  0.00075424 \\
    4.932   0.0007591 \\
    4.936  0.00076362 \\
    4.94  0.00076756 \\
    4.944  0.00077073 \\
    4.948  0.00077301 \\
    4.952  0.00077434 \\
    4.956  0.00077469 \\
    4.96  0.00077411 \\
    4.964  0.00077264 \\
    4.968   0.0007704 \\
    4.972  0.00076751 \\
    4.976  0.00076414 \\
    4.98  0.00076048 \\
    4.984  0.00075677 \\
    4.988  0.00075324 \\
    4.992  0.00075017 \\
    4.996  0.00074781 \\
    5  0.00074639 \\
    5.004  0.00074607 \\
    5.008  0.00074689 \\
    5.012  0.00074878 \\
    5.016  0.00075153 \\
    5.02  0.00075489 \\
    5.024  0.00075861 \\
    5.028  0.00076243 \\
    5.032  0.00076614 \\
    5.036  0.00076956 \\
    5.04  0.00077257 \\
    5.044  0.00077506 \\
    5.048  0.00077698 \\
    5.052   0.0007783 \\
    5.056    0.000779 \\
    5.06  0.00077912 \\
    5.064  0.00077869 \\
    5.068  0.00077777 \\
    5.072  0.00077643 \\
    5.076  0.00077476 \\
    5.08  0.00077285 \\
    5.084   0.0007708 \\
    5.088   0.0007687 \\
    5.092  0.00076667 \\
    5.096   0.0007648 \\
    5.1  0.00076318 \\
    5.104  0.00076189 \\
    5.108  0.00076099 \\
    5.112  0.00076052 \\
    5.116  0.00076051 \\
    5.12  0.00076096 \\
    5.124  0.00076185 \\
    5.128  0.00076313 \\
    5.132  0.00076474 \\
    5.136  0.00076662 \\
    5.14  0.00076869 \\
    5.144  0.00077088 \\
    5.148  0.00077311 \\
    5.152  0.00077531 \\
    5.156  0.00077743 \\
    5.16  0.00077941 \\
    5.164  0.00078121 \\
    5.168   0.0007828 \\
    5.172  0.00078415 \\
    5.176  0.00078526 \\
    5.18  0.00078611 \\
    5.184   0.0007867 \\
    5.188  0.00078706 \\
    5.192  0.00078718 \\
    5.196  0.00078708 \\
    5.2  0.00078679 \\
    5.204  0.00078634 \\
    5.208  0.00078575 \\
    5.212  0.00078504 \\
    5.216  0.00078425 \\
    5.22  0.00078341 \\
    5.224  0.00078254 \\
    5.228  0.00078167 \\
    5.232  0.00078083 \\
    5.236  0.00078003 \\
    5.24  0.00077931 \\
    5.244  0.00077866 \\
    5.248  0.00077812 \\
    5.252  0.00077769 \\
    5.256  0.00077737 \\
    5.26  0.00077718 \\
    5.264  0.00077711 \\
    5.268  0.00077716 \\
    5.272  0.00077733 \\
    5.276  0.00077761 \\
    5.28    0.000778 \\
    5.284  0.00077847 \\
    5.288  0.00077902 \\
    5.292  0.00077964 \\
    5.296  0.00078031 \\
    5.3  0.00078101 \\
    5.304  0.00078174 \\
    5.308  0.00078248 \\
    5.312  0.00078321 \\
    5.316  0.00078392 \\
    5.32  0.00078461 \\
    5.324  0.00078525 \\
    5.328  0.00078585 \\
    5.332  0.00078639 \\
    5.336  0.00078686 \\
    5.34  0.00078727 \\
    5.344   0.0007876 \\
    5.348  0.00078786 \\
    5.352  0.00078804 \\
    5.356  0.00078814 \\
    5.36  0.00078817 \\
    5.364  0.00078813 \\
    5.368  0.00078802 \\
    5.372  0.00078785 \\
    5.376  0.00078761 \\
    5.38  0.00078732 \\
    5.384  0.00078697 \\
    5.388  0.00078659 \\
    5.392  0.00078617 \\
    5.396  0.00078571 \\
    5.4  0.00078524 \\
    5.404  0.00078474 \\
    5.408  0.00078424 \\
    5.412  0.00078373 \\
    5.416  0.00078323 \\
    5.42  0.00078273 \\
    5.424  0.00078224 \\
    5.428  0.00078177 \\
    5.432  0.00078133 \\
    5.436  0.00078091 \\
    5.44  0.00078051 \\
    5.444  0.00078015 \\
    5.448  0.00077982 \\
    5.452  0.00077953 \\
    5.456  0.00077927 \\
    5.46  0.00077904 \\
    5.464  0.00077885 \\
    5.468   0.0007787 \\
    5.472  0.00077858 \\
    5.476  0.00077849 \\
    5.48  0.00077843 \\
    5.484   0.0007784 \\
    5.488  0.00077839 \\
    5.492  0.00077841 \\
    5.496  0.00077844 \\
    5.5  0.00077849 \\
    5.504  0.00077855 \\
    5.508  0.00077862 \\
    5.512   0.0007787 \\
    5.516  0.00077878 \\
    5.52  0.00077886 \\
    5.524  0.00077894 \\
    5.528  0.00077901 \\
    5.532  0.00077908 \\
    5.536  0.00077913 \\
    5.54  0.00077917 \\
    5.544   0.0007792 \\
    5.548  0.00077921 \\
    5.552  0.00077921 \\
    5.556  0.00077918 \\
    5.56  0.00077914 \\
    5.564  0.00077908 \\
    5.568    0.000779 \\
    5.572   0.0007789 \\
    5.576  0.00077878 \\
    5.58  0.00077864 \\
    5.584  0.00077849 \\
    5.588  0.00077832 \\
    5.592  0.00077814 \\
    5.596  0.00077794 \\
};
\addlegendentry{Error R-FF}
\addplot[line width=0.1mm, color=red]
    table[row sep=crcr]{
    0  0 \\
    5.5  0 \\
};
\addlegendentry{Error FB-FF}
\addplot[line width=0.3mm, color=black]
  table[row sep=crcr]{
  0 0.00859 \\
  5.5 0.00859 \\
  };
\addlegendentry{Backlash Angle}
\addplot[line width=0.3mm, color=black]
  table[row sep=crcr]{
  0 -0.00859 \\
  5.5 -0.00859 \\
  };
  \fill[gray,fill opacity=0.25] (axis cs:0,-0.00859) rectangle (axis cs:5.5, 0.00859);
\end{axis}
\end{tikzpicture}

%% file: figs/sim_elastic_torque_plot_v3.tex
\begin{tikzpicture}
\begin{axis}[%
width=0.951\fwidth,
height=\fheight,
at={(0\fwidth,0\fheight)},
scale only axis,
xmin=0,
xmax=5.5,
xtick={0,1,...,5},
xlabel style={font=\color{white!15!black}},
xlabel={Time in $\mathrm{s}$},
ymin=-5,
ymax=6,
ytick={4,2,...,-4},
ylabel style={font=\color{white!15!black}},
ylabel={Elastic torque $\tau_E$ in $\mathrm{kNm}$},
axis background/.style={fill=white},
xmajorgrids,
ymajorgrids,
legend cell align={left},
legend pos=north east
]
\addplot[line width=0.1mm] [color=blue]
  table[row sep=crcr, x expr=\thisrow{X}, y expr=\thisrow{Y}*0.001]{
  X Y \\
    0 -1.1435e-14 \\
    0.004  8.6318e-10 \\
    0.008  9.8301e-08 \\
    0.012  1.6208e-06 \\
    0.016  1.1858e-05 \\
    0.02  5.5397e-05 \\
    0.024  0.00019466 \\
    0.028  0.00056176 \\
    0.032   0.0014031 \\
    0.036   0.0031381 \\
    0.04   0.0064319 \\
    0.044    0.012283 \\
    0.048    0.022127 \\
    0.052    0.037946 \\
    0.056    0.062402 \\
    0.06    0.098974 \\
    0.064      0.1521 \\
    0.068     0.22734 \\
    0.072     0.33152 \\
    0.076     0.47291 \\
    0.08     0.66139 \\
    0.084     0.90856 \\
    0.088       1.228 \\
    0.092      1.6352 \\
    0.096      2.1481 \\
    0.1      2.7867 \\
    0.104      3.5737 \\
    0.108      4.5343 \\
    0.112      5.6964 \\
    0.116      7.0909 \\
    0.12      8.7516 \\
    0.124       10.716 \\
    0.128      13.0251 \\
    0.132      15.7244 \\
    0.136      18.8646 \\
    0.14      22.5031 \\
    0.144      26.7053 \\
    0.148      31.5485 \\
    0.152      37.1259 \\
    0.156      43.5527 \\
    0.16      50.9777 \\
    0.164      59.5988 \\
    0.168       69.694 \\
    0.172      81.6683 \\
    0.176      96.1536 \\
    0.18        114.21 \\
    0.184      137.8084 \\
    0.188      171.1955 \\
    0.192      225.7616 \\
    0.196      337.5682 \\
    0.2      520.7759 \\
    0.204      654.4256 \\
    0.208      681.5646 \\
    0.212      602.1954 \\
    0.216      452.9787 \\
    0.22      314.9901 \\
    0.224      244.6694 \\
    0.228      214.9283 \\
    0.232      207.5821 \\
    0.236      219.8512 \\
    0.24      261.6146 \\
    0.244      373.8978 \\
    0.248      584.5247 \\
    0.252      791.4893 \\
    0.256      921.9046 \\
    0.26      941.7833 \\
    0.264      851.5632 \\
    0.268      685.3303 \\
    0.272      502.2185 \\
    0.276      374.4246 \\
    0.28       333.511 \\
    0.284      368.9646 \\
    0.288      500.2102 \\
    0.292      709.2856 \\
    0.296      925.8932 \\
    0.3      1086.8818 \\
    0.304      1148.7914 \\
    0.308      1098.9283 \\
    0.312      959.0272 \\
    0.316      778.4505 \\
    0.32      619.1775 \\
    0.324      536.3944 \\
    0.328      559.0003 \\
    0.332       686.752 \\
    0.336      887.4074 \\
    0.34      1104.5685 \\
    0.344      1277.5573 \\
    0.348      1359.6968 \\
    0.352       1332.448 \\
    0.356      1210.9632 \\
    0.36      1039.2746 \\
    0.364      876.6409 \\
    0.368      779.2769 \\
    0.372       783.021 \\
    0.376      892.5507 \\
    0.38      1080.0726 \\
    0.384      1293.8822 \\
    0.388      1474.4291 \\
    0.392      1572.6303 \\
    0.396      1564.8858 \\
    0.4      1460.2038 \\
    0.404      1297.2817 \\
    0.408      1132.4819 \\
    0.412       1022.439 \\
    0.416       1006.661 \\
    0.42      1095.4735 \\
    0.424       1266.981 \\
    0.428      1473.9301 \\
    0.432      1658.3353 \\
    0.436      1769.3317 \\
    0.44      1778.7587 \\
    0.444      1689.7006 \\
    0.448      1535.3928 \\
    0.452      1368.8748 \\
    0.456      1246.6079 \\
    0.46      1211.1188 \\
    0.464      1278.0005 \\
    0.468      1431.2455 \\
    0.472      1628.3032 \\
    0.476      1813.2659 \\
    0.48      1934.0966 \\
    0.484      1958.5889 \\
    0.488       1884.164 \\
    0.492      1738.5252 \\
    0.496      1571.0092 \\
    0.5      1437.3351 \\
    0.504      1382.4622 \\
    0.508      1426.8315 \\
    0.512      1560.1993 \\
    0.516      1744.9224 \\
    0.52      1927.6418 \\
    0.524      2055.7398 \\
    0.528      2093.4937 \\
    0.532      2032.9677 \\
    0.536      1896.3209 \\
    0.54      1728.8648 \\
    0.544       1585.041 \\
    0.548      1511.6609 \\
    0.552      1533.5646 \\
    0.556      1646.0934 \\
    0.56      1816.6552 \\
    0.564      1994.8608 \\
    0.568      2128.0797 \\
    0.572      2177.6123 \\
    0.576      2130.5102 \\
    0.58      2003.4347 \\
    0.584      1837.4051 \\
    0.588      1685.0903 \\
    0.592      1594.5819 \\
    0.596      1594.6475 \\
    0.6      1685.9853 \\
    0.604      1841.1391 \\
    0.608      2013.0593 \\
    0.612       2149.642 \\
    0.616      2209.7535 \\
    0.62      2175.8123 \\
    0.624      2059.0815 \\
    0.628      1896.0784 \\
    0.632       1737.246 \\
    0.636      1631.3998 \\
    0.64      1610.7454 \\
    0.644      1681.0663 \\
    0.648      1820.0699 \\
    0.652      1984.3633 \\
    0.656      2122.8769 \\
    0.66      2192.5838 \\
    0.664      2171.6774 \\
    0.668      2066.1722 \\
    0.672      1907.9306 \\
    0.676      1744.7618 \\
    0.68      1625.6679 \\
    0.684      1585.7929 \\
    0.688      1635.6947 \\
    0.692      1758.2179 \\
    0.696      1913.8901 \\
    0.7      2053.1481 \\
    0.704      2131.6055 \\
    0.708      2123.6617 \\
    0.712       2030.279 \\
    0.716      1878.5672 \\
    0.72      1713.3402 \\
    0.724      1583.2754 \\
    0.728      1525.9489 \\
    0.732      1556.3491 \\
    0.736      1662.3799 \\
    0.74      1808.7005 \\
    0.744       1947.687 \\
    0.748      2034.1152 \\
    0.752      2039.0393 \\
    0.756      1958.6079 \\
    0.76       1815.134 \\
    0.764      1650.1244 \\
    0.768      1511.4474 \\
    0.772      1438.6061 \\
    0.776      1450.6471 \\
    0.78      1540.4075 \\
    0.784      1676.8364 \\
    0.788      1814.6414 \\
    0.792      1908.2651 \\
    0.796      1925.8774 \\
    0.8      1859.0899 \\
    0.804      1725.4256 \\
    0.808      1562.8212 \\
    0.812      1417.8852 \\
    0.816      1331.5471 \\
    0.82      1326.5159 \\
    0.824      1400.3923 \\
    0.828      1526.5214 \\
    0.832       1662.292 \\
    0.836      1762.2959 \\
    0.84      1792.2875 \\
    0.844        1739.65 \\
    0.848      1617.1731 \\
    0.852        1459.01 \\
    0.856      1310.0933 \\
    0.86      1212.2888 \\
    0.864      1191.5546 \\
    0.868      1250.0441 \\
    0.872      1365.5554 \\
    0.876      1498.4624 \\
    0.88      1603.9647 \\
    0.884      1645.8705 \\
    0.888      1607.6727 \\
    0.892      1497.5291 \\
    0.896      1345.6458 \\
    0.9      1194.9014 \\
    0.904      1087.6226 \\
    0.908      1052.5899 \\
    0.912      1096.2625 \\
    0.916      1200.9001 \\
    0.92      1330.1212 \\
    0.924      1440.1619 \\
    0.928      1493.3513 \\
    0.932      1469.6524 \\
    0.936      1372.7361 \\
    0.94      1228.7456 \\
    0.944      1078.1674 \\
    0.948      963.3334 \\
    0.952      915.4113 \\
    0.956      944.8831 \\
    0.96      1038.4358 \\
    0.964       1163.149 \\
    0.968      1276.6925 \\
    0.972      1340.3745 \\
    0.976      1331.0069 \\
    0.98      1247.9557 \\
    0.984      1113.2332 \\
    0.988       964.637 \\
    0.992       844.076 \\
    0.996       784.666 \\
    1       800.573 \\
    1.004      882.8454 \\
    1.008       1002.223 \\
    1.012      1118.1715 \\
    1.016      1191.4198 \\
    1.02      1196.0161 \\
    1.024      1127.2338 \\
    1.028      1002.9259 \\
    1.032      857.9472 \\
    1.036      733.4045 \\
    1.04      663.9401 \\
    1.044      666.8861 \\
    1.048      737.5497 \\
    1.052      850.6835 \\
    1.056      967.8937 \\
    1.06      1049.7274 \\
    1.064       1067.837 \\
    1.068      1013.6096 \\
    1.072      900.7192 \\
    1.076      760.8544 \\
    1.08      634.0545 \\
    1.084      556.2046 \\
    1.088      546.6733 \\
    1.092      604.5706 \\
    1.096      709.9593 \\
    1.1       827.178 \\
    1.104      916.8319 \\
    1.108      948.4042 \\
    1.112      909.4462 \\
    1.116      809.2782 \\
    1.12      676.1233 \\
    1.124      548.9071 \\
    1.128        465.01 \\
    1.132      443.4482 \\
    1.136      484.6106 \\
    1.14      577.7526 \\
    1.144      692.8244 \\
    1.148      790.3778 \\
    1.152      837.3639 \\
    1.156      816.8536 \\
    1.16       732.859 \\
    1.164       609.208 \\
    1.168      483.6164 \\
    1.172      396.0913 \\
    1.176      363.1804 \\
    1.18      380.5227 \\
    1.184      447.5648 \\
    1.188      552.1244 \\
    1.192      657.7548 \\
    1.196      726.7875 \\
    1.2      735.2209 \\
    1.204       678.439 \\
    1.208      572.4497 \\
    1.212      451.2144 \\
    1.216      358.1886 \\
    1.22      312.4799 \\
    1.224      304.3477 \\
    1.228      329.0383 \\
    1.232      392.1486 \\
    1.236      488.8447 \\
    1.24      584.4491 \\
    1.244        641.68 \\
    1.248      639.6638 \\
    1.252      577.4558 \\
    1.256      474.6843 \\
    1.26      370.4857 \\
    1.264      301.9208 \\
    1.268      268.9601 \\
    1.272      259.7847 \\
    1.276      269.9492 \\
    1.28      302.1109 \\
    1.284      363.4154 \\
    1.288       448.882 \\
    1.292      525.6456 \\
    1.296       560.727 \\
    1.3      539.9242 \\
    1.304      469.6092 \\
    1.308      377.6476 \\
    1.312      302.9957 \\
    1.316      258.5159 \\
    1.32      235.3134 \\
    1.324      226.1916 \\
    1.328      228.2543 \\
    1.332      241.4527 \\
    1.336      268.1954 \\
    1.34      312.8739 \\
    1.344      375.3385 \\
    1.348       436.639 \\
    1.352      468.0156 \\
    1.356        454.22 \\
    1.36      400.2399 \\
    1.364      330.8701 \\
    1.368      273.8988 \\
    1.372      235.8777 \\
    1.376      211.6521 \\
    1.38       196.636 \\
    1.384      188.2016 \\
    1.388      184.9227 \\
    1.392      186.0246 \\
    1.396       191.093 \\
    1.4       199.874 \\
    1.404      212.0739 \\
    1.408       227.007 \\
    1.412      243.0157 \\
    1.416      256.8608 \\
    1.42      264.0269 \\
    1.424      260.8677 \\
    1.428      247.2525 \\
    1.432      226.6746 \\
    1.436      203.5345 \\
    1.44      180.7845 \\
    1.444      159.6504 \\
    1.448       140.346 \\
    1.452      122.6648 \\
    1.456       106.281 \\
    1.46      90.8609 \\
    1.464      76.0981 \\
    1.468      61.7091 \\
    1.472      47.4274 \\
    1.476      32.9843 \\
    1.48       18.086 \\
    1.484      2.3805 \\
    1.488     -14.5992 \\
    1.492     -33.5392 \\
    1.496     -55.5693 \\
    1.5     -82.8331 \\
    1.504     -120.3184 \\
    1.508      -184.834 \\
    1.512     -393.6624 \\
    1.516     -821.8679 \\
    1.52     -1099.4803 \\
    1.524      -1137.436 \\
    1.528     -934.3955 \\
    1.532     -564.3121 \\
    1.536     -254.6022 \\
    1.54      -166.991 \\
    1.544     -132.5515 \\
    1.548     -118.0834 \\
    1.552     -117.6305 \\
    1.556     -132.1345 \\
    1.56     -171.5724 \\
    1.564      -306.167 \\
    1.568     -804.0508 \\
    1.572     -1325.9068 \\
    1.576     -1672.8966 \\
    1.58     -1750.8406 \\
    1.584     -1548.7736 \\
    1.588     -1141.8941 \\
    1.592     -668.2902 \\
    1.596     -323.6134 \\
    1.6     -240.5699 \\
    1.604     -268.2161 \\
    1.608     -503.8031 \\
    1.612     -1007.5564 \\
    1.616     -1558.2009 \\
    1.62     -1997.2261 \\
    1.624     -2206.6003 \\
    1.628       -2151.62 \\
    1.632     -1905.8739 \\
    1.636     -1607.6412 \\
    1.64     -1388.4597 \\
    1.644     -1337.5467 \\
    1.648     -1485.2785 \\
    1.652     -1798.8865 \\
    1.656     -2194.2198 \\
    1.66     -2562.2002 \\
    1.664     -2802.5459 \\
    1.668     -2854.5621 \\
    1.672     -2715.6032 \\
    1.676     -2441.6137 \\
    1.68     -2129.6723 \\
    1.684     -1887.9835 \\
    1.688     -1802.5904 \\
    1.692     -1911.0456 \\
    1.696     -2191.0941 \\
    1.7     -2567.7738 \\
    1.704     -2936.6703 \\
    1.708     -3196.1088 \\
    1.712     -3278.3581 \\
    1.716     -3170.2675 \\
    1.72     -2917.0487 \\
    1.724     -2608.1177 \\
    1.728     -2349.4274 \\
    1.732     -2230.8491 \\
    1.736     -2298.6418 \\
    1.74     -2541.4408 \\
    1.744     -2894.0068 \\
    1.748     -3257.5014 \\
    1.752     -3529.9886 \\
    1.756     -3637.7476 \\
    1.76     -3557.7777 \\
    1.764     -3324.6246 \\
    1.768     -3019.5041 \\
    1.772     -2745.1438 \\
    1.776     -2594.1343 \\
    1.78     -2620.5386 \\
    1.784     -2823.4634 \\
    1.788     -3147.5823 \\
    1.792     -3500.3724 \\
    1.796      -3780.695 \\
    1.8     -3909.8909 \\
    1.804     -3855.8309 \\
    1.808     -3642.5667 \\
    1.812     -3342.6857 \\
    1.816     -3054.7959 \\
    1.82     -2873.1241 \\
    1.824     -2858.5986 \\
    1.828     -3020.2904 \\
    1.832     -3312.8646 \\
    1.836     -3650.7504 \\
    1.84     -3934.6036 \\
    1.844     -4081.8762 \\
    1.848     -4052.0891 \\
    1.852       -3859.07 \\
    1.856     -3566.4581 \\
    1.86     -3267.9335 \\
    1.864     -3058.3119 \\
    1.868     -3004.4231 \\
    1.872     -3124.7174 \\
    1.876     -3383.8229 \\
    1.88      -3703.655 \\
    1.884     -3987.5926 \\
    1.888     -4150.2357 \\
    1.892      -4143.581 \\
    1.896     -3971.5929 \\
    1.9     -3688.7501 \\
    1.904     -3383.0862 \\
    1.908     -3149.0033 \\
    1.912     -3058.2559 \\
    1.916     -3138.0211 \\
    1.92     -3362.7602 \\
    1.924      -3662.305 \\
    1.928     -3943.6129 \\
    1.932     -4119.4467 \\
    1.936       -4135.14 \\
    1.94     -3985.2385 \\
    1.944      -3714.952 \\
    1.948     -3406.0418 \\
    1.952     -3151.5483 \\
    1.956     -3027.1706 \\
    1.96     -3068.1029 \\
    1.964     -3258.4157 \\
    1.968      -3536.185 \\
    1.972     -3812.7229 \\
    1.976     -3999.9384 \\
    1.98       -4037.39 \\
    1.984      -3910.723 \\
    1.988     -3655.8706 \\
    1.992     -3347.7916 \\
    1.996     -3077.2764 \\
    2     -2922.9973 \\
    2.004     -2927.4086 \\
    2.008     -3083.8784 \\
    2.012     -3338.9546 \\
    2.016     -3608.9972 \\
    2.02     -3806.0053 \\
    2.024     -3864.6654 \\
    2.028     -3762.3073 \\
    2.032     -3525.6751 \\
    2.036     -3222.4908 \\
    2.04      -2940.474 \\
    2.044     -2760.3164 \\
    2.048      -2730.938 \\
    2.052     -2854.6136 \\
    2.056     -3086.4923 \\
    2.06     -3348.5897 \\
    2.064     -3553.8921 \\
    2.068     -3633.1224 \\
    2.072     -3555.9365 \\
    2.076     -3340.0669 \\
    2.08     -3045.6606 \\
    2.084     -2756.6178 \\
    2.088     -2554.7217 \\
    2.092     -2494.5373 \\
    2.096     -2586.7832 \\
    2.1     -2795.2447 \\
    2.104     -3048.1134 \\
    2.108     -3260.2052 \\
    2.112     -3359.1844 \\
    2.116     -3307.7213 \\
    2.12     -3114.7989 \\
    2.124     -2832.7452 \\
    2.128     -2540.9683 \\
    2.132     -2321.4505 \\
    2.136     -2233.5637 \\
    2.14     -2295.9437 \\
    2.144     -2480.9573 \\
    2.148     -2723.4032 \\
    2.152     -2940.7097 \\
    2.156     -3058.3743 \\
    2.16     -3032.8085 \\
    2.164     -2864.5842 \\
    2.168     -2598.0588 \\
    2.172     -2307.5556 \\
    2.176     -2074.4053 \\
    2.18     -1961.9408 \\
    2.184     -1996.1338 \\
    2.188     -2157.7925 \\
    2.192     -2388.6653 \\
    2.196     -2609.5106 \\
    2.2     -2744.5293 \\
    2.204     -2744.6288 \\
    2.208     -2602.3788 \\
    2.212      -2354.104 \\
    2.216     -2068.5313 \\
    2.22     -1825.5379 \\
    2.224     -1691.5733 \\
    2.228     -1699.3169 \\
    2.232     -1837.7993 \\
    2.236     -2055.9733 \\
    2.24      -2278.575 \\
    2.244       -2429.35 \\
    2.248     -2454.4761 \\
    2.252     -2338.9911 \\
    2.256     -2111.2104 \\
    2.26     -1833.8357 \\
    2.264     -1584.5437 \\
    2.268     -1432.0655 \\
    2.272     -1415.1229 \\
    2.276     -1530.6761 \\
    2.28     -1735.0503 \\
    2.284     -1957.5353 \\
    2.288     -2122.2253 \\
    2.292     -2171.3536 \\
    2.296     -2082.9505 \\
    2.3     -1877.4268 \\
    2.304     -1611.1132 \\
    2.308     -1358.7981 \\
    2.312     -1190.6774 \\
    2.316     -1150.8204 \\
    2.32     -1243.7572 \\
    2.324     -1433.2705 \\
    2.328     -1653.7043 \\
    2.332     -1830.2624 \\
    2.336     -1902.0203 \\
    2.34     -1840.5714 \\
    2.344     -1658.6025 \\
    2.348     -1405.8097 \\
    2.352      -1153.469 \\
    2.356     -972.4514 \\
    2.36     -911.4572 \\
    2.364     -982.1586 \\
    2.368     -1155.8097 \\
    2.372     -1372.2337 \\
    2.376     -1558.4553 \\
    2.38     -1651.1718 \\
    2.384     -1616.1478 \\
    2.388     -1458.5955 \\
    2.392     -1221.3939 \\
    2.396     -971.7505 \\
    2.4     -780.4815 \\
    2.404     -700.2021 \\
    2.408     -749.0295 \\
    2.412     -905.8072 \\
    2.416     -1116.2595 \\
    2.42     -1309.8527 \\
    2.424     -1421.6603 \\
    2.428     -1412.2397 \\
    2.432     -1279.6227 \\
    2.436     -1059.7484 \\
    2.44     -815.2746 \\
    2.444     -616.5749 \\
    2.448     -519.9603 \\
    2.452     -546.2793 \\
    2.456     -683.5056 \\
    2.46     -885.5838 \\
    2.464     -1084.4131 \\
    2.468     -1213.9741 \\
    2.472     -1230.0291 \\
    2.476     -1123.4722 \\
    2.48     -922.9609 \\
    2.484     -686.0894 \\
    2.488     -484.2394 \\
    2.492     -381.0413 \\
    2.496     -381.3145 \\
    2.5     -480.5205 \\
    2.504     -664.7484 \\
    2.508     -867.1321 \\
    2.512     -1018.3341 \\
    2.516     -1067.8991 \\
    2.52       -997.07 \\
    2.524     -824.2337 \\
    2.528      -599.621 \\
    2.532      -399.234 \\
    2.536     -299.7175 \\
    2.54     -277.1551 \\
    2.544     -306.5529 \\
    2.548     -411.1347 \\
    2.552     -595.7699 \\
    2.556     -780.2248 \\
    2.56     -895.9857 \\
    2.564     -903.9427 \\
    2.568     -798.8629 \\
    2.572     -610.7552 \\
    2.576     -404.5806 \\
    2.58     -280.2554 \\
    2.584     -233.3999 \\
    2.588     -221.0682 \\
    2.592     -232.5093 \\
    2.596     -274.6658 \\
    2.6     -377.9636 \\
    2.604     -547.9918 \\
    2.608     -698.2354 \\
    2.612     -766.6596 \\
    2.616     -729.0319 \\
    2.62      -595.225 \\
    2.624     -412.6424 \\
    2.628     -279.6765 \\
    2.632     -220.0849 \\
    2.636     -192.8189 \\
    2.64     -181.7708 \\
    2.644     -182.1987 \\
    2.648     -193.7029 \\
    2.652      -219.589 \\
    2.656     -270.4622 \\
    2.66     -369.4863 \\
    2.664     -503.9466 \\
    2.668     -597.6403 \\
    2.672     -607.1872 \\
    2.676     -527.6184 \\
    2.68     -391.4127 \\
    2.684      -277.513 \\
    2.688     -216.5904 \\
    2.692     -183.0975 \\
    2.696     -163.2223 \\
    2.7      -151.647 \\
    2.704     -146.1091 \\
    2.708      -145.622 \\
    2.712     -149.8925 \\
    2.716     -159.1709 \\
    2.72     -174.3438 \\
    2.724     -197.3235 \\
    2.728     -231.8754 \\
    2.732     -284.2019 \\
    2.736     -355.1848 \\
    2.74     -417.4593 \\
    2.744     -433.1181 \\
    2.748     -392.1186 \\
    2.752     -318.7729 \\
    2.756     -253.2536 \\
    2.76      -208.085 \\
    2.764     -177.2005 \\
    2.768     -155.1414 \\
    2.772     -138.8872 \\
    2.776     -126.7882 \\
    2.78     -117.8897 \\
    2.784     -111.6059 \\
    2.788     -107.5603 \\
    2.792     -105.4959 \\
    2.796     -105.2269 \\
    2.8     -106.6149 \\
    2.804      -109.542 \\
    2.808     -113.9008 \\
    2.812     -119.5784 \\
    2.816     -126.4338 \\
    2.82     -134.2649 \\
    2.824     -142.7668 \\
    2.828     -151.4709 \\
    2.832     -159.6824 \\
    2.836     -166.4707 \\
    2.84     -170.7618 \\
    2.844     -171.6061 \\
    2.848     -168.5124 \\
    2.852     -161.6516 \\
    2.856     -151.7521 \\
    2.86       -139.78 \\
    2.864     -126.6228 \\
    2.868     -112.9348 \\
    2.872     -99.1224 \\
    2.876     -85.3954 \\
    2.88     -71.8289 \\
    2.884     -58.4102 \\
    2.888     -45.0705 \\
    2.892      -31.702 \\
    2.896     -18.1628 \\
    2.9     -4.2734 \\
    2.904      10.1945 \\
    2.908      25.5439 \\
    2.912      42.1984 \\
    2.916      60.7933 \\
    2.92      82.3642 \\
    2.924      108.8024 \\
    2.928      144.1621 \\
    2.932      199.6464 \\
    2.936      318.6128 \\
    2.94      542.8105 \\
    2.944      693.7226 \\
    2.948      693.0604 \\
    2.952      546.6156 \\
    2.956      334.1312 \\
    2.96      216.3763 \\
    2.964      164.2854 \\
    2.968      135.1393 \\
    2.972      117.6106 \\
    2.976      108.0296 \\
    2.98      105.1913 \\
    2.984        109.13 \\
    2.988       121.072 \\
    2.992      144.6135 \\
    2.996       191.468 \\
    3      325.1535 \\
    3.004      682.8696 \\
    3.008      1010.1542 \\
    3.012       1175.589 \\
    3.016      1134.9796 \\
    3.02      907.8197 \\
    3.024      571.7511 \\
    3.028      293.6536 \\
    3.032       205.234 \\
    3.036      179.5641 \\
    3.04      182.7809 \\
    3.044      219.8234 \\
    3.048      362.3037 \\
    3.052      734.8085 \\
    3.056      1118.6509 \\
    3.06      1380.7045 \\
    3.064      1447.1268 \\
    3.068      1304.5517 \\
    3.072      1003.8341 \\
    3.076      644.4442 \\
    3.08      363.3626 \\
    3.084      271.8136 \\
    3.088      286.6143 \\
    3.092      442.1163 \\
    3.096      791.2796 \\
    3.1      1186.1393 \\
    3.104      1504.9855 \\
    3.108      1657.3637 \\
    3.112      1608.1191 \\
    3.116      1395.3946 \\
    3.12      1123.6377 \\
    3.124      919.7569 \\
    3.128      879.9667 \\
    3.132      1037.1096 \\
    3.136      1355.1117 \\
    3.14      1744.5607 \\
    3.144      2092.3998 \\
    3.148      2297.6668 \\
    3.152      2303.0592 \\
    3.156      2112.6636 \\
    3.16      1790.5354 \\
    3.164      1440.7811 \\
    3.168      1175.4096 \\
    3.172      1079.8986 \\
    3.176      1187.0646 \\
    3.18      1467.2275 \\
    3.184      1837.6072 \\
    3.188      2187.9868 \\
    3.192      2414.6835 \\
    3.196      2452.3568 \\
    3.2      2293.8963 \\
    3.204      1992.3378 \\
    3.208      1644.2991 \\
    3.212       1360.109 \\
    3.216      1229.8632 \\
    3.22      1295.8481 \\
    3.224      1539.7744 \\
    3.228      1888.6697 \\
    3.232      2237.5378 \\
    3.236      2481.7482 \\
    3.24      2549.1649 \\
    3.244      2422.1489 \\
    3.248      2142.7255 \\
    3.252      1799.4046 \\
    3.256      1499.7876 \\
    3.26      1337.4366 \\
    3.264      1363.2072 \\
    3.268       1569.827 \\
    3.272      1894.3958 \\
    3.276      2237.9567 \\
    3.28      2496.0504 \\
    3.284      2590.8194 \\
    3.288      2494.7953 \\
    3.292       2239.104 \\
    3.296      1903.6396 \\
    3.3      1592.3066 \\
    3.304      1401.0033 \\
    3.308      1388.2098 \\
    3.312      1557.2022 \\
    3.316      1855.3001 \\
    3.32      2190.3023 \\
    3.324      2458.9827 \\
    3.328      2578.8413 \\
    3.332      2513.3459 \\
    3.336      2282.9406 \\
    3.34      1958.5084 \\
    3.344      1639.3734 \\
    3.348      1422.6713 \\
    3.352      1373.5308 \\
    3.356      1505.2222 \\
    3.36      1775.3212 \\
    3.364      2098.9935 \\
    3.368      2375.2428 \\
    3.372      2518.0012 \\
    3.376      2482.4927 \\
    3.38       2278.796 \\
    3.384      1968.4995 \\
    3.388      1645.5542 \\
    3.392      1407.2754 \\
    3.396      1324.4448 \\
    3.4      1419.6977 \\
    3.404      1660.7945 \\
    3.408      1970.7741 \\
    3.412      2251.7951 \\
    3.416      2415.2786 \\
    3.42        2409.07 \\
    3.424      2233.2893 \\
    3.428      1940.0567 \\
    3.432      1617.2506 \\
    3.436      1361.3595 \\
    3.44       1247.813 \\
    3.444      1307.9175 \\
    3.448      1519.4462 \\
    3.452      1813.7116 \\
    3.456       2096.874 \\
    3.46      2278.8754 \\
    3.464      2301.0799 \\
    3.468      2154.1379 \\
    3.472      1880.6333 \\
    3.476      1561.7702 \\
    3.48      1292.2608 \\
    3.484      1151.1805 \\
    3.488      1177.7577 \\
    3.492      1359.5123 \\
    3.496      1636.3271 \\
    3.5      1919.1281 \\
    3.504      2117.3739 \\
    3.508      2166.8676 \\
    3.512      2049.3521 \\
    3.516       1797.908 \\
    3.52      1486.5611 \\
    3.524       1207.364 \\
    3.528      1042.0473 \\
    3.532      1036.9678 \\
    3.536      1189.0393 \\
    3.54      1446.9096 \\
    3.544      1726.9487 \\
    3.548       1939.083 \\
    3.552      2014.4865 \\
    3.556      1926.6206 \\
    3.56      1699.1912 \\
    3.564      1398.6404 \\
    3.568      1113.5485 \\
    3.572      927.3352 \\
    3.576      892.6553 \\
    3.58      1015.3769 \\
    3.584      1253.0213 \\
    3.588       1527.989 \\
    3.592      1751.5739 \\
    3.596      1851.2538 \\
    3.6      1792.8874 \\
    3.604       1591.024 \\
    3.608      1304.2136 \\
    3.612      1016.8274 \\
    3.616      813.0505 \\
    3.62      750.9814 \\
    3.624      844.8505 \\
    3.628      1061.1592 \\
    3.632      1328.8354 \\
    3.636      1561.3716 \\
    3.64      1683.4711 \\
    3.644      1654.1074 \\
    3.648      1478.9669 \\
    3.652       1208.489 \\
    3.656      922.1774 \\
    3.66      704.1575 \\
    3.664      617.1971 \\
    3.668      682.6044 \\
    3.672      876.4097 \\
    3.676      1134.6274 \\
    3.68      1373.6211 \\
    3.684      1516.1994 \\
    3.688       1515.162 \\
    3.692      1367.6486 \\
    3.696      1115.8154 \\
    3.7      833.7004 \\
    3.704      604.8216 \\
    3.708      496.6746 \\
    3.712      533.0047 \\
    3.716      701.4068 \\
    3.72      947.6262 \\
    3.724      1190.7858 \\
    3.728      1352.4846 \\
    3.732      1379.8446 \\
    3.736       1261.514 \\
    3.74      1030.9621 \\
    3.744      756.0503 \\
    3.748      519.7885 \\
    3.752      397.7309 \\
    3.756      403.3789 \\
    3.76      533.6179 \\
    3.764      761.6541 \\
    3.768      1006.9747 \\
    3.772      1189.3727 \\
    3.776      1249.8039 \\
    3.78      1167.0297 \\
    3.784      963.9203 \\
    3.788      700.3536 \\
    3.792      459.0708 \\
    3.796      330.4834 \\
    3.8      309.5593 \\
    3.804      374.3647 \\
    3.808      555.1185 \\
    3.812      796.1464 \\
    3.816      1005.8817 \\
    3.82      1116.0404 \\
    3.824      1090.3398 \\
    3.828      934.4484 \\
    3.832      694.1583 \\
    3.836      446.5519 \\
    3.84      300.6025 \\
    3.844      257.2589 \\
    3.848      265.9422 \\
    3.852       335.639 \\
    3.856      515.0032 \\
    3.86      745.7809 \\
    3.864      928.9122 \\
    3.868      1003.8711 \\
    3.872      945.5962 \\
    3.876      770.1822 \\
    3.88      531.2076 \\
    3.884      329.8163 \\
    3.888      245.9653 \\
    3.892      220.2213 \\
    3.896      224.1782 \\
    3.9      260.3215 \\
    3.904      364.0937 \\
    3.908       566.006 \\
    3.912       764.424 \\
    3.916        878.36 \\
    3.92      870.2435 \\
    3.924      741.0864 \\
    3.928      530.6181 \\
    3.932      331.0958 \\
    3.936      238.3802 \\
    3.94      202.5806 \\
    3.944      190.9813 \\
    3.948      196.1474 \\
    3.952      220.2299 \\
    3.956      278.7496 \\
    3.96      415.8502 \\
    3.964      606.8359 \\
    3.968      744.6987 \\
    3.972      777.5063 \\
    3.976      693.6834 \\
    3.98      519.5292 \\
    3.984      334.2651 \\
    3.988      237.3928 \\
    3.992      195.1333 \\
    3.996      175.1759 \\
    4       168.018 \\
    4.004      171.0142 \\
    4.008      184.7296 \\
    4.012       213.547 \\
    4.016      271.4431 \\
    4.02      391.8225 \\
    4.024       551.512 \\
    4.028       655.699 \\
    4.032      660.8193 \\
    4.036      564.3408 \\
    4.04      404.0107 \\
    4.044      274.9113 \\
    4.048      211.5683 \\
    4.052      178.6605 \\
    4.056       160.249 \\
    4.06       150.797 \\
    4.064      148.1909 \\
    4.068      151.8054 \\
    4.072      162.0528 \\
    4.076      180.6542 \\
    4.08       211.959 \\
    4.084      266.9328 \\
    4.088      366.0427 \\
    4.092      488.3156 \\
    4.096      559.4598 \\
    4.1      544.5478 \\
    4.104      449.4319 \\
    4.108      325.5448 \\
    4.112      242.1148 \\
    4.116      195.9432 \\
    4.12      168.1127 \\
    4.124      150.3839 \\
    4.128      139.2947 \\
    4.132      133.2453 \\
    4.136      131.4904 \\
    4.14      133.7642 \\
    4.144      140.1724 \\
    4.148      151.2338 \\
    4.152       168.098 \\
    4.156      193.0979 \\
    4.16      231.0085 \\
    4.164      290.6149 \\
    4.168      374.5004 \\
    4.172      446.3075 \\
    4.176       461.589 \\
    4.18      411.5727 \\
    4.184      325.9451 \\
    4.188      253.5557 \\
    4.192      206.6415 \\
    4.196      175.9231 \\
    4.2      154.7959 \\
    4.204      139.9556 \\
    4.208      129.7055 \\
    4.212      123.1329 \\
    4.216      119.7389 \\
    4.22      119.2737 \\
    4.224      121.6593 \\
    4.228      126.9726 \\
    4.232      135.4659 \\
    4.236       147.633 \\
    4.24      164.3544 \\
    4.244      187.1577 \\
    4.248      218.6305 \\
    4.252       262.395 \\
    4.256      318.5292 \\
    4.26      369.7765 \\
    4.264      386.7398 \\
    4.268      358.5925 \\
    4.272      303.1748 \\
    4.276      249.5297 \\
    4.28      209.2008 \\
    4.284      180.1278 \\
    4.288      158.8174 \\
    4.292      142.9734 \\
    4.296      131.2337 \\
    4.3      122.7917 \\
    4.304      117.1674 \\
    4.308      114.0776 \\
    4.312      113.3722 \\
    4.316      114.9996 \\
    4.32      118.9981 \\
    4.324      125.5031 \\
    4.328      134.7746 \\
    4.332      147.2548 \\
    4.336      163.6544 \\
    4.34      185.0984 \\
    4.344      213.2298 \\
    4.348      249.7715 \\
    4.352      293.3866 \\
    4.356      332.4781 \\
    4.36      347.2877 \\
    4.364      328.8323 \\
    4.368      288.3905 \\
    4.372      245.5169 \\
    4.376      210.2292 \\
    4.38      183.1696 \\
    4.384      162.5565 \\
    4.388      146.8056 \\
    4.392      134.8459 \\
    4.396       125.992 \\
    4.4      119.8139 \\
    4.404      116.0482 \\
    4.408      114.5471 \\
    4.412        115.25 \\
    4.416      118.1714 \\
    4.42      123.4056 \\
    4.424       131.142 \\
    4.428      141.7025 \\
    4.432      155.6005 \\
    4.436       173.635 \\
    4.44      196.9803 \\
    4.444      227.0982 \\
    4.448      264.5244 \\
    4.452      304.5195 \\
    4.456      332.6613 \\
    4.46      333.7858 \\
    4.464      307.3695 \\
    4.468      268.0179 \\
    4.472      230.5724 \\
    4.476      200.3352 \\
    4.48      176.9721 \\
    4.484      159.0441 \\
    4.488      145.3597 \\
    4.492      135.1061 \\
    4.496      127.7569 \\
    4.5      122.9847 \\
    4.504      120.5977 \\
    4.508      120.5055 \\
    4.512      122.7036 \\
    4.516      127.2716 \\
    4.52      134.3877 \\
    4.524      144.3638 \\
    4.528      157.7059 \\
    4.532      175.1995 \\
    4.536      198.0184 \\
    4.54      227.6624 \\
    4.544      264.8604 \\
    4.548      305.3966 \\
    4.552        335.33 \\
    4.556       338.945 \\
    4.56      314.2891 \\
    4.564      275.1027 \\
    4.568      236.9471 \\
    4.572      206.0322 \\
    4.576       182.284 \\
    4.58      164.2406 \\
    4.584       150.654 \\
    4.588      140.6714 \\
    4.592      133.7467 \\
    4.596      129.5466 \\
    4.6      127.8878 \\
    4.604      128.7025 \\
    4.608      132.0253 \\
    4.612      138.0003 \\
    4.616      146.9122 \\
    4.62      159.2382 \\
    4.624      175.7425 \\
    4.628      197.6033 \\
    4.632      226.4264 \\
    4.636      263.4234 \\
    4.64      305.5693 \\
    4.644      339.7562 \\
    4.648      348.6722 \\
    4.652      327.3151 \\
    4.656      287.8683 \\
    4.66      247.5833 \\
    4.664      214.7063 \\
    4.668      189.6779 \\
    4.672      170.9235 \\
    4.676      157.0312 \\
    4.68      147.0375 \\
    4.684      140.3359 \\
    4.688      136.5665 \\
    4.692      135.5423 \\
    4.696      137.2148 \\
    4.7      141.6645 \\
    4.704      149.1131 \\
    4.708      159.9773 \\
    4.712      174.9456 \\
    4.716      195.1114 \\
    4.72      222.0547 \\
    4.724      257.3708 \\
    4.728      299.6941 \\
    4.732      338.3414 \\
    4.736      355.2088 \\
    4.74      340.9093 \\
    4.744      303.7464 \\
    4.748      261.6883 \\
    4.752      226.1872 \\
    4.756      199.1165 \\
    4.76      178.9953 \\
    4.764       164.235 \\
    4.768      153.7224 \\
    4.772      146.7538 \\
    4.776      142.9135 \\
    4.78      141.9852 \\
    4.784      143.9116 \\
    4.788      148.7785 \\
    4.792      156.8361 \\
    4.796      168.5446 \\
    4.8       184.661 \\
    4.804      206.3335 \\
    4.808      235.0343 \\
    4.812      271.4813 \\
    4.816      311.6548 \\
    4.82      342.3425 \\
    4.824      348.2787 \\
    4.828      326.3073 \\
    4.832      288.2889 \\
    4.836      249.8318 \\
    4.84       218.377 \\
    4.844      194.4259 \\
    4.848      176.5817 \\
    4.852      163.5377 \\
    4.856      154.3849 \\
    4.86      148.5414 \\
    4.864      145.6614 \\
    4.868      145.5648 \\
    4.872      148.2071 \\
    4.876      153.6676 \\
    4.88      162.1571 \\
    4.884      174.0448 \\
    4.888      189.8805 \\
    4.892      210.3532 \\
    4.896       235.969 \\
    4.9      265.8578 \\
    4.904      295.2976 \\
    4.908      314.7357 \\
    4.912      315.4008 \\
    4.916      297.0618 \\
    4.92      268.2006 \\
    4.924      238.4444 \\
    4.928      212.7361 \\
    4.932      192.0643 \\
    4.936      175.9378 \\
    4.94      163.6315 \\
    4.944      154.5422 \\
    4.948      148.2256 \\
    4.952      144.3771 \\
    4.956      142.7947 \\
    4.96       143.353 \\
    4.964      145.9833 \\
    4.968      150.6681 \\
    4.972      157.4212 \\
    4.976      166.2758 \\
    4.98      177.2436 \\
    4.984       190.238 \\
    4.988      204.9018 \\
    4.992      220.3343 \\
    4.996      234.7726 \\
    5      245.5826 \\
    5.004       250.063 \\
    5.008      246.9079 \\
    5.012      237.0623 \\
    5.016       223.068 \\
    5.02      207.6056 \\
    5.024      192.5517 \\
    5.028      178.8808 \\
    5.032      166.9656 \\
    5.036      156.8648 \\
    5.04      148.5024 \\
    5.044      141.7612 \\
    5.048      136.5141 \\
    5.052      132.6458 \\
    5.056      130.0488 \\
    5.06       128.629 \\
    5.064       128.295 \\
    5.068       128.961 \\
    5.072      130.5372 \\
    5.076      132.9277 \\
    5.08       136.024 \\
    5.084      139.6971 \\
    5.088      143.7938 \\
    5.092      148.1255 \\
    5.096      152.4663 \\
    5.1      156.5574 \\
    5.104      160.1129 \\
    5.108       162.846 \\
    5.112      164.5074 \\
    5.116      164.9195 \\
    5.12      164.0099 \\
    5.124      161.8249 \\
    5.128      158.5151 \\
    5.132      154.3016 \\
    5.136       149.437 \\
    5.14      144.1701 \\
    5.144      138.7208 \\
    5.148      133.2708 \\
    5.152      127.9608 \\
    5.156      122.8911 \\
    5.16      118.1349 \\
    5.164       113.738 \\
    5.168      109.7271 \\
    5.172      106.1151 \\
    5.176      102.9047 \\
    5.18      100.0901 \\
    5.184       97.658 \\
    5.188      95.5913 \\
    5.192      93.8705 \\
    5.196      92.4698 \\
    5.2      91.3639 \\
    5.204      90.5225 \\
    5.208      89.9149 \\
    5.212      89.5097 \\
    5.216      89.2717 \\
    5.22      89.1677 \\
    5.224      89.1626 \\
    5.228      89.2212 \\
    5.232      89.3098 \\
    5.236       89.395 \\
    5.24      89.4454 \\
    5.244      89.4318 \\
    5.248      89.3284 \\
    5.252       89.113 \\
    5.256      88.7676 \\
    5.26      88.2771 \\
    5.264      87.6328 \\
    5.268      86.8305 \\
    5.272      85.8699 \\
    5.276      84.7548 \\
    5.28      83.4924 \\
    5.284      82.0945 \\
    5.288      80.5748 \\
    5.292      78.9482 \\
    5.296      77.2317 \\
    5.3      75.4435 \\
    5.304      73.6006 \\
    5.308        71.72 \\
    5.312      69.8196 \\
    5.316      67.9145 \\
    5.32      66.0191 \\
    5.324      64.1483 \\
    5.328      62.3129 \\
    5.332      60.5236 \\
    5.336      58.7899 \\
    5.34      57.1201 \\
    5.344      55.5201 \\
    5.348      53.9951 \\
    5.352      52.5494 \\
    5.356      51.1853 \\
    5.36      49.9042 \\
    5.364      48.7071 \\
    5.368      47.5929 \\
    5.372      46.5613 \\
    5.376      45.6091 \\
    5.38      44.7338 \\
    5.384      43.9317 \\
    5.388      43.1986 \\
    5.392      42.5298 \\
    5.396      41.9201 \\
    5.4      41.3645 \\
    5.404      40.8569 \\
    5.408       40.391 \\
    5.412      39.9618 \\
    5.416      39.5622 \\
    5.42      39.1872 \\
    5.424      38.8299 \\
    5.428      38.4858 \\
    5.432      38.1484 \\
    5.436      37.8133 \\
    5.44      37.4756 \\
    5.444      37.1307 \\
    5.448      36.7754 \\
    5.452      36.4053 \\
    5.456      36.0188 \\
    5.46      35.6123 \\
    5.464      35.1851 \\
    5.468      34.7349 \\
    5.472      34.2616 \\
    5.476      33.7645 \\
    5.48      33.2437 \\
    5.484      32.7005 \\
    5.488      32.1353 \\
    5.492      31.5495 \\
    5.496      30.9453 \\
    5.5      30.3246 \\
    5.504      29.6895 \\
    5.508      29.0425 \\
    5.512      28.3862 \\
    5.516      27.7238 \\
    5.52      27.0579 \\
    5.524      26.3911 \\
    5.528      25.7263 \\
    5.532      25.0664 \\
    5.536      24.4139 \\
    5.54      23.7718 \\
    5.544      23.1421 \\
    5.548      22.5271 \\
    5.552      21.9287 \\
    5.556      21.3491 \\
    5.56      20.7897 \\
    5.564      20.2517 \\
    5.568      19.7364 \\
    5.572      19.2448 \\
    5.576      18.7771 \\
    5.58      18.3342 \\
    5.584      17.9157 \\
    5.588      17.5219 \\
    5.592      17.1521 \\
    5.596      16.8062 \\
};
\addlegendentry{R-FF Elastic Torque}
\addplot[line width=0.1mm] [color=red]
  table[row sep=crcr, x expr=\thisrow{X}, y expr=\thisrow{Y}*0.001]{
  X Y \\
      0  0 \\
    0 -1.1435e-14 \\
    0.04     0.17936 \\
    0.08      4.9753 \\
    0.12      32.3079 \\
    0.16      114.4971 \\
    0.2      278.8381 \\
    0.24      485.7489 \\
    0.28      672.4403 \\
    0.32      862.4259 \\
    0.36      1065.0721 \\
    0.4      1267.9428 \\
    0.44      1457.1985 \\
    0.48      1621.2374 \\
    0.52      1751.5986 \\
    0.56      1843.2342 \\
    0.6      1894.3429 \\
    0.64      1905.9153 \\
    0.68       1881.127 \\
    0.72      1824.6862 \\
    0.76      1742.2145 \\
    0.8      1639.7055 \\
    0.84      1523.0891 \\
    0.88      1397.9058 \\
    0.92      1269.0864 \\
    0.96      1140.8246 \\
    1      1016.5249 \\
    1.04      898.8096 \\
    1.08       789.568 \\
    1.12      690.0329 \\
    1.16      600.8726 \\
    1.2      522.2882 \\
    1.24      454.1094 \\
    1.28      395.8853 \\
    1.32       346.638 \\
    1.36       299.104 \\
    1.4      232.2786 \\
    1.44      118.7182 \\
    1.48     -63.4238 \\
    1.52     -322.8283 \\
    1.56     -653.3508 \\
    1.6     -1036.5717 \\
    1.64      -1932.798 \\
    1.68     -2353.8048 \\
    1.72      -2731.858 \\
    1.76     -3057.7598 \\
    1.8     -3315.0577 \\
    1.84     -3494.0655 \\
    1.88     -3591.5246 \\
    1.92      -3609.695 \\
    1.96     -3555.1439 \\
    2     -3437.4485 \\
    2.04     -3267.9629 \\
    2.08     -3058.7459 \\
    2.12     -2821.6959 \\
    2.16     -2567.9074 \\
    2.2     -2307.2373 \\
    2.24     -2048.0546 \\
    2.28     -1797.1417 \\
    2.32     -1559.7119 \\
    2.36     -1339.5106 \\
    2.4     -1138.9699 \\
    2.44     -959.3945 \\
    2.48     -801.1569 \\
    2.52     -663.8906 \\
    2.56     -546.6691 \\
    2.6     -448.1667 \\
    2.64     -366.7973 \\
    2.68     -300.8282 \\
    2.72     -248.4098 \\
    2.76      -205.633 \\
    2.8     -162.0553 \\
    2.84     -102.9604 \\
    2.88     -15.4338 \\
    2.92      107.2455 \\
    2.96      264.2113 \\
    3      447.9695 \\
    3.04      646.5687 \\
    3.08      846.1381 \\
    3.12      1305.2697 \\
    3.16      1695.7637 \\
    3.2      1825.5412 \\
    3.24       1917.011 \\
    3.28      1968.0391 \\
    3.32      1979.3186 \\
    3.36      1953.7978 \\
    3.4      1896.0389 \\
    3.44      1811.5952 \\
    3.48      1706.4606 \\
    3.52      1586.6186 \\
    3.56      1457.7013 \\
    3.6       1324.754 \\
    3.64      1192.0946 \\
    3.68      1063.2507 \\
    3.72      940.9593 \\
    3.76      827.2103 \\
    3.8      723.3203 \\
    3.84      630.0236 \\
    3.88      547.5703 \\
    3.92      475.8235 \\
    3.96       414.351 \\
    4      362.5091 \\
    4.04      319.5131 \\
    4.08      284.4879 \\
    4.12      256.5273 \\
    4.16      234.7428 \\
    4.2      218.2585 \\
    4.24      206.2706 \\
    4.28      198.0245 \\
    4.32      192.8558 \\
    4.36      190.1613 \\
    4.4      189.4232 \\
    4.44      190.1956 \\
    4.48      192.0956 \\
    4.52      194.8083 \\
    4.56      198.0704 \\
    4.6      201.6619 \\
    4.64      205.3859 \\
    4.68      209.0334 \\
    4.72      212.3409 \\
    4.76      214.9227 \\
    4.8      216.2523 \\
    4.84      215.6684 \\
    4.88         212.5 \\
    4.92      206.2254 \\
    4.96      196.6631 \\
    5      184.0676 \\
    5.04      169.0901 \\
    5.08      152.6316 \\
    5.12      135.6339 \\
    5.16       118.924 \\
    5.2      103.1191 \\
    5.24      88.6104 \\
    5.28      75.5945 \\
    5.32      64.1215 \\
    5.36      54.1408 \\
    5.4      45.5465 \\
    5.44       38.202 \\
    5.48      31.9631 \\
    5.52      26.6872 \\
    5.56      22.2425 \\
};
\addlegendentry{FB-FF Elastic Torque}
\draw[dashed, draw=black] (axis cs:3.0, 0) rectangle (axis cs:3.2, 2.8);
\addplot [color=black, dotted, forget plot]
  table[row sep=crcr]{%
3.0 0 \\
4.05	-3.95 \\
}; 
\addplot [color=black, dotted, forget plot]
  table[row sep=crcr]{%
3.2 2.8 \\
4.9	0 \\
}; 
\end{axis}
\begin{axis}[%
width=0.15\fwidth,
height=0.35\fheight,
at={(0.7\fwidth,0.11\fheight)}, 
scale only axis,
xmin=3.0,
xmax=3.2,
xtick = {3,3.1,3.2},
axis background/.style={fill=white},
xmajorgrids,
ymajorgrids,
legend style={legend cell align=left, align=left, draw=white!15!black}
]
\draw[ultra thin] (axis cs:\pgfkeysvalueof{/pgfplots/xmin},0) -- (axis cs:\pgfkeysvalueof{/pgfplots/xmax},0);
\draw[ultra thin] (axis cs:0,\pgfkeysvalueof{/pgfplots/ymin}) -- (axis cs:0,\pgfkeysvalueof{/pgfplots/ymax});
\draw[ultra thin] (axis cs:\pgfkeysvalueof{/pgfplots/xmin},0) -- (axis cs:\pgfkeysvalueof{/pgfplots/xmax},0);
\draw[ultra thin] (axis cs:0,\pgfkeysvalueof{/pgfplots/ymin}) -- (axis cs:0,\pgfkeysvalueof{/pgfplots/ymax});
\addplot [color=blue]
  table[row sep=crcr, x expr=\thisrow{X}, y expr=\thisrow{Y}*0.001]{
  X Y \\
    2.9992      281.8224 \\
    3.0032      606.0178 \\
    3.0072      955.1868 \\
    3.0112      1158.4633 \\
    3.0152       1159.572 \\
    3.0192      965.1662 \\
    3.0232      642.2596 \\
    3.0272      330.5123 \\
    3.0312      215.4374 \\
    3.0352       182.095 \\
    3.0392      179.9822 \\
    3.0432       208.299 \\
    3.0472       314.868 \\
    3.0512       653.123 \\
    3.0552      1048.3934 \\
    3.0592      1342.3604 \\
    3.0632      1451.0959 \\
    3.0672      1348.1959 \\
    3.0712      1072.3475 \\
    3.0752      715.2528 \\
    3.0792      404.2403 \\
    3.0832      279.4087 \\
    3.0872      276.2545 \\
    3.0912      392.5902 \\
    3.0952       713.364 \\
    3.0992      1109.6903 \\
    3.1032      1452.2727 \\
    3.1072      1642.8802 \\
    3.1112       1633.378 \\
    3.1152      1446.6933 \\
    3.1192      1176.4772 \\
    3.1232      949.6952 \\
    3.1272      872.1914 \\
    3.1312      990.7399 \\
    3.1352      1282.3755 \\
    3.1392      1666.3271 \\
    3.1432      2031.4654 \\
    3.1472      2271.5945 \\
    3.1512      2318.7066 \\
    3.1552      2164.0818 \\
    3.1592      1860.8417 \\
    3.1632      1507.3908 \\
    3.1672      1216.9975 \\
    3.1712      1082.9425 \\
    3.1752      1149.9618 \\
    3.1792      1400.7331 \\
    3.1832      1761.4749 \\
    3.1872      2124.8619 \\
    3.1912      2383.1359 \\
    3.1952      2461.1971 \\
    3.1992      2339.5156 \\
    3.2032      2059.8727 \\
    3.2072      1712.2481 \\
    3.2112      1406.9661 \\
    3.2152      1240.7221 \\
    3.2192      1266.9499 \\
    3.2232      1479.6108 \\
    3.2272      1815.3193 \\
    3.2312      2173.0708 \\
    3.2352      2445.4303 \\
    3.2392      2551.5541 \\
  };
\draw[ultra thin] (axis cs:\pgfkeysvalueof{/pgfplots/xmin},0) -- (axis cs:\pgfkeysvalueof{/pgfplots/xmax},0);
\draw[ultra thin] (axis cs:0,\pgfkeysvalueof{/pgfplots/ymin}) -- (axis cs:0,\pgfkeysvalueof{/pgfplots/ymax});
\addplot [color=red]
  table[row sep=crcr, x expr=\thisrow{X}, y expr=\thisrow{Y}*0.001]{
  X Y \\
    2.9992      444.1049 \\
    3.0032       463.487 \\
    3.0072      483.0092 \\
    3.0112      502.6585 \\
    3.0152      522.4209 \\
    3.0192      542.2818 \\
    3.0232      562.2264 \\
    3.0272      582.2394 \\
    3.0312      602.3059 \\
    3.0352       622.411 \\
    3.0392       642.541 \\
    3.0432      662.6821 \\
    3.0472      682.8213 \\
    3.0512      702.9453 \\
    3.0552      723.0407 \\
    3.0592      743.0936 \\
    3.0632      763.0899 \\
    3.0672      783.0149 \\
    3.0712      802.8544 \\
    3.0752      822.5951 \\
    3.0792      842.2258 \\
    3.0832      861.7412 \\
    3.0872      881.1514 \\
    3.0912      900.5103 \\
    3.0952      919.9998 \\
    3.0992      940.1798 \\
    3.1032      962.7143 \\
    3.1072      992.3528 \\
    3.1112      1041.2726 \\
    3.1152      1131.9791 \\
    3.1192      1274.0765 \\
    3.1232      1419.0958 \\
    3.1272       1513.705 \\
    3.1312      1563.6979 \\
    3.1352       1592.344 \\
    3.1392      1612.8467 \\
    3.1432      1630.3772 \\
    3.1472      1646.7239 \\
    3.1512      1662.4753 \\
    3.1552       1677.818 \\
    3.1592       1692.807 \\
    3.1632      1707.4546 \\
    3.1672      1721.7599 \\
    3.1712      1735.7185 \\
    3.1752      1749.3251 \\
    3.1792      1762.5748 \\
    3.1832      1775.4629 \\
    3.1872       1787.985 \\
    3.1912      1800.1364 \\
    3.1952      1811.9124 \\
    3.1992      1823.3081 \\
    3.2032      1834.3189 \\
    3.2072      1844.9403 \\
    3.2112      1855.1685 \\
    3.2152      1865.0003 \\
    3.2192      1874.4335 \\
    3.2232      1883.4665 \\
    3.2272      1892.0982 \\
    3.2312      1900.3275 \\
    3.2352      1908.1537 \\
    3.2392      1915.5753 \\
};
\end{axis}
\end{tikzpicture}

%% file: sec_results_v2.tex
\section{Experimental Results}
\label{sec:experiments}

\begin{figure}
    \centering
    \includegraphics[width = 0.4\textwidth]{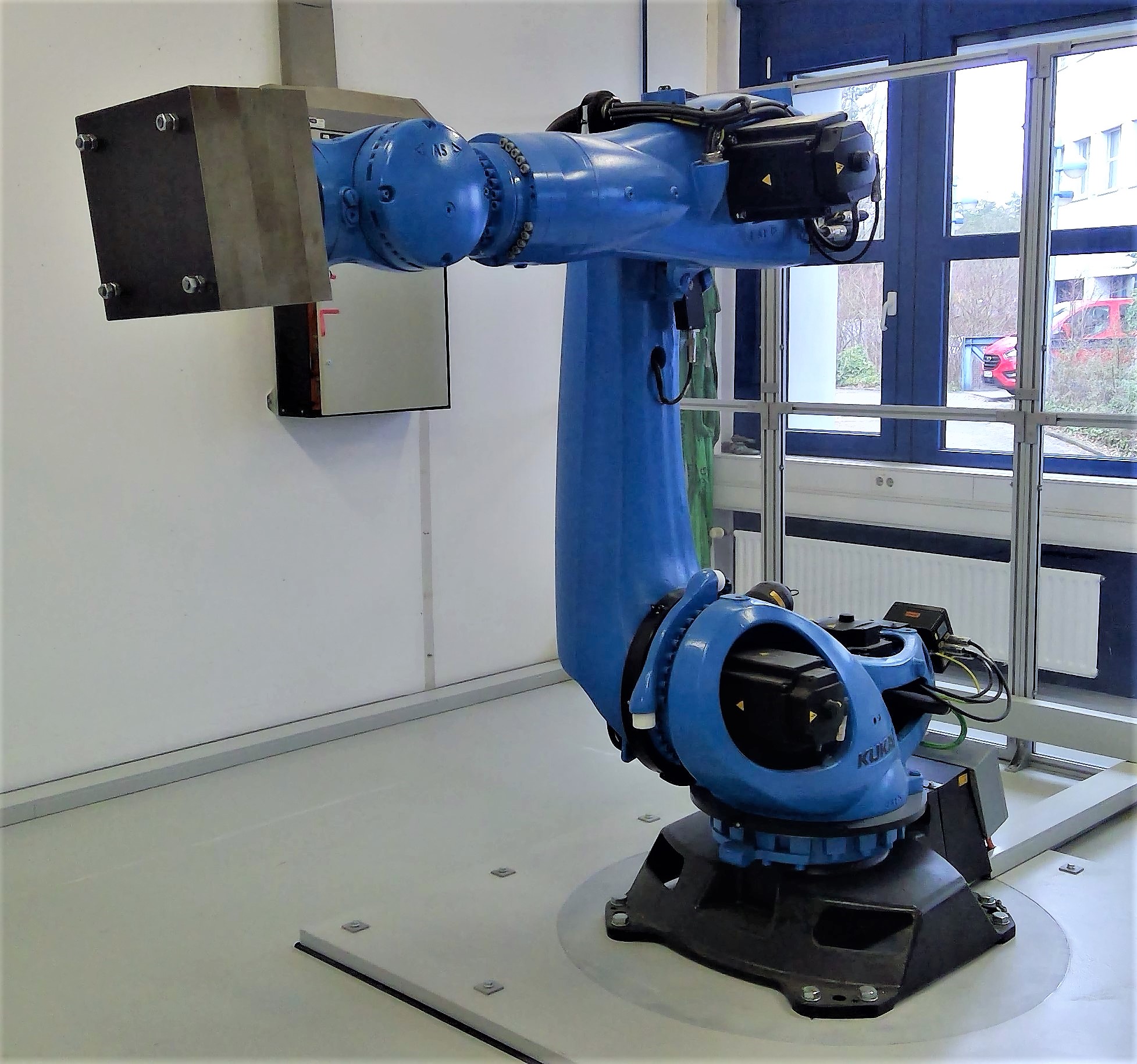}
    \caption{\textit{KUKA Quantec KR300 Ultra SE}.}
    \label{fig:robot_picture}
\end{figure}

The presented feed forward and feedback control laws are validated on a \textit{KUKA Quantec KR300 Ultra SE} robot, pictured in Fig.~\ref{fig:robot_picture}. 
The hardware and robot operating system are based on components from \textit{KUKA} and  industrial supplier \textit{B\&R Automation}. The \textit{B\&R Automation} robot operating system allows to implement the control algorithms as presented in  Fig.~\ref{fig:closed-loop-model}.
Except for the velocity control which runs in a task class of $0.2\, \mathrm{ms}$, all other presented algorithms run in a task class of $0.8\, \mathrm{ms}$. On the six-joint manipulator, SE are mounted on the base, shoulder and elbow joints, i.e. joints $1$, $2$ and $3$, and are utilized for position control as illustrated in Fig.~\ref{fig:closed-loop-model}. The motor encoders of the first three joints are used for velocity control only. Joints $4$, $5$ and $6$ are not equipped with SE and apply motor encoders for position and velocity control. Due to the long lever, we focus our results on the base, shoulder and elbow joints. Due to SE, it is possible to directly measure elastic joint effects and therefore validate the main contributions of this work.

As an experimental scenario, a full robot movement in Cartesian space was chosen. The movement of joints $1$, $2$ and $3$ can be measured using the same SE applied for position control. Each SE achieves an accuracy of $\pm 0.017 \ \mathrm{deg}$. All experiments are carried out on the cold robot, where both gearboxes and motors have approximately a temperature of $24.7$\textdegree{}C, measured before and after the experiments. This leads to a significant effect of nonlinear friction in the base, shoulder and elbow joints. The experiment has been carried out $10$ times in a row and the measurements are representative and reproducible. In order to account for a milling spindle in a robot machining application, and to increase the dynamical loads on each joint, we applied a payload of $150$~$kg$ on the robot's tool center point (TCP) as presented in Fig.~\ref{fig:robot_picture}.

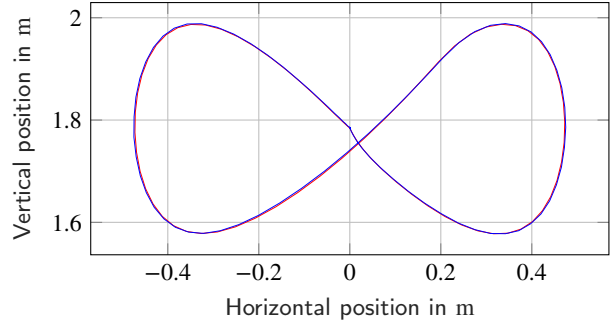
\begin{figure}
	\centering
	\setlength\fheight{3.5cm}
	\setlength\fwidth{7cm}
	\input{figs/ex_reference_plot_cart_v1}
	\caption{Reference and measured trajectory in Cartesian coordinates of the planar movement. Measurement in blue and reference in red.}
    \label{fig:reference_plot_cart}
\end{figure}

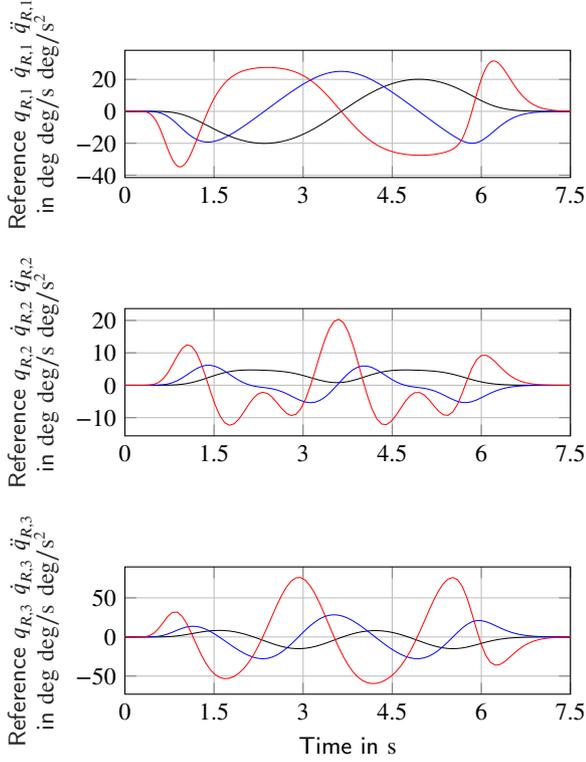
\begin{figure}
	\centering
	\setlength\fheight{3.5cm}
	\setlength\fwidth{7cm}
	\input{figs/ex_reference_plot_joint_v1}
	\caption{Joint space reference trajectories for each joint. Angular position in black, velocity in blue and acceleration in red.}
    \label{fig:reference_plot_joint}
\end{figure}

The Cartesian reference trajectory of the movement is shown in Fig.~\ref{fig:reference_plot_cart}. The Cartesian movement starts in a homing position, as presented in Fig.~\ref{fig:robot_picture}, and performs a planar eight-knot movement. The trajectory is based on the Lemniscate of Gerono, which can be parametrized as
\begin{align}
    x(t) &= a \, \cos( \varphi(t) ), \\
    y(t) &= b \, \sin( \varphi(t) ) \, \cos( \varphi(t) ) \nonumber
\end{align}
with the horizontal and vertical length parameters $a$, $b$. The Cartesian angle $\varphi(t)$ performs a $7^{th}$ order continuously differentiable acceleration and deceleration trajectory. We applied an additional joint space filter after estimating the inverse kinematics. In order to explicitly test effects of backlash, lost-motion and Coulomb friction we chose a Cartesian movement with several changes in direction and parts with link velocities close to zero. The joint space reference trajectory is shown in Fig.~\ref{fig:reference_plot_joint}.

Besides the Cartesian reference trajectory, Fig.~\ref{fig:reference_plot_cart} also presents the main result: the actual movement of the TCP based on the forward kinematics of the measured link angles. For a detailed discussion, we argue in the following with the measured joint angles and measured motor torques, instead of the estimated Cartesian coordinates with potential errors in the kinematic model. 

\begin{figure}
	\centering
	\setlength\fheight{3.5cm}
	\setlength\fwidth{7cm}
	\input{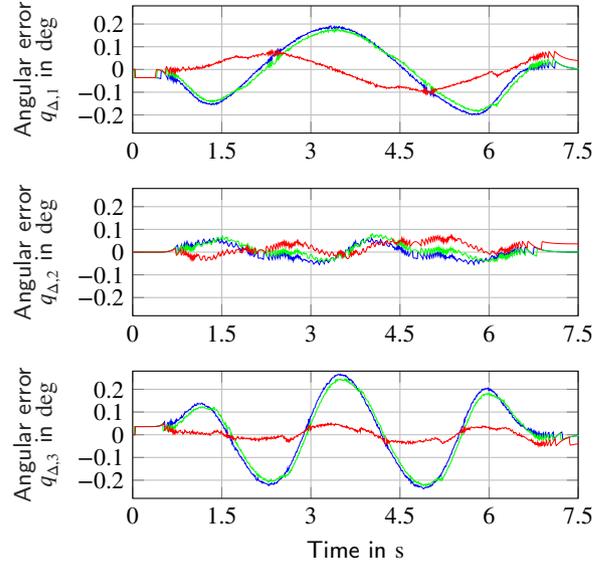}
	\caption{Measured angular error $q_{\Delta,i}$ of each joint. Error with conventional feedback controller (C-FB) in blue, model-based feedback controller (MB-FB) in green and flatness based feed forward controller (FB-FF) in red.}
    \label{fig:error_A_plot}
\end{figure}

The angular position improvements are shown in Fig.~\ref{fig:error_A_plot}, which present the angular displacement $q_{\Delta, i} = q_{R,i} - q_i $ of joint $1$, $2$ and $3$ for three cases. We apply the same control parameters, filter constants and dynamic model parameters in all cases. We apply and analyse a movement with a conventional feedback control (C-FB) law as a baseline. The subsequent measurement shows that a model-based feedback controller (MB-FF) leads to an improvement for all joints. A further improvement can be achieved applying the FB-FF and MB-FF. For joints~$1$, $2$ and $3$, the maximum path angular error does not exceed $\pm 0.08 \, \mathrm{deg}$. Note that the angular resolution of the SE is only $\pm 0.017 \, \mathrm{deg}$. Therefore, the achieved performance of the feedback control law is based only on $5$ measurement increments. All in all, the FB-FF and MB-FB improve the mean error of joint~$1$ by $49 \, \%$ and of joint~$3$ by $77 \, \%$ compared to C-FB. A detailed analysis of this improvement is presented in Tab.~\ref{tab:error}. We refer to $\text{mean}( \text{abs}( q_{\Delta, i}))$ as the mean angular error and to $\text{max}( \text{abs}( q_{\Delta, 1}))$ as the maximum angular error in Tab.~\ref{tab:error}.

\begin{table}
	\centering
	\caption{Measured maximum and mean angular errors for all presented algorithms.}
	\label{tab:error}
	\begin{tabularx}{\linewidth}{Xccccc} 
		\toprule
		algorithm & measure & joint~$1$ & joint~$2$ & joint~$3$ & unit \\
		\midrule
		C-FB & max & $ 0.200$ & $ 0.062$ & $ 0.266$ & $\mathrm{deg}$ \\
		C-FB & mean & $ 0.087$ & $ 0.020$ & $ 0.106$ & $\mathrm{deg}$ \\
		MB-FB & max & $ 0.184$ & $ 0.081$ & $ 0.245$ & $\mathrm{deg}$ \\
		MB-FB & mean & $ 0.081$ & $ 0.019$ & $ 0.101$ & $\mathrm{deg}$ \\
		R-FF & max & $ 0.135$ & $  0.053$ & $ 0.080$ & $\mathrm{deg}$ \\
		R-FF & mean & $0.052$ & $ 0.018$ & $ 0.034$ & $\mathrm{deg}$ \\
		FB-FF & max & $ 0.110$ & $  0.073$ & $ 0.050$ & $\mathrm{deg}$ \\
		FB-FF & mean & $0.044$ & $ 0.023$ & $ 0.024$ & $\mathrm{deg}$ \\
		\bottomrule
	\end{tabularx}
\end{table}

\begin{figure}
	\centering
	\setlength\fheight{3.5cm}
	\setlength\fwidth{7cm}
	\input{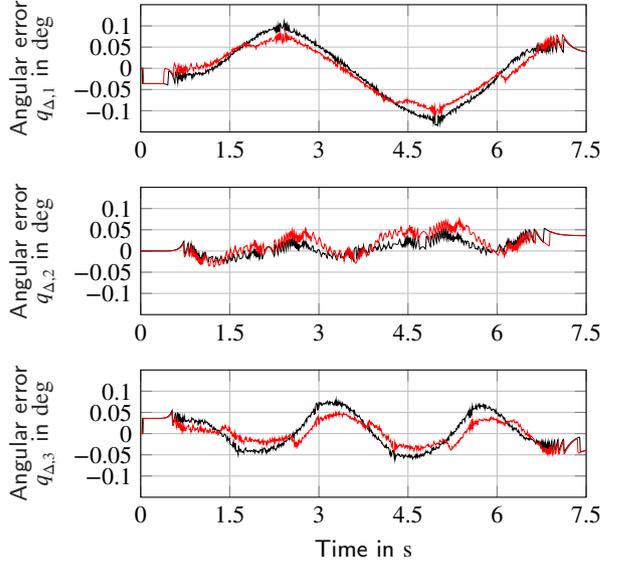}
	\caption{Measured angular error $q_{\Delta,i}$ of each joint. Flatness based feed forward controller (FB-FF) in red and nonlinear rigid model feed forward controller (R-FF) in black.}
    \label{fig:error_B_plot}
\end{figure}

As a more comprehensive validation, we compare nonlinear rigid model feed forward control (R-FF with C-FB) and FB-FF with MB-FB on the same trajectory. Model-based feedback control is not applicable if the rigid joint model neglects elasticity. The experimental result is shown in Fig.~\ref{fig:error_B_plot}. The flatness based controller leads to a better performance for joints~$1$ and $3$ than the nonlinear rigid model feed forward control law. The mean error for joint~$1$ is improved by $15 \, \%$ and for joint~$3$ by $29 \, \%$. The detailed analysis is given in Tab.~\ref{tab:error}. Regarding joint~$2$, both controllers achieve a similar performance of less than $0.073 \, \mathrm{deg}$ angular error. 

\begin{figure}
	\centering
	\setlength\fheight{3.5cm}
	\setlength\fwidth{7cm}
	\input{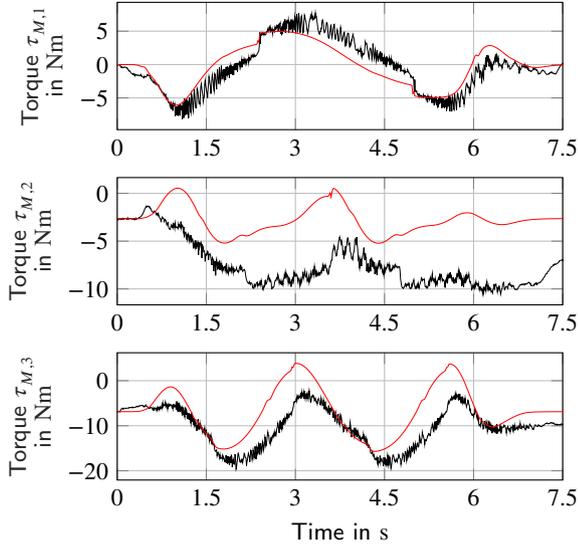}
	\caption{Measured motor torque $\tau_{M,i}$ of each joint in black. The stand-alone flatness based feed forward torque (FB-FF) executed is presented for comparison in red.}
    \label{fig:torque_plot}
\end{figure}

Fig.~\ref{fig:torque_plot} shows the measured motor torque for the proposed flatness based controller including a modified feedback controller for all major axis. For a comparison, we added the pure feed forward torque in the same figure. This measurement demonstrates that the elastic joint model captures the dynamics of joint~$1$ precisely. Regarding joints~$2$ and $3$, the measured motor torque significantly differs from the flatness based calculation. Note that the standstill measured motor torques at the beginning of the movement ($0 \, \mathrm{s}$) and at its end ($7.5 \, \mathrm{s}$) differ significantly. The measured joint angles, see Fig.~\ref{fig:error_A_plot}, confirm that the end pose is identical to the start pose within the measurement resolution of $\pm 0.017 \, \mathrm{deg}$. So, gravity loads and hydraulic spring loads on joint~$2$, which only depend on the pose, should be identical at the beginning and at the end. Inertia, Coriolis, centripetal and friction forces are zero in standstill. This leads to the conclusion that a significant asymmetrical friction is present in joint~$2$ and $3$. For joint~$2$, the hydraulic counterbalance reduces the gravity torque. Therefore, we neglected both, gravity torque and counterbalance for joint~$2$ in our model.

In a critical review we find that there is a considerable difference between simulation and real robot results. As remarked before, this is mainly due to model errors, sensor noise and external disturbances which are neglected in the presented simulation. The gap between simulation and experimental results decreases if, for example, model errors are included in the simulation. However, in this case the difference in feed forward torques between FB-FF and R-FF decreases, since both feed forward control algorithms can not account for model errors. Reviewing the feed forward torque comparison, simulation Fig.~\ref{fig:sim_ff_torque} demonstrates that the nonlinear rigid model feed forward controller (R-FF) already captures the major robot dynamics. A similar result is found in the angular displacement simulation in Fig.~\ref{fig:sim_error}, where the flatness based controller is better, but the R-FF controller nevertheless achieves an accuracy of $ 0.06 \, \mathrm{deg}$. We find this result to be surprising, since the R-FF does not model any joint elasticity. To be clear, the flatness based control is equal or better in all categories. The computed feed forward torque difference is just not significant compared to other error sources like noise, model errors and external disturbances.

The main advantages of the flatness based controller originate from a different perspective. First of all, note in Fig.~\ref{fig:sim_elastic_torque} the gearbox oscillations are caused by backlash and Coulomb friction. These oscillations feature a considerable amplitude of $ 1500 \, \mathrm{Nm}$. Using an accurate flatness based controller eliminates these oscillations. 

Second, all in all only the combined performance of feed forward and feedback controller is relevant. It is well-known, that a motor-side velocity controller can achieve a greater bandwidth than a link-side velocity controller. However, for implementing a motor-side velocity controller the calculation of the motor reference velocity $\dot{\theta}_{R,i} $ is non-trivial. Usually, elastic joints are neglected and an incorrect reference is applied, i.e. $\dot{\theta}_{R,i} = u_i \, \dot{q}_{R,i}$. In order to incorporate elastic joint effects, the full nonlinear model dynamics \eqref{eq:feedforward} needs to be taken into account. For a rigid joint model, like the nonlinear controller, it is impossible to calculate the correct motor reference velocity since elastic joint effects are neglected in the first place. The flatness based controller is able to fulfil this task and requires only a few microseconds of computation time due to AD. Therefore, the flatness based controller enables a better feedback velocity controller which is a core contribution of this work. We presented in experiments that the proposed algorithm (FB-FF with MB-FB) leads to an improvement of mean accuracy for joint~$3$ of $29\, \% $ compared to R-FF and of $ 77 \, \%$ compared to C-FB.


\section{Conclusion}
\label{sec:conculsion}
It was shown that the derived flatness based feed forward controller is capable of utilizing complex, nonlinear dynamical model of the joint flexibility and improving the position precision significantly. Joint backlash and lost-motion can be modeled using a novel nonlinear, continuously differentiable function. This enables standard AD tools, which allow fast run time cycles since the flatness based feed forward controller only requires a few microseconds of computation time online. 

During the experiments, we encountered significant difficulties in identifying nonlinear model parameters. Especially the experiments on joints $2$ and $3$ suggest, that an asymmetrical friction is present. Therefore, future works will further investigate on friction identification for the joints and the hydraulic counterbalance, including Stribeck and asymmetrical friction effects. Our future work will focus on online, nonlinear system identification. Parameter identification during run time enables model adaption to time-varying parameters, such as temperature dependent joint friction. 
Regarding the vision of robot machining applications, the improvement in trajectory tracking accuracy developed in this contribution enhances industrial robots capabilities for robot machining applications. Future works consider applying and evaluating this algorithm in a robot machining application.

%% file: figs/ex_reference_plot_cart_v1.tex
\begin{tikzpicture}
\begin{axis}[
width=1.2\fwidth,
height=0.7\fwidth,
xlabel={Horizontal position in $\mathrm{m}$},
xlabel style={font=\color{white!15!black}},
ylabel={Vertical position in $\mathrm{m}$},
ylabel style={font=\color{white!15!black}},
axis background/.style={fill=white},
xmajorgrids,
ymajorgrids,
legend cell align={left},
legend pos=north east
]
\addplot[line width=0.1mm, color=red]
    table[row sep=crcr]{
    8.3316e-17       1.784 \\
    -1.3399e-14       1.784 \\
    -1.6572e-10       1.784 \\
    -3.5422e-08       1.784 \\
    5.4731e-07       1.784 \\
    1.2433e-05       1.784 \\
    0.00011792      1.7838 \\
    0.00056953       1.783 \\
    0.0018349      1.7809 \\
    0.0045163      1.7762 \\
    0.0093841      1.7683 \\
    0.017519      1.7567 \\
    0.030233      1.7416 \\
    0.048554      1.7232 \\
    0.072736       1.702 \\
    0.10217      1.6789 \\
    0.13563      1.6551 \\
    0.17167      1.6321 \\
    0.20886      1.6115 \\
    0.24595      1.5948 \\
    0.28193      1.5832 \\
    0.31602      1.5777 \\
    0.34763      1.5788 \\
    0.37628      1.5868 \\
    0.40164      1.6016 \\
    0.42343      1.6229 \\
    0.44143        1.65 \\
    0.45546       1.682 \\
    0.46537      1.7178 \\
    0.47105       1.756 \\
    0.4724      1.7954 \\
    0.46941      1.8343 \\
    0.46209      1.8712 \\
    0.45049      1.9049 \\
    0.43474       1.934 \\
    0.41497      1.9576 \\
    0.39137      1.9746 \\
    0.36415      1.9846 \\
    0.33355      1.9872 \\
    0.30008       1.982 \\
    0.26511       1.968 \\
    0.23014      1.9448 \\
    0.19538       1.914 \\
    0.15941      1.8789 \\
    0.12029      1.8415 \\
    0.077375      1.8029 \\
    0.031435      1.7639 \\
    -0.016455      1.7259 \\
    -0.065382      1.6903 \\
    -0.11438      1.6582 \\
    -0.16234      1.6304 \\
    -0.20825      1.6079 \\
    -0.25129      1.5912 \\
    -0.29098      1.5812 \\
    -0.32701      1.5782 \\
    -0.35924      1.5823 \\
    -0.38759      1.5936 \\
    -0.41199      1.6116 \\
    -0.43238      1.6359 \\
    -0.44868      1.6656 \\
    -0.46081      1.6996 \\
    -0.46868      1.7369 \\
    -0.47222      1.7759 \\
    -0.47141      1.8152 \\
    -0.46624      1.8533 \\
    -0.45675      1.8887 \\
    -0.44305      1.9203 \\
    -0.42525      1.9467 \\
    -0.40352       1.967 \\
    -0.37806      1.9806 \\
    -0.34911      1.9869 \\
    -0.31697      1.9858 \\
    -0.28216      1.9775 \\
    -0.2455      1.9628 \\
    -0.20823      1.9433 \\
    -0.17187      1.9206 \\
    -0.13796      1.8969 \\
    -0.10777      1.8739 \\
    -0.082038      1.8532 \\
    -0.06095      1.8355 \\
    -0.044242      1.8212 \\
    -0.031395      1.8102 \\
    -0.021786       1.802 \\
    -0.014787      1.7961 \\
    -0.0098226       1.792 \\
    -0.0063928      1.7892 \\
    -0.0040805      1.7873 \\
    -0.0025591      1.7861 \\
    -0.0015792      1.7853 \\
    -0.00095987      1.7848 \\
    -0.00057666      1.7845 \\
    -0.00034076      1.7843 \\
    -0.00019983      1.7842 \\
    -0.00011638      1.7841 \\
    -6.6686e-05      1.7841 \\
    -3.7748e-05       1.784 \\
    -2.1292e-05       1.784 \\
    -1.2165e-05       1.784 \\
    -6.0706e-06       1.784 \\
    -3.0289e-06       1.784 \\
  };
  \addplot[line width=0.1mm, color=blue]
    table[row sep=crcr]{
    -0.00096782      1.7848 \\
    -0.00096782      1.7848 \\
    -0.00096782      1.7848 \\
    -0.00096786      1.7848 \\
    -0.00096881      1.7848 \\
    -3.4676e-05      1.7848 \\
    -3.317e-05      1.7847 \\
    0.00058463      1.7838 \\
    0.0014199      1.7818 \\
    0.0031992       1.778 \\
    0.0074487      1.7707 \\
    0.014685      1.7599 \\
    0.026336      1.7456 \\
    0.043428       1.728 \\
    0.066528      1.7075 \\
    0.095248      1.6849 \\
    0.1283      1.6609 \\
    0.16401      1.6378 \\
    0.20124      1.6168 \\
    0.23853      1.5986 \\
    0.27504      1.5857 \\
    0.30967      1.5782 \\
    0.34199      1.5778 \\
    0.37126      1.5837 \\
    0.39718      1.5971 \\
    0.42006      1.6163 \\
    0.43921       1.642 \\
    0.45412      1.6733 \\
    0.46514      1.7082 \\
    0.47173      1.7467 \\
    0.47455      1.7857 \\
    0.47246      1.8252 \\
    0.46591      1.8631 \\
    0.4551      1.8973 \\
    0.44031       1.928 \\
    0.42121      1.9534 \\
    0.39852      1.9722 \\
    0.372      1.9841 \\
    0.34199      1.9886 \\
    0.30908      1.9852 \\
    0.2746      1.9729 \\
    0.23986      1.9514 \\
    0.2054      1.9221 \\
    0.16974      1.8879 \\
    0.13062      1.8514 \\
    0.08757      1.8129 \\
    0.041509       1.774 \\
    -0.0066331       1.736 \\
    -0.05594      1.6998 \\
    -0.10548      1.6668 \\
    -0.15373       1.638 \\
    -0.20056      1.6137 \\
    -0.24414      1.5955 \\
    -0.28458      1.5836 \\
    -0.32127      1.5784 \\
    -0.35414      1.5809 \\
    -0.38328      1.5903 \\
    -0.40851      1.6068 \\
    -0.42988      1.6294 \\
    -0.44719       1.658 \\
    -0.46042      1.6911 \\
    -0.46946      1.7279 \\
    -0.47448      1.7666 \\
    -0.47409      1.8064 \\
    -0.47005      1.8449 \\
    -0.4614      1.8811 \\
    -0.44862      1.9134 \\
    -0.43145       1.942 \\
    -0.41064       1.964 \\
    -0.38592      1.9796 \\
    -0.35765      1.9878 \\
    -0.32609      1.9883 \\
    -0.29165      1.9814 \\
    -0.25517      1.9681 \\
    -0.21758      1.9493 \\
    -0.18054      1.9269 \\
    -0.14617      1.9034 \\
    -0.11518        1.88 \\
    -0.08794      1.8588 \\
    -0.065611      1.8403 \\
    -0.047491      1.8253 \\
    -0.033582      1.8131 \\
    -0.022945      1.8045 \\
    -0.015222      1.7974 \\
    -0.0092987      1.7929 \\
    -0.0056028      1.7893 \\
    -0.002798      1.7871 \\
    -0.00091687      1.7866 \\
    2.6543e-05      1.7852 \\
    0.00098063      1.7848 \\
    0.00097469      1.7846 \\
    0.00097143      1.7847 \\
    0.00096967       1.784 \\
    0.00096869      1.7839 \\
    0.00096814       1.784 \\
    0.00096781       1.784 \\
    0.00096806      1.7832 \\
    0.00096794      1.7832 \\
    0.00096788      1.7832 \\
    0.00096784      1.7832 \\
  };
\end{axis}
\end{tikzpicture}

%% file: figs/ex_reference_plot_joint_v1.tex
        \begin{tabular}{r}
            \begin{tikzpicture}
                \begin{axis}[
                    width=0.7\linewidth,
                    height=0.2\linewidth,
                    scale only axis,
                    xmin=0,
                    xmax=7.5,
                    xtick={0, 1.5, 3, 4.5, 6, 7.5},
                    ylabel= Reference $q_{R,1}$ $\dot{q}_{R,1}$ $\ddot{q}_{R,1}$ \\ in $\mathrm{deg}$ $\mathrm{deg}/\mathrm{s}$  $\mathrm{deg}/\mathrm{s^2}$,
                    ylabel style={font=\color{white!15!black}, align=center},
                    axis background/.style={fill=white},
                    xmajorgrids,
                    ymajorgrids
                    ]
                    \addplot [line width=0.1mm, color=black, on layer=foreground]
                    table[row sep=crcr, x expr=\thisrow{X}, y expr=\thisrow{Y}]{
                    X Y \\
    0  3.9887e-15 \\
    0.040004  3.8583e-15 \\
    0.080008 -3.2286e-12 \\
    0.12001  -8.495e-10 \\
    0.16002 -3.8197e-08 \\
    0.20002 -6.5531e-07 \\
    0.24002 -6.1235e-06 \\
    0.28003 -3.7681e-05 \\
    0.32003 -0.00017101 \\
    0.36004 -0.00061585 \\
    0.40004    -0.00185 \\
    0.44004  -0.0048037 \\
    0.48005   -0.011069 \\
    0.52005   -0.023095 \\
    0.56006    -0.04431 \\
    0.60006   -0.079148 \\
    0.64006    -0.13294 \\
    0.68007    -0.21167 \\
    0.72007    -0.32168 \\
    0.76008    -0.46922 \\
    0.80008     -0.6601 \\
    0.84008    -0.89929 \\
    0.88009     -1.1906 \\
    0.92009     -1.5366 \\
    0.9601     -1.9382 \\
    1.0001     -2.3949 \\
    1.0401     -2.9051 \\
    1.0801     -3.4656 \\
    1.1201     -4.0721 \\
    1.1601     -4.7199 \\
    1.2001     -5.4032 \\
    1.2401     -6.1161 \\
    1.2801     -6.8523 \\
    1.3201     -7.6056 \\
    1.3601     -8.3699 \\
    1.4001     -9.1393 \\
    1.4401     -9.9083 \\
    1.48015     -10.6716 \\
    1.52015     -11.4244 \\
    1.56016     -12.1624 \\
    1.60016     -12.8816 \\
    1.64016     -13.5784 \\
    1.68017     -14.2498 \\
    1.72017     -14.8929 \\
    1.76018     -15.5053 \\
    1.80018     -16.0848 \\
    1.84018     -16.6297 \\
    1.88019     -17.1382 \\
    1.92019     -17.6091 \\
    1.9602     -18.0412 \\
    2.0002     -18.4335 \\
    2.0402      -18.785 \\
    2.08021     -19.0951 \\
    2.12021     -19.3632 \\
    2.16022     -19.5889 \\
    2.20022     -19.7715 \\
    2.24022      -19.911 \\
    2.28023     -20.0069 \\
    2.32023     -20.0592 \\
    2.36024     -20.0676 \\
    2.40024      -20.032 \\
    2.44024     -19.9526 \\
    2.48025     -19.8292 \\
    2.52025     -19.6621 \\
    2.56026     -19.4512 \\
    2.60026     -19.1968 \\
    2.64026     -18.8993 \\
    2.68027     -18.5588 \\
    2.72027     -18.1759 \\
    2.76028     -17.7509 \\
    2.80028     -17.2846 \\
    2.84028     -16.7774 \\
    2.88029     -16.2303 \\
    2.92029      -15.644 \\
    2.9603     -15.0196 \\
    3.0003     -14.3581 \\
    3.0403     -13.6608 \\
    3.08031     -12.9291 \\
    3.12031     -12.1644 \\
    3.16032     -11.3684 \\
    3.20032     -10.5429 \\
    3.2403     -9.6897 \\
    3.2803     -8.8108 \\
    3.3203     -7.9085 \\
    3.3603      -6.985 \\
    3.4003     -6.0426 \\
    3.4403      -5.084 \\
    3.4803     -4.1115 \\
    3.5204      -3.128 \\
    3.5604     -2.1361 \\
    3.6004     -1.1385 \\
    3.6404      -0.138 \\
    3.6804     0.86255 \\
    3.7204      1.8604 \\
    3.7604      2.8528 \\
    3.8004       3.837 \\
    3.8404      4.8103 \\
    3.8804      5.7703 \\
    3.9204      6.7142 \\
    3.9604      7.6398 \\
    4.0004      8.5446 \\
    4.0404      9.4266 \\
    4.08041      10.2835 \\
    4.12041      11.1134 \\
    4.16042      11.9145 \\
    4.20042      12.6851 \\
    4.24042      13.4235 \\
    4.28043      14.1284 \\
    4.32043      14.7983 \\
    4.36044       15.432 \\
    4.40044      16.0284 \\
    4.44044      16.5865 \\
    4.48045      17.1055 \\
    4.52045      17.5845 \\
    4.56046      18.0228 \\
    4.60046      18.4198 \\
    4.64046       18.775 \\
    4.68047      19.0878 \\
    4.72047      19.3579 \\
    4.76048      19.5849 \\
    4.80048      19.7686 \\
    4.84048      19.9087 \\
    4.88049      20.0051 \\
    4.92049      20.0575 \\
    4.9605      20.0659 \\
    5.0005      20.0304 \\
    5.0405      19.9507 \\
    5.08051      19.8272 \\
    5.12051      19.6597 \\
    5.16052      19.4485 \\
    5.20052      19.1938 \\
    5.24052      18.8959 \\
    5.28053      18.5551 \\
    5.32053      18.1717 \\
    5.36054      17.7464 \\
    5.40054      17.2796 \\
    5.44054      16.7721 \\
    5.48055      16.2246 \\
    5.52055      15.6381 \\
    5.56056      15.0137 \\
    5.60056      14.3531 \\
    5.64056      13.6585 \\
    5.68057      12.9327 \\
    5.72057      12.1798 \\
    5.76058      11.4049 \\
    5.80058      10.6141 \\
    5.8406      9.8148 \\
    5.8806       9.015 \\
    5.9206      8.2232 \\
    5.9606       7.448 \\
    6.0006      6.6975 \\
    6.0406      5.9792 \\
    6.0806      5.2993 \\
    6.1206      4.6631 \\
    6.1606      4.0742 \\
    6.2006      3.5348 \\
    6.2406      3.0459 \\
    6.2806       2.607 \\
    6.3206      2.2169 \\
    6.3606      1.8732 \\
    6.4006      1.5731 \\
    6.4406      1.3133 \\
    6.4806        1.09 \\
    6.5207     0.89977 \\
    6.5607     0.73874 \\
    6.6007     0.60341 \\
    6.6407     0.49043 \\
    6.6807      0.3967 \\
    6.7207     0.31941 \\
    6.7607     0.25604 \\
    6.8007     0.20437 \\
    6.8407     0.16246 \\
    6.8807     0.12863 \\
    6.9207     0.10147 \\
    6.9607    0.079744 \\
    7.0007    0.062451 \\
    7.0407    0.048743 \\
    7.0807     0.03792 \\
    7.1207    0.029407 \\
    7.1607    0.022736 \\
    7.2007    0.017527 \\
    7.2407    0.013474 \\
    7.2807    0.010329 \\
    7.3207   0.0078981 \\
    7.3607   0.0060238 \\
    7.4007    0.004583 \\
    7.4407   0.0034786 \\
    7.4807   0.0026342 \\
    7.5208   0.0019905 \\
    7.5608   0.0015008 \\
    7.6008   0.0011292 \\
    7.6408  0.00084795 \\
    7.6808   0.0006355 \\
    7.7208  0.00047539 \\
    7.7608  0.00035496 \\
    7.8008  0.00026458 \\
    7.8408  0.00019687 \\
    7.8808  0.00014624 \\
    7.9208  0.00010846 \\
    7.9608  8.0314e-05 \\
                    };
                    \addplot [line width=0.1mm, color=blue, on layer=foreground]
                    table[row sep=crcr, x expr=\thisrow{X}, y expr=\thisrow{Y}]{
                    X Y \\
    0  0 \\
    0.040004 -5.1195e-14 \\
    0.080008 -5.6538e-10 \\
    0.12001 -9.5317e-08 \\
    0.16002 -3.0964e-06 \\
    0.20002 -4.0913e-05 \\
    0.24002  -0.0003065 \\
    0.28003  -0.0015542 \\
    0.32003  -0.0059295 \\
    0.36004   -0.018223 \\
    0.40004   -0.047263 \\
    0.44004    -0.10696 \\
    0.48005    -0.21644 \\
    0.52005    -0.39905 \\
    0.56006    -0.68011 \\
    0.60006     -1.0839 \\
    0.64006     -1.6304 \\
    0.68007     -2.3324 \\
    0.72007     -3.1935 \\
    0.76008     -4.2071 \\
    0.80008     -5.3568 \\
    0.84008     -6.6175 \\
    0.88009     -7.9578 \\
    0.92009     -9.3421 \\
    0.960096     -10.7336 \\
    1.0001     -12.0961 \\
    1.0401     -13.3965 \\
    1.08011     -14.6057 \\
    1.12011     -15.6999 \\
    1.16012     -16.6608 \\
    1.20012     -17.4758 \\
    1.24012     -18.1375 \\
    1.28013     -18.6431 \\
    1.32013     -18.9938 \\
    1.36014      -19.194 \\
    1.40014     -19.2506 \\
    1.44014     -19.1722 \\
    1.48015     -18.9688 \\
    1.52015     -18.6506 \\
    1.56016     -18.2283 \\
    1.60016     -17.7125 \\
    1.64016     -17.1131 \\
    1.68017     -16.4397 \\
    1.72017     -15.7013 \\
    1.76018     -14.9058 \\
    1.80018     -14.0607 \\
    1.84018     -13.1727 \\
    1.88019     -12.2476 \\
    1.92019     -11.2907 \\
    1.9602     -10.3067 \\
    2.0002     -9.2995 \\
    2.0402     -8.2728 \\
    2.0802     -7.2297 \\
    2.1202     -6.1729 \\
    2.1602     -5.1048 \\
    2.2002     -4.0275 \\
    2.2402     -2.9429 \\
    2.2802     -1.8526 \\
    2.3202    -0.75828 \\
    2.3602      0.3387 \\
    2.4002      1.4369 \\
    2.4402      2.5351 \\
    2.4802      3.6319 \\
    2.5203      4.7258 \\
    2.5603      5.8155 \\
    2.6003      6.8995 \\
    2.6403      7.9761 \\
    2.6803      9.0435 \\
    2.72027      10.0998 \\
    2.76028       11.143 \\
    2.80028      12.1707 \\
    2.84028      13.1807 \\
    2.88029      14.1702 \\
    2.92029      15.1366 \\
    2.9603      16.0768 \\
    3.0003      16.9878 \\
    3.0403      17.8665 \\
    3.08031      18.7095 \\
    3.12031      19.5135 \\
    3.16032       20.275 \\
    3.20032      20.9908 \\
    3.24032      21.6574 \\
    3.28033      22.2717 \\
    3.32033      22.8306 \\
    3.36034      23.3312 \\
    3.40034      23.7708 \\
    3.44034      24.1472 \\
    3.48035      24.4583 \\
    3.52035      24.7024 \\
    3.56036      24.8782 \\
    3.60036      24.9848 \\
    3.64036      25.0217 \\
    3.68037       24.989 \\
    3.72037      24.8869 \\
    3.76038      24.7162 \\
    3.80038      24.4781 \\
    3.84038      24.1741 \\
    3.88039      23.8062 \\
    3.92039      23.3765 \\
    3.9604      22.8874 \\
    4.0004      22.3416 \\
    4.0404      21.7421 \\
    4.08041      21.0917 \\
    4.12041      20.3936 \\
    4.16042       19.651 \\
    4.20042       18.867 \\
    4.24042      18.0448 \\
    4.28043      17.1875 \\
    4.32043      16.2981 \\
    4.36044      15.3795 \\
    4.40044      14.4347 \\
    4.44044      13.4661 \\
    4.48045      12.4764 \\
    4.52045      11.4679 \\
    4.56046      10.4428 \\
    4.6005      9.4032 \\
    4.6405      8.3508 \\
    4.6805      7.2876 \\
    4.7205       6.215 \\
    4.7605      5.1347 \\
    4.8005      4.0478 \\
    4.8405      2.9558 \\
    4.8805      1.8599 \\
    4.9205     0.76115 \\
    4.9605    -0.33918 \\
    5.0005       -1.44 \\
    5.0405     -2.5401 \\
    5.0805     -3.6383 \\
    5.1205     -4.7333 \\
    5.1605     -5.8238 \\
    5.2005     -6.9084 \\
    5.2405     -7.9853 \\
    5.2805      -9.053 \\
    5.32053     -10.1095 \\
    5.36054     -11.1527 \\
    5.40054     -12.1804 \\
    5.44054       -13.19 \\
    5.48055     -14.1782 \\
    5.52055       -15.14 \\
    5.56056     -16.0677 \\
    5.60056     -16.9493 \\
    5.64056     -17.7677 \\
    5.68057     -18.5001 \\
    5.72057     -19.1197 \\
    5.76058     -19.5976 \\
    5.80058      -19.906 \\
    5.84058     -20.0217 \\
    5.88059     -19.9288 \\
    5.92059     -19.6211 \\
    5.9606     -19.1026 \\
    6.0006     -18.3877 \\
    6.0406     -17.4992 \\
    6.08061     -16.4671 \\
    6.12061     -15.3251 \\
    6.16062     -14.1091 \\
    6.20062      -12.854 \\
    6.24062     -11.5926 \\
    6.28063     -10.3536 \\
    6.3206     -9.1608 \\
    6.3606      -8.033 \\
    6.4006      -6.984 \\
    6.4406     -6.0223 \\
    6.4806     -5.1527 \\
    6.5207     -4.3758 \\
    6.5607     -3.6897 \\
    6.6007     -3.0901 \\
    6.6407     -2.5713 \\
    6.6807     -2.1264 \\
    6.7207     -1.7482 \\
    6.7607     -1.4292 \\
    6.8007     -1.1621 \\
    6.8407    -0.94015 \\
    6.8807    -0.75686 \\
    6.9207    -0.60646 \\
    6.9607    -0.48378 \\
    7.0007    -0.38427 \\
    7.0407    -0.30398 \\
    7.0807    -0.23952 \\
    7.1207    -0.18802 \\
    7.1607    -0.14706 \\
    7.2007    -0.11463 \\
    7.2407   -0.089052 \\
    7.2807   -0.068961 \\
    7.3207   -0.053239 \\
    7.3607    -0.04098 \\
    7.4007   -0.031455 \\
    7.4407   -0.024077 \\
    7.4807   -0.018381 \\
    7.5208   -0.013997 \\
    7.5608   -0.010633 \\
    7.6008  -0.0080581 \\
    7.6408  -0.0060928 \\
    7.6808  -0.0045967 \\
    7.7208  -0.0034606 \\
    7.7608  -0.0025999 \\
    7.8008  -0.0019494 \\
    7.8408  -0.0014588 \\
    7.8808  -0.0010897 \\
    7.9208 -0.00081248 \\
    7.9608 -0.00060473 \\
                    };
                    \addplot [line width=0.1mm, color=red, on layer=foreground]
                    table[row sep=crcr, x expr=\thisrow{X}, y expr=\thisrow{Y}]{
                    X Y \\
    0  0 \\
    0.040004 -1.8324e-11 \\
    0.080008 -9.1219e-08 \\
    0.12001  -9.812e-06 \\
    0.16002 -0.00022874 \\
    0.20002  -0.0023106 \\
    0.24002   -0.013766 \\
    0.28003   -0.057012 \\
    0.32003    -0.18107 \\
    0.36004    -0.46984 \\
    0.40004     -1.0399 \\
    0.44004     -2.0247 \\
    0.48005     -3.5472 \\
    0.52005     -5.6891 \\
    0.56006     -8.4647 \\
    0.60006      -11.807 \\
    0.640064     -15.5698 \\
    0.680068     -19.5434 \\
    0.720072     -23.4803 \\
    0.760076     -27.1258 \\
    0.80008     -30.2459 \\
    0.840084     -32.6504 \\
    0.880088     -34.2072 \\
    0.920092     -34.8484 \\
    0.960096     -34.5678 \\
    1.0001     -33.4138 \\
    1.0401     -31.4768 \\
    1.08011     -28.8771 \\
    1.12011     -25.7508 \\
    1.16012     -22.2389 \\
    1.20012     -18.4778 \\
    1.24012     -14.5919 \\
    1.28013     -10.6896 \\
    1.3201     -6.8609 \\
    1.3601     -3.1766 \\
    1.4001      0.3106 \\
    1.4401      3.5647 \\
    1.4801       6.564 \\
    1.5202      9.2987 \\
    1.56016      11.7684 \\
    1.60016      13.9801 \\
    1.64016       15.946 \\
    1.68017      17.6815 \\
    1.72017      19.2045 \\
    1.76018      20.5337 \\
    1.80018      21.6877 \\
    1.84018      22.6849 \\
    1.88019      23.5427 \\
    1.92019      24.2773 \\
    1.9602      24.9034 \\
    2.0002      25.4344 \\
    2.0402      25.8822 \\
    2.08021      26.2571 \\
    2.12021      26.5683 \\
    2.16022      26.8234 \\
    2.20022      27.0289 \\
    2.24022      27.1899 \\
    2.28023      27.3108 \\
    2.32023      27.3944 \\
    2.36024      27.4431 \\
    2.40024       27.458 \\
    2.44024      27.4394 \\
    2.48025      27.3869 \\
    2.52025      27.2994 \\
    2.56026      27.1749 \\
    2.60026      27.0109 \\
    2.64026      26.8045 \\
    2.68027      26.5522 \\
    2.72027        26.25 \\
    2.76028      25.8939 \\
    2.80028      25.4794 \\
    2.84028      25.0022 \\
    2.88029       24.458 \\
    2.92029      23.8425 \\
    2.9603      23.1519 \\
    3.0003      22.3829 \\
    3.0403      21.5326 \\
    3.08031      20.5991 \\
    3.12031       19.581 \\
    3.16032      18.4783 \\
    3.20032      17.2916 \\
    3.24032      16.0231 \\
    3.28033      14.6756 \\
    3.32033      13.2536 \\
    3.36034      11.7625 \\
    3.40034      10.2089 \\
    3.4403      8.6002 \\
    3.4803       6.945 \\
    3.5204      5.2525 \\
    3.5604      3.5325 \\
    3.6004      1.7954 \\
    3.6404    0.051684 \\
    3.6804     -1.6881 \\
    3.7204     -3.4136 \\
    3.7604     -5.1147 \\
    3.8004     -6.7819 \\
    3.8404     -8.4062 \\
    3.8804     -9.9795 \\
    3.92039     -11.4948 \\
    3.9604     -12.9458 \\
    4.0004     -14.3274 \\
    4.0404     -15.6355 \\
    4.08041     -16.8671 \\
    4.12041     -18.0203 \\
    4.16042     -19.0942 \\
    4.20042     -20.0888 \\
    4.24042     -21.0048 \\
    4.28043     -21.8441 \\
    4.32043     -22.6087 \\
    4.36044     -23.3017 \\
    4.40044     -23.9261 \\
    4.44044     -24.4858 \\
    4.48045     -24.9845 \\
    4.52045     -25.4262 \\
    4.56046     -25.8149 \\
    4.60046     -26.1545 \\
    4.64046     -26.4488 \\
    4.68047     -26.7013 \\
    4.72047     -26.9151 \\
    4.76048     -27.0933 \\
    4.80048     -27.2381 \\
    4.84048     -27.3516 \\
    4.88049     -27.4352 \\
    4.92049     -27.4899 \\
    4.9605     -27.5161 \\
    5.0005     -27.5134 \\
    5.0405     -27.4813 \\
    5.08051     -27.4183 \\
    5.12051     -27.3227 \\
    5.16052     -27.1918 \\
    5.20052      -27.023 \\
    5.24052     -26.8127 \\
    5.28053     -26.5573 \\
    5.32053     -26.2526 \\
    5.36054     -25.8942 \\
    5.40054     -25.4759 \\
    5.44054     -24.9859 \\
    5.48055     -24.3971 \\
    5.52055     -23.6562 \\
    5.56056     -22.6746 \\
    5.60056     -21.3309 \\
    5.64056     -19.4871 \\
    5.68057     -17.0171 \\
    5.72057      -13.838 \\
    5.7606     -9.9368 \\
    5.8006     -5.3843 \\
    5.8406    -0.33245 \\
    5.8806      5.0022 \\
    5.92059      10.3652 \\
    5.9606      15.4952 \\
    6.0006       20.152 \\
    6.0406      24.1399 \\
    6.08061      27.3209 \\
    6.12061      29.6211 \\
    6.16062      31.0277 \\
    6.20062       31.581 \\
    6.24062      31.3631 \\
    6.28063      30.4844 \\
    6.32063      29.0711 \\
    6.36064      27.2537 \\
    6.40064      25.1584 \\
    6.44064      22.8997 \\
    6.48065      20.5768 \\
    6.52065      18.2716 \\
    6.56066      16.0477 \\
    6.60066      13.9519 \\
    6.64066      12.0155 \\
    6.68067       10.257 \\
    6.7207      8.6838 \\
    6.7607      7.2953 \\
    6.8007      6.0844 \\
    6.8407      5.0398 \\
    6.8807      4.1478 \\
    6.9207      3.3929 \\
    6.9607      2.7595 \\
    7.0007      2.2321 \\
    7.0407      1.7962 \\
    7.0807      1.4383 \\
    7.1207      1.1464 \\
    7.1607     0.90967 \\
    7.2007     0.71877 \\
    7.2407     0.56565 \\
    7.2807     0.44343 \\
    7.3207     0.34635 \\
    7.3607     0.26957 \\
    7.4007      0.2091 \\
    7.4407     0.16167 \\
    7.4807     0.12462 \\
    7.5208     0.09577 \\
    7.5608     0.07339 \\
    7.6008    0.056086 \\
    7.6408    0.042749 \\
    7.6808    0.032501 \\
    7.7208    0.024649 \\
    7.7608    0.018651 \\
    7.8008     0.01408 \\
    7.8408    0.010606 \\
    7.8808    0.007972 \\
    7.9208   0.0059801 \\
    7.9608   0.0044771 \\
                    };
                \end{axis}
            \end{tikzpicture} \\
            \begin{tikzpicture}
                \begin{axis}[
                    width=0.7\linewidth,
                    height=0.2\linewidth,
                    scale only axis,
                    xmin=0,
                    xmax=7.5,
                    xtick={0, 1.5, 3, 4.5, 6, 7.5},
ylabel= Reference $q_{R,2}$ $\dot{q}_{R,2}$ $\ddot{q}_{R,2}$ \\ in $\mathrm{deg}$ $\mathrm{deg}/\mathrm{s}$  $\mathrm{deg}/\mathrm{s^2}$,
                    ylabel style={font=\color{white!15!black}, align=center},
                    xmajorgrids,
                    ymajorgrids
                    ]
                    \addplot [line width=0.1mm, color=black, on layer=foreground]
                    table[row sep=crcr, x expr=\thisrow{X}, y expr=\thisrow{Y}+92.4941]{
                    X Y \\
    0     -92.4941 \\
    0.080008     -92.4941 \\
    0.160016     -92.4941 \\
    0.240024     -92.4941 \\
    0.320032     -92.4941 \\
    0.40004      -92.494 \\
    0.480048     -92.4935 \\
    0.560056     -92.4911 \\
    0.640064     -92.4834 \\
    0.720072     -92.4629 \\
    0.80008     -92.4177 \\
    0.880088     -92.3319 \\
    0.960096     -92.1886 \\
    1.0401     -91.9736 \\
    1.12011     -91.6797 \\
    1.20012     -91.3093 \\
    1.28013     -90.8745 \\
    1.36014     -90.3967 \\
    1.44014     -89.9026 \\
    1.52015     -89.4208 \\
    1.60016     -88.9778 \\
    1.68017     -88.5947 \\
    1.76018      -88.285 \\
    1.84018     -88.0537 \\
    1.92019     -87.8979 \\
    2.0002     -87.8082 \\
    2.08021     -87.7715 \\
    2.16022     -87.7735 \\
    2.24022     -87.8019 \\
    2.32023     -87.8477 \\
    2.40024     -87.9079 \\
    2.48025     -87.9847 \\
    2.56026     -88.0856 \\
    2.64026      -88.221 \\
    2.72027     -88.4022 \\
    2.80028     -88.6384 \\
    2.88029     -88.9337 \\
    2.9603     -89.2856 \\
    3.0403     -89.6834 \\
    3.12031     -90.1083 \\
    3.20032     -90.5351 \\
    3.28033     -90.9342 \\
    3.36034     -91.2745 \\
    3.44034     -91.5273 \\
    3.52035     -91.6694 \\
    3.60036     -91.6862 \\
    3.68037     -91.5735 \\
    3.76038     -91.3384 \\
    3.84038     -90.9986 \\
    3.92039     -90.5805 \\
    4.0004     -90.1162 \\
    4.08041     -89.6396 \\
    4.16042     -89.1826 \\
    4.24042     -88.7721 \\
    4.32043     -88.4272 \\
    4.40044      -88.158 \\
    4.48045     -87.9661 \\
    4.56046     -87.8453 \\
    4.64046     -87.7841 \\
    4.72047     -87.7687 \\
    4.80048     -87.7853 \\
    4.88049     -87.8232 \\
    4.9605     -87.8764 \\
    5.0405     -87.9443 \\
    5.12051     -88.0321 \\
    5.20052     -88.1489 \\
    5.28053     -88.3061 \\
    5.36054     -88.5142 \\
    5.44054     -88.7801 \\
    5.52055     -89.1048 \\
    5.60056     -89.4815 \\
    5.68057     -89.8952 \\
    5.76058     -90.3239 \\
    5.84058     -90.7428 \\
    5.92059     -91.1289 \\
    6.0006     -91.4654 \\
    6.08061     -91.7437 \\
    6.16062     -91.9632 \\
    6.24062     -92.1291 \\
    6.32063     -92.2497 \\
    6.40064     -92.3343 \\
    6.48065     -92.3919 \\
    6.56066     -92.4301 \\
    6.64066     -92.4547 \\
    6.72067     -92.4703 \\
    6.80068     -92.4799 \\
    6.88069     -92.4858 \\
    6.9607     -92.4893 \\
    7.0407     -92.4913 \\
    7.12071     -92.4925 \\
    7.20072     -92.4932 \\
    7.28073     -92.4936 \\
    7.36074     -92.4938 \\
    7.44074      -92.494 \\
    7.52075      -92.494 \\
    7.60076     -92.4941 \\
    7.68077     -92.4941 \\
    7.76078     -92.4941 \\
    7.84078     -92.4941 \\
    7.92079     -92.4941 \\
                    };
                    \addplot [line width=0.1mm, color=blue, on layer=foreground]
                    table[row sep=crcr, x expr=\thisrow{X}, y expr=\thisrow{Y}]{
                    X Y \\
    0  0 \\
    0.080008  2.5648e-11 \\
    0.16002  1.4024e-07 \\
    0.24002   1.405e-05 \\
    0.32003  0.00028465 \\
    0.40004    0.002515 \\
    0.48005    0.013552 \\
    0.56006    0.052232 \\
    0.64006     0.15618 \\
    0.72007     0.38055 \\
    0.80008     0.78279 \\
    0.88009      1.3978 \\
    0.9601      2.2157 \\
    1.0401      3.1742 \\
    1.1201      4.1673 \\
    1.2001      5.0678 \\
    1.2801      5.7541 \\
    1.3601      6.1335 \\
    1.4401       6.158 \\
    1.5202      5.8302 \\
    1.6002       5.199 \\
    1.6802      4.3486 \\
    1.7602      3.3824 \\
    1.8402      2.4058 \\
    1.9202      1.5101 \\
    2.0002     0.76031 \\
    2.0802     0.18749 \\
    2.1602    -0.21271 \\
    2.2402    -0.47653 \\
    2.3202    -0.66368 \\
    2.4002    -0.84539 \\
    2.4802     -1.0906 \\
    2.5603     -1.4524 \\
    2.6403     -1.9563 \\
    2.7203     -2.5935 \\
    2.8003     -3.3194 \\
    2.8803     -4.0587 \\
    2.9603     -4.7156 \\
    3.0403     -5.1878 \\
    3.1203     -5.3814 \\
    3.2003     -5.2257 \\
    3.2803     -4.6842 \\
    3.3603     -3.7622 \\
    3.4403     -2.5087 \\
    3.5204     -1.0135 \\
    3.6004     0.60246 \\
    3.6804      2.2003 \\
    3.7604        3.64 \\
    3.8404      4.7984 \\
    3.9204      5.5845 \\
    4.0004      5.9505 \\
    4.0804      5.8969 \\
    4.1604      5.4696 \\
    4.2404      4.7518 \\
    4.3204      3.8495 \\
    4.4004       2.876 \\
    4.4804      1.9355 \\
    4.5605      1.1097 \\
    4.6405     0.44857 \\
    4.7205   -0.034102 \\
    4.8005    -0.35918 \\
    4.8805    -0.57588 \\
    4.9605     -0.7513 \\
    5.0405    -0.95726 \\
    5.1205     -1.2563 \\
    5.2005     -1.6887 \\
    5.2805     -2.2634 \\
    5.3605      -2.953 \\
    5.4405     -3.6966 \\
    5.5206     -4.4063 \\
    5.6006     -4.9789 \\
    5.6806     -5.3154 \\
    5.7606     -5.3494 \\
    5.8406     -5.0733 \\
    5.9206     -4.5418 \\
    6.0006      -3.851 \\
    6.0806     -3.1058 \\
    6.1606     -2.3935 \\
    6.2406     -1.7707 \\
    6.3206     -1.2629 \\
    6.4006    -0.87175 \\
    6.4806     -0.5844 \\
    6.5607    -0.38165 \\
    6.6407    -0.24345 \\
    6.7207    -0.15205 \\
    6.8007   -0.093171 \\
    6.8807   -0.056113 \\
    6.9607   -0.033269 \\
    7.0407   -0.019444 \\
    7.1207   -0.011217 \\
    7.2007  -0.0063937 \\
    7.2807  -0.0036045 \\
    7.3607  -0.0020115 \\
    7.4407   -0.001112 \\
    7.5208  -0.0006094 \\
    7.6008 -0.00033127 \\
    7.6808 -0.00017872 \\
    7.7608 -9.5745e-05 \\
    7.8408 -5.0955e-05 \\
    7.9208 -2.6951e-05 \\
                    };
                    \addplot [line width=0.1mm, color=red, on layer=foreground]
                    table[row sep=crcr, x expr=\thisrow{X}, y expr=\thisrow{Y}]{
                    X Y \\
    0  0 \\
    0.080008  4.1321e-09 \\
    0.16002  1.0366e-05 \\
    0.24002  0.00063532 \\
    0.32003   0.0089487 \\
    0.40004    0.059563 \\
    0.48005     0.25388 \\
    0.56006     0.79144 \\
    0.64006      1.9253 \\
    0.72007      3.8087 \\
    0.80008      6.3239 \\
    0.88009      9.0306 \\
    0.960096      11.2839 \\
    1.0401      12.4496 \\
    1.12011      12.1045 \\
    1.20012      10.1487 \\
    1.2801      6.8107 \\
    1.3601      2.5717 \\
    1.4401     -1.9525 \\
    1.5202      -6.138 \\
    1.6002     -9.4617 \\
    1.68017     -11.5772 \\
    1.76018     -12.3521 \\
    1.84018     -11.8675 \\
    1.92019     -10.3854 \\
    2.0002     -8.2932 \\
    2.0802     -6.0359 \\
    2.1602     -4.0454 \\
    2.2402     -2.6764 \\
    2.3202     -2.1532 \\
    2.4002     -2.5358 \\
    2.4802     -3.7077 \\
    2.5603     -5.3906 \\
    2.6403     -7.1873 \\
    2.7203     -8.6464 \\
    2.8003     -9.3401 \\
    2.8803     -8.9384 \\
    2.9603     -7.2669 \\
    3.0403     -4.3379 \\
    3.1203     -0.3528 \\
    3.2003      4.3228 \\
    3.2803      9.2032 \\
    3.36034      13.7424 \\
    3.44034      17.4016 \\
    3.52035      19.7203 \\
    3.60036      20.3817 \\
    3.68037      19.2641 \\
    3.76038       16.466 \\
    3.84038      12.2998 \\
    3.9204      7.2496 \\
    4.0004      1.9012 \\
    4.0804     -3.1446 \\
    4.1604     -7.3594 \\
    4.24042     -10.3603 \\
    4.32043     -11.9556 \\
    4.40044     -12.1601 \\
    4.48045     -11.1779 \\
    4.5605      -9.359 \\
    4.6405     -7.1391 \\
    4.7205     -4.9714 \\
    4.8005     -3.2596 \\
    4.8805     -2.2999 \\
    4.9605     -2.2377 \\
    5.0405     -3.0439 \\
    5.1205     -4.5172 \\
    5.2005     -6.3121 \\
    5.2805     -7.9949 \\
    5.3605     -9.1163 \\
    5.4405     -9.2869 \\
    5.5206     -8.2345 \\
    5.6006     -5.8665 \\
    5.6806     -2.3994 \\
    5.7606      1.5663 \\
    5.8406      5.2164 \\
    5.9206      7.8612 \\
    6.0006       9.183 \\
    6.0806      9.2604 \\
    6.1606      8.4268 \\
    6.2406      7.0901 \\
    6.3206      5.6027 \\
    6.4006       4.204 \\
    6.4806      3.0202 \\
    6.5607      2.0905 \\
    6.6407      1.4013 \\
    6.7207      0.9133 \\
    6.8007     0.58075 \\
    6.8807      0.3613 \\
    6.9607     0.22043 \\
    7.0407     0.13216 \\
    7.1207    0.077992 \\
    7.2007    0.045373 \\
    7.2807    0.026056 \\
    7.3607    0.014786 \\
    7.4407   0.0082999 \\
    7.5208   0.0046126 \\
    7.6008   0.0025398 \\
    7.6808   0.0013865 \\
    7.7608  0.00075095 \\
    7.8408  0.00040372 \\
    7.9208  0.00021555 \\
                    };
                \end{axis}
            \end{tikzpicture} \\
            \begin{tikzpicture}
                \begin{axis}[
                    width=0.7\linewidth,
                    height=0.2\linewidth,
                    scale only axis,
                    xmin=0,
                    xmax=7.5,
                    xtick={0, 1.5, 3, 4.5, 6, 7.5},
                    xlabel={Time in $\mathrm{s}$},
ylabel= Reference $q_{R,3}$ $\dot{q}_{R,3}$ $\ddot{q}_{R,3}$ \\ in $\mathrm{deg}$ $\mathrm{deg}/\mathrm{s}$  $\mathrm{deg}/\mathrm{s^2}$,
                    ylabel style={font=\color{white!15!black}, align=center},
                    xmajorgrids,
                    ymajorgrids
                    ]
                        \addplot [line width=0.1mm, color=black, on layer=foreground]
                    table[row sep=crcr, x expr=\thisrow{X}, y expr=\thisrow{Y} - 92.4317]{
                    X Y \\
    0      92.4317 \\
    0.080008      92.4317 \\
    0.160016      92.4317 \\
    0.240024      92.4317 \\
    0.320032      92.4319 \\
    0.40004       92.434 \\
    0.480048      92.4455 \\
    0.560056      92.4864 \\
    0.640064      92.5939 \\
    0.720072      92.8178 \\
    0.80008      93.2077 \\
    0.880088      93.7959 \\
    0.960096      94.5846 \\
    1.0401      95.5419 \\
    1.12011      96.6069 \\
    1.20012      97.6998 \\
    1.28013      98.7345 \\
    1.36014      99.6284 \\
    1.440144      100.3104 \\
    1.520152      100.7247 \\
    1.60016      100.8317 \\
    1.680168      100.6077 \\
    1.760176      100.0434 \\
    1.84018      99.1423 \\
    1.92019        97.92 \\
    2.0002      96.4028 \\
    2.08021      94.6281 \\
    2.16022      92.6446 \\
    2.24022      90.5113 \\
    2.32023      88.2983 \\
    2.40024      86.0843 \\
    2.48025      83.9551 \\
    2.56026      81.9995 \\
    2.64026      80.3045 \\
    2.72027      78.9501 \\
    2.80028      78.0031 \\
    2.88029      77.5119 \\
    2.9603      77.5027 \\
    3.0403      77.9776 \\
    3.12031      78.9139 \\
    3.20032      80.2666 \\
    3.28033      81.9714 \\
    3.36034      83.9497 \\
    3.44034      86.1132 \\
    3.52035      88.3696 \\
    3.60036      90.6274 \\
    3.68037      92.8001 \\
    3.76038      94.8097 \\
    3.84038      96.5892 \\
    3.92039      98.0839 \\
    4.0004      99.2521 \\
    4.080408      100.0646 \\
    4.160416      100.5035 \\
    4.240424      100.5618 \\
    4.320432      100.2418 \\
    4.40044      99.5542 \\
    4.48045      98.5182 \\
    4.56046      97.1614 \\
    4.64046        95.52 \\
    4.72047      93.6398 \\
    4.80048      91.5765 \\
    4.88049      89.3959 \\
    4.9605      87.1733 \\
    5.0405      84.9919 \\
    5.12051      82.9397 \\
    5.20052      81.1054 \\
    5.28053      79.5733 \\
    5.36054      78.4174 \\
    5.44054      77.6958 \\
    5.52055       77.446 \\
    5.60056      77.6803 \\
    5.68057      78.3802 \\
    5.76058      79.4901 \\
    5.84058      80.9167 \\
    5.92059      82.5373 \\
    6.0006      84.2195 \\
    6.08061      85.8422 \\
    6.16062      87.3129 \\
    6.24062      88.5757 \\
    6.32063      89.6094 \\
    6.40064      90.4207 \\
    6.48065      91.0341 \\
    6.56066      91.4824 \\
    6.64066      91.8005 \\
    6.72067      92.0201 \\
    6.80068      92.1682 \\
    6.88069      92.2657 \\
    6.9607      92.3288 \\
    7.0407      92.3688 \\
    7.12071      92.3937 \\
    7.20072      92.4091 \\
    7.28073      92.4184 \\
    7.36074      92.4239 \\
    7.44074      92.4272 \\
    7.52075      92.4291 \\
    7.60076      92.4302 \\
    7.68077      92.4309 \\
    7.76078      92.4312 \\
    7.84078      92.4314 \\
    7.92079      92.4315 \\
                    };
                    \addplot [line width=0.1mm, color=blue, on layer=foreground]
                    table[row sep=crcr, x expr=\thisrow{X}, y expr=\thisrow{Y}]{
                    X Y \\
    0  0 \\
    0.080008  7.1181e-10 \\
    0.16002  3.8919e-06 \\
    0.24002  0.00038509 \\
    0.32003   0.0074378 \\
    0.40004    0.059049 \\
    0.48005     0.26837 \\
    0.56006     0.83288 \\
    0.64006      1.9605 \\
    0.72007      3.7444 \\
    0.80008      6.0729 \\
    0.88009      8.6354 \\
    0.960096      11.0114 \\
    1.0401      12.7898 \\
    1.12011      13.6638 \\
    1.20012      13.4758 \\
    1.28013      12.2145 \\
    1.3601      9.9821 \\
    1.4401      6.9508 \\
    1.5202       3.323 \\
    1.6002    -0.69761 \\
    1.6802     -4.9213 \\
    1.7602     -9.1778 \\
    1.84018     -13.3138 \\
    1.92019     -17.1873 \\
    2.0002     -20.6612 \\
    2.08021     -23.5987 \\
    2.16022     -25.8623 \\
    2.24022     -27.3172 \\
    2.32023      -27.839 \\
    2.40024     -27.3257 \\
    2.48025     -25.7131 \\
    2.56026     -22.9901 \\
    2.64026     -19.2114 \\
    2.72027     -14.5045 \\
    2.8003     -9.0675 \\
    2.8803     -3.1569 \\
    2.9603      2.9322 \\
    3.0403      8.8904 \\
    3.12031      14.4204 \\
    3.20032      19.2587 \\
    3.28033      23.1932 \\
    3.36034      26.0735 \\
    3.44034      27.8152 \\
    3.52035      28.3981 \\
    3.60036      27.8603 \\
    3.68037      26.2885 \\
    3.76038      23.8065 \\
    3.84038      20.5624 \\
    3.92039      16.7158 \\
    4.0004      12.4263 \\
    4.0804      7.8457 \\
    4.1604      3.1123 \\
    4.2404     -1.6498 \\
    4.3204     -6.3287 \\
    4.40044     -10.8196 \\
    4.48045     -15.0197 \\
    4.56046     -18.8223 \\
    4.64046     -22.1133 \\
    4.72047     -24.7711 \\
    4.80048      -26.669 \\
    4.88049     -27.6828 \\
    4.9605     -27.7018 \\
    5.0405     -26.6438 \\
    5.12051      -24.471 \\
    5.20052     -21.2052 \\
    5.28053     -16.9383 \\
    5.36054     -11.8345 \\
    5.4405      -6.124 \\
    5.5206   -0.092723 \\
    5.6006      5.9129 \\
    5.68057      11.4665 \\
    5.76058      16.0825 \\
    5.84058      19.3179 \\
    5.92059      20.9135 \\
    6.0006      20.8808 \\
    6.08061      19.4876 \\
    6.16062      17.1608 \\
    6.24062      14.3629 \\
    6.32063      11.4943 \\
    6.4006        8.84 \\
    6.4806       6.562 \\
    6.5607      4.7191 \\
    6.6407      3.2987 \\
    6.7207      2.2476 \\
    6.8007      1.4964 \\
    6.8807     0.97557 \\
    6.9607     0.62403 \\
    7.0407     0.39229 \\
    7.1207     0.24272 \\
    7.2007       0.148 \\
    7.2807    0.089044 \\
    7.3607    0.052914 \\
    7.4407    0.031087 \\
    7.5208    0.018071 \\
    7.6008    0.010402 \\
    7.6808   0.0059328 \\
    7.7608   0.0033551 \\
    7.8408   0.0018822 \\
    7.9208   0.0010481 \\
                    };
                    \addplot [line width=0.1mm, color=red, on layer=foreground]
                    table[row sep=crcr, x expr=\thisrow{X}, y expr=\thisrow{Y}]{
                    X Y \\
    0  0 \\
    0.080008  1.1475e-07 \\
    0.16002   0.0002875 \\
    0.24002    0.017291 \\
    0.32003     0.22689 \\
    0.40004      1.2951 \\
    0.48005      4.3652 \\
    0.560056      10.2087 \\
    0.640064      18.1945 \\
    0.720072       26.163 \\
    0.80008      31.3763 \\
    0.880088      31.7768 \\
    0.960096      26.7403 \\
    1.0401      17.0675 \\
    1.1201      4.4536 \\
    1.2001     -9.1699 \\
    1.28013     -22.1374 \\
    1.36014     -33.2989 \\
    1.44014     -42.0476 \\
    1.52015     -48.2119 \\
    1.60016     -51.8967 \\
    1.68017     -53.3307 \\
    1.76018     -52.7526 \\
    1.84018      -50.343 \\
    1.92019     -46.2008 \\
    2.0002     -40.3531 \\
    2.08021     -32.7896 \\
    2.16022     -23.5128 \\
    2.24022     -12.5944 \\
    2.3202    -0.23285 \\
    2.40024      13.2004 \\
    2.48025      27.1363 \\
    2.56026      40.8208 \\
    2.64026      53.3711 \\
    2.72027      63.8728 \\
    2.80028      71.5002 \\
    2.88029       75.631 \\
    2.9603      75.9305 \\
    3.0403      72.3871 \\
    3.12031      65.2961 \\
    3.20032      55.2046 \\
    3.28033      42.8332 \\
    3.36034      28.9922 \\
    3.44034      14.5056 \\
    3.5204     0.14715 \\
    3.60036     -13.4087 \\
    3.68037     -25.6222 \\
    3.76038     -36.1095 \\
    3.84038     -44.6494 \\
    3.92039     -51.1726 \\
    4.0004     -55.7355 \\
    4.08041     -58.4798 \\
    4.16042     -59.5855 \\
    4.24042     -59.2237 \\
    4.32043     -57.5196 \\
    4.40044     -54.5301 \\
    4.48045     -50.2401 \\
    4.56046     -44.5763 \\
    4.64046     -37.4373 \\
    4.72047     -28.7354 \\
    4.80048     -18.4463 \\
    4.8805      -6.662 \\
    4.9605      6.3611 \\
    5.0405      20.1683 \\
    5.12051      34.1029 \\
    5.20052       47.342 \\
    5.28053      58.9766 \\
    5.36054      68.1235 \\
    5.44054        74.03 \\
    5.52055      76.0215 \\
    5.60056      73.2133 \\
    5.68057      64.5791 \\
    5.76058      49.8526 \\
    5.84058      30.4653 \\
    5.9206      9.4485 \\
    6.0006     -9.6853 \\
    6.08061     -24.2241 \\
    6.16062     -32.9571 \\
    6.24062     -36.1373 \\
    6.32063     -34.9768 \\
    6.40064     -31.0415 \\
    6.48065     -25.7867 \\
    6.56066     -20.3129 \\
    6.64066     -15.3099 \\
    6.72067      -11.114 \\
    6.8007     -7.8107 \\
    6.8807     -5.3358 \\
    6.9607     -3.5549 \\
    7.0407     -2.3161 \\
    7.1207     -1.4792 \\
    7.2007    -0.92781 \\
    7.2807    -0.57253 \\
    7.3607    -0.34809 \\
    7.4407    -0.20877 \\
    7.5208    -0.12366 \\
    7.6008   -0.072416 \\
    7.6808   -0.041959 \\
    7.7608   -0.024074 \\
    7.8408   -0.013688 \\
    7.9208  -0.0077167 \\
                    };
                \end{axis}
            \end{tikzpicture}
        \end{tabular}

%% file: sec_appendix_v3.tex
\section{Robot Model Parameters}
\label{app:robo_para}
The Tab.~\ref{tab:alltableparameters} shows the parameters utilized of \textit{KUKA Quantec Ultra SE}.

\begin{table*}
	\centering
	\caption{Model parameters of \textit{KUKA Quantec KR300 Ultra SE}.}
	\label{tab:alltableparameters}
	\begin{tabularx}{\linewidth}{Xcccccccc} 
		\toprule
		description & symbol & joint~$1$ & joint~$2$ & joint~$3$ & joint~$4$ & joint~$5$ & joint~$6$ & unit \\
		\midrule
		Coulomb friction & $f_{c}$ & $ 200$ & $ 150$ & $ 180$ & $ 150$ & $ 150$ & $ 150 $ & $\mathrm{Nm}$ \\
		viscous friction & $f_{v}$ & $ 800$ & $ 500$ & $ 600$ & $ 100$ & $ 100$ & $ 100 $ & $\mathrm{Nms}/\mathrm{rad}$ \\
	    friction smoothness factor & $s_F$ & $ 500$ & $ 300$ & $ 100$ & $ 200$ & $ 200$ & $ 200 $ & $\mathrm{s}/\mathrm{rad}$ \\
		proportional speed gain  & $K_V$ & $ 0.015 $ & $ 0.015 $& $ 0.015 $& $ 0.015 $& $ 0.015 $& $ 0.015 $& $\mathrm{Nms}/\mathrm{rad}$ \\
		proportional position gain  & $K_P$ & $ 20 $ & $ 20 $& $ 20 $& $ 20 $& $ 20 $& $ 20 $& $1/\mathrm{s}$ \\
		backlash angle & $\phi_{B*} $ & $0.15 $ & $0.15 $ & $0.15 $ & $0.15 $ & $0.15 $ & $0.15 $ & $10^{-3}\quad \mathrm{rad}$ \\
		lost-motion angle & $\phi_{LM}$ & $0.15 $ & $0.15 $ & $0.15 $ & $0.15 $ & $0.15 $ & $0.15 $ & $10^{-3}\quad \mathrm{rad}$ \\
		torsional rigidity stiffness & $c_{TR}$ & $     8.4225  $ & $    8.9381  $ & $    5.5691  $ & $    1.6845  $ & $    1.6845  $ & $    1.0726  $ & $10^{6} \quad\mathrm{Nm}/\mathrm{rad}$ \\
		stiffness smoothness factor & $s_{E2}$ & $ 0.02$ & $ 0.015$ & $ 0.015$ & $ 0.015$ & $ 0.015$ & $ 0.015 $ & $\mathrm{1}/\mathrm{Nm}$ \\
		gearbox ratio & $u_G$ & $1798/7$ & $1872/7$ & $ 757/3$ & $ 221/1$ & $ 5032/21$ & $ 206793/1340 $ &  $-$\\
		motor inertia & $J$ & $ 0.0138 $ & $ 0.0177 $ & $ 0.0177$ & $ 0.0150$ & $ 0.0150$ & $ 0.0150 $ & $\mathrm{kg}\mathrm{m^2}$ \\
		\bottomrule
	\end{tabularx}
\end{table*}

\section{Continously Differentiable Stiffness}
\label{app:stiffness}

The idea for a continuously differentiable function for the nonlinear stiffness is borrowed from control theory. Consider a PT1-System with a ramp input and set the slope of the ramp equal to the linear stiffness.

\begin{equation}
T \, \frac{\diff y(t)}{\diff t}  +  y(t) = c_{TR} \, t
\label{eq:PT1_example}
\end{equation}

The system output slope will converge towards the input slope, with a constant time offset. Although we are not looking for a time-domain function at all, we can still use the algebraic solution of \eqref{eq:PT1_example}, by replacing the time variable $t$ with the torsion angle $\Delta q$, the output $y(t)$ with the elastic torque $ \tau_{E}(\Delta q) $ and setting all initial conditions to zero. In this work, we used a $3^{rd}$ order system to increase the curvature of the function. Further curvature can be achieved by applying an arbitrary higher order. If the order greater than $1$ is applied, two issues have to be considered. First, all poles of the transfer function must coincide. Second, the poles must be adapted to the order $n \in \mathbb{N}^{ + }$ of the ODE by setting $T = \phi_B/n$. This ensures that asymptote of the solution matches the full-contact stiffness. We define the variable-order ODE



\begin{equation}
\sum_{k = 0}^n    \left(\frac{\phi_B}{n} \right)^k \, {{n}\choose{k}} \, \frac{\diff^k y(t)}{\diff t^k} = c_{FC} \, t
\label{eq:var_order_example}
\end{equation}

with the binomial coefficient $ {{n}\choose{k}} $.

\section{Feed Forward Controller}
\label{app:code}

We present the flatness based control algorithm, which is obtained by \textit{MATLAB Symbolic Toolbox} with additional manual modifications and transferred into C. We limited all exponential functions to $exp( \cdot ) \leq 1e30$ in order to avoid Not-A-Number and infinity errors when using division. We extended the code for estimating all robot joints.

\begin{lstlisting}
double[3][6] res = feedforward(
    double[6] q0, double[6] q1,
    double[6] q2, double[6] q3,
    double[6] q4, double[6] qB,
    double[6] J, double[6] M,
    double[6] sE, double[6] sF,
    double[6] f_v, double[6] f_c,
    double[6] u, double[6] c,
    double[6] tau_c, double[6] tau_g,
    double[6] tau_a)
{

    /* Description */
    // Computes the flatness based torque
    // and motor reference variables for
    // each axis of an industrial robot.

    /* Input */
    // Variables and parameters of
    // the current state for all 6 axis.
    // Type double.

    /* Output */
    // Result matrix. First column is
    // the flatness based torque in Nm,
    // second column is the motor reference
    // angle in rad, third column is
    // the motor reference velocity in rad/s.
    // Type double.

    /* define variables */
    // define axis index
    int i_ax;
    // define temporary variables
    double t[36];
    // limit exponential function
    double exp_max = 1e30;
    // define variables for each axis
    double M_i, J_i, tau_c_i, tau_g_i, tau_a_i,
        c_i, qB_i, sF_i, sE_i, f_v_i, f_c_i,
        u_i, q0_i, q1_i, q2_i, q3_i, q4_i;

    /* for each axis */
    for (i_ax = 0; i_ax < 6; i_ax++)
    {

        /* get parameters for each axis */
        // link inertia, kgm^2
        M_i = M[i_ax];
        // motor inertia, kgm^2
        J_i = J[i_ax];
        // backlash angle, rad
        qB_i = qB[i_ax];
        // Coriolis and centripetal torque, Nm
        tau_c_i = tau_c[i_ax];
        // gravity torque, Nm
        tau_g_i = tau_g[i_ax];
        // acceleration torque, Nm
        tau_a_i = tau_a[i_ax];
        // transmission factor, without unit
        u_i = u[i_ax];
        // stiffness factor, Nm/rad
        c_i = c[i_ax];
        // friction smoothness factor, s/rad
        sF_i = sF[i_ax];
        // elastic torque smoothness factor, 
        // rad/Nm
        sE_i = sE[i_ax];
        // viscoul friction coefficient, Nms/rad
        f_v_i = f_v[i_ax];
        // coulomb friction coefficient, Nm
        f_c_i = f_c[i_ax];

        /* get variables for each axis */
        q0_i = q0[i_ax]; // link angle
        q1_i = q1[i_ax]; // link velocity
        q2_i = q2[i_ax]; // link acceleration
        q3_i = q3[i_ax]; // link jerk
        q4_i = q4[i_ax]; // link jerk derivative

        /* compute feed forward torque, 
        motor angle and motor velocity */
        t[0] = M_i*q2_i;
        t[1] = M_i*q3_i;
        t[2] = M_i*q4_i;
        t[3] = f_v_i*q1_i;
        t[4] = f_v_i*q2_i;
        t[5] = f_v_i*q3_i;
        t[6] = q1_i*sF_i;
        t[7] = q2_i*q2_i;
        t[8] = sE_i*sE_i;
        t[9] = sF_i*sF_i;
        t[10] = -f_c_i;
        t[11] = 1.0/c_i;
        t[12] = t[6]*2.0;
        t[13] = -t6;
        t[14] = -t12;
        t[15] = fmin(exp(t13), exp_max);
        t[16] = fmin(exp(t14), exp_max);
        t[17] = t[15]+1.0;
        t[18] = 1.0/t17;
        t[19] = t[18]*t[18];
        t[20] = t[18]*t[18];
        t[21] = f_c_i*t[18]*2.0;
        t[22] = f_c_i*q2_i*sF_i 
               *t[15]*t[19]*2.0;
        t[23] = f_c_i*q3_i*sF_i 
               *t[15]*t[19]*2.0;
        t[24] = f_c_i*t[7]*t[9] 
               *t[15]*t[19]*2.0;
        t[25] = f_c_i*t[7]*t[9] 
               *t[16]*t[20]*4.0;
        t[26] = t[0]+t[3]+t[10]+t[21] 
               +tau_c_i+tau_g_i+tau_a_i;
        t[27] = -t[24];
        t[28] = sE_i*t[26];
        t[29] = t[1]+t[4]+t[22];
        t[30] = -t[28];
        t[31] = t[29]*t[29];
        t[32] = t[2]+t[5]+t[23]+t[25] 
               +t[27];
        t[33] = fmin(exp(t[30]), exp_max);
        t[34] = t[33]+1.0;
        t[35] = 1.0/(t[34]*t[34]);

        /* feed forward torque in Nm */
        res[0][i_ax] = t[26]/u_i+J_i*u_i*
                        (q2_i+t[11]*t[32] 
                       +qB_i*sE_i*t[33] 
                       *t[35]*t[32]*2.0 
                       -qB_i*t[8]*t[31] 
                       *t[33]*t[35]*2.0 
                       +qB_i*t[8]*t[31] 
                       *1.0/(t[34]*t[34] 
                       *t[34])*fmin(exp(t[28] 
                       *-2.0), exp_max)*4.0);
        
        /* motor reference angle in rad */
        res[1][i_ax] = u_i*(q0_i-qB_i 
                       +t[11]*t[26] 
                       +(qB_i*2.0)/t[34]);
        
        /* motor reference velocity in rad/s */
        res[2][i_ax] = u_i*(q1_i+t[11] 
                       *t[29]+qB_i*sE_i 
                       *t[29]*t[33] 
                       *t[35]*2.0);
    }
}
 \end{lstlisting}

